\documentclass[sigconf]{acmart}
\AtBeginDocument{%
  }

\usepackage{xcolor}
\usepackage{multirow}
\usepackage{adjustbox}

\usepackage{graphicx}
\usepackage{subcaption}
\usepackage{bm}
\usepackage{tcolorbox}
\usepackage{balance}

\usepackage{pifont}
\newcommand{\xmark}{\ding{55}}
\newcommand{\cmark}{\ding{51}}
\newcommand*\colourcheck[1]{%
  \expandafter\newcommand\csname #1check\endcsname{\textcolor{#1}{\cmark}}%
}
\newcommand*\colourx[1]{%
  \expandafter\newcommand\csname #1x\endcsname{\textcolor{#1}{\xmark}}%
}
\colourcheck{green}
\colourx{red}
\colourcheck{teal}

\usepackage{tabularx}
\usepackage{makecell}
\usepackage{multirow}
\usepackage{comment}
\usepackage{booktabs}
\usepackage{subcaption}
\usepackage{algorithm}
\usepackage{algpseudocode}
\usepackage{tcolorbox}
\usepackage{placeins}
\newcolumntype{Y}{>{\centering\arraybackslash}X}
\newcolumntype{P}[1]{>{\centering\arraybackslash}p{#1}}

\DeclareMathSizes{10}{9}{7}{5}

\copyrightyear{2026}
\acmYear{2026}
\setcopyright{cc}
\setcctype{by}
\acmConference[MM '26]{Proceedings of the 34th ACM International Conference on Multimedia}{November 10--14, 2026}{Rio de Janeiro, Brazil}
\acmBooktitle{Proceedings of the 34th ACM International Conference on Multimedia (MM '26), November 10--14, 2026, Rio de Janeiro, Brazil}
\acmDOI{10.1145/3767308.3836090}
\acmISBN{979-8-4007-2213-4/2026/11}

\begin{document}

\title{A Large-scale Evaluation of Text-guided Models for Facial Editing}

\author{Rahul Nair}
\email{rnair21@asu.edu}
\affiliation{%
  \institution{Arizona State University}
  \city{Tempe}
  \state{AZ}
  \country{USA}
}

\author{Saurav Pandit}
\email{saurav@prelight.ai}
\affiliation{%
  \institution{Prelight Inc}
  \city{New York}
  \country{USA}
}

\author{Hannah Kerner}
\email{hkerner@asu.edu}
\affiliation{%
  \institution{Arizona State University}
  \city{Tempe}
  \state{AZ}
  \country{USA}
}

\renewcommand{\shortauthors}{Rahul Nair, Saurav Pandit, and Hannah Kerner}


\begin{abstract}
Facial appearance editing powers popular applications like FaceApp and Photoshop. Generative Adversarial Networks (GANs) and 3D Morphable Models (3DMMs) have been widely used for facial editing. GANs can perform varied facial edits (e.g., changing hair color, hairstyle), but often produce unstable edits. 3DMMs produce stable edits, but can only alter pose and facial expression. Recently, text-guided diffusion models like Nano Banana have become popular for image editing. Text-guided models are a compelling alternative to GANs and 3DMMs since they can produce both stable and varied image edits. While text-guided models have been widely tested for whole-scene edits (e.g., ``make the woman play a guitar''), they have not been comprehensively tested for facial editing. We conducted the first large-scale evaluation ($\sim1$M images evaluated) of six popular text-guided models on a sequential facial editing task. We present Face-Edit-Attributes, the largest collection of $169$ facial editing attributes focused on hair, accessories, and pose edits. We compared model performance using two popular celebrity face datasets: CelebA and CelebSET. Our results show that most models performed hair and accessory edits well, but struggled with editing pose. All models over-edit (e.g., changing hair color when asked only to change the hairstyle). We also evaluated demographic biases in each model. Our results show surprising biases in overediting: almost all models created more overedits for dark-skinned male faces and old faces. The code and data for our results (including our repository of $\sim 1$M images) can be accessed \href{https://github.com/rahul1801/Face-Edit-Bench}{\textcolor{blue}{here}}.
\end{abstract}






\maketitle


\begin{figure*}[t]
  \centering
  \includegraphics[width=1.00\linewidth]{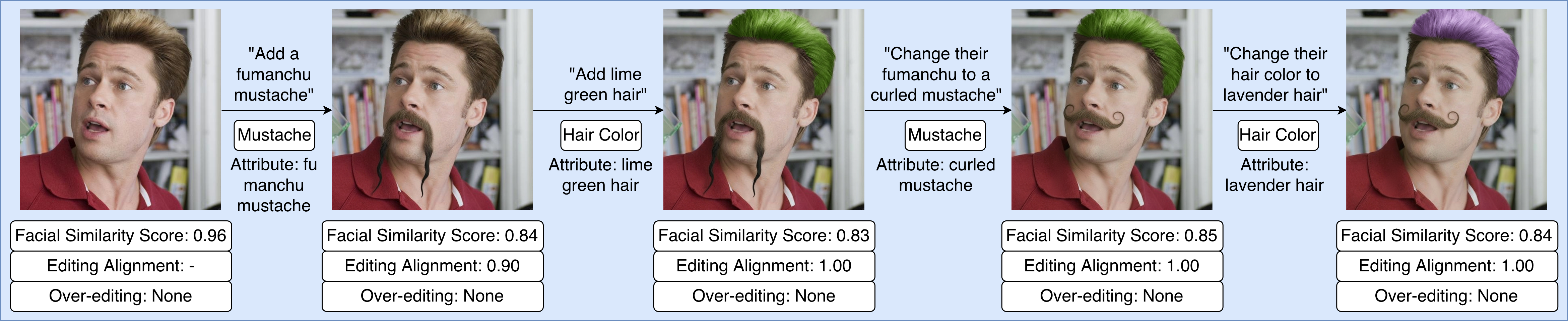}
  \vspace{-0.7cm}
  \caption{An example editing session from our study. Each image goes through a fixed sequence of $4$ edits.}
  \label{fig:teaser_fig}
  \vspace{-0.3cm}
\end{figure*}

\section{Introduction}
Facial appearance editing is an important computer vision task that powers popular applications such as FaceApp \cite{faceapp} and Photoshop \cite{photoshop}. The goal is to perform controlled edits on a person's face (e.g., changing a person's head pose) without altering additional elements in the image (e.g., changing hair color, background objects) or modifying the person's identity. Facial editing has widespread applications, such as virtual try-ons (e.g., adding glasses, hats, and new hair colors) and content creation. 

In recent years, Generative Adversarial Networks (GANs) \cite{choi2018stargan, chu2020sscgan, kwak2020cafe, zhou2022tigan, liu2019stgan} and 3D-morphable models (3DMMs) \cite{ding2023diffusionrig, hong2022headnerf, jang2025controlface, khan2025instaface} have been widely used for facial appearance editing. Both models have their pros and cons. While GANs can perform varied edits (e.g., changing hair color, hairstyle), they often produce unstable edits. While 3DMMs produce stable edits, they can only edit head pose and facial expression. This is because 3DMMs rely on 3D facial renderings where only pose and expression can be controlled. 

Recently, text-guided image editing diffusion models like Nano Banana \cite{nano}, which allow users to edit images using prompts (e.g., ``make the person look leftward''), have become popular for image editing tasks. Text-guided models provide a compelling alternative for facial editing to GANs and 3DMMs. Text-guided models are trained on a massive corpus of image-and-text pairs, enabling them to perform a much larger variety of facial edits compared to GANs (e.g., adding different facial hair styles, different hat styles). Text-guided models also produce consistent and photorealistic edits due to their use of diffusion models. 

There are several key requirements when using text-guided models for real-world facial editing applications: $(1)$ Models should be able to execute granular facial edits (e.g., adding specific facial hair styles like a handlebar mustache or hat types like a slouchy beanie hat). $(2)$ Models should perform edits fairly on male and female faces (no gender bias), on light and dark-skinned faces (no skin-tone bias), and on young and old faces (no age bias). $(3)$ Models should not overedit --- for instance, if a model is prompted to edit a person's hair color, it should not change their hairstyle.

Previous studies comprehensively evaluate text-guided models on their ability to execute generic whole-scene edits (e.g., making a person ride a dragon) \cite{zhang2023magicbrush, wei2024omniedit, brooks2023instructpix2pix, hui2024hq, zhao2024ultraedit, pathiraja2025refedit}. However, previous studies have not comprehensively evaluated text-guided models' ability to perform granular facial editing tasks. They also have not examined bias in these models across demographic groups. 

\begin{figure}[t]
  \centering
  \vspace{0.2cm}
  \includegraphics[width=\linewidth]{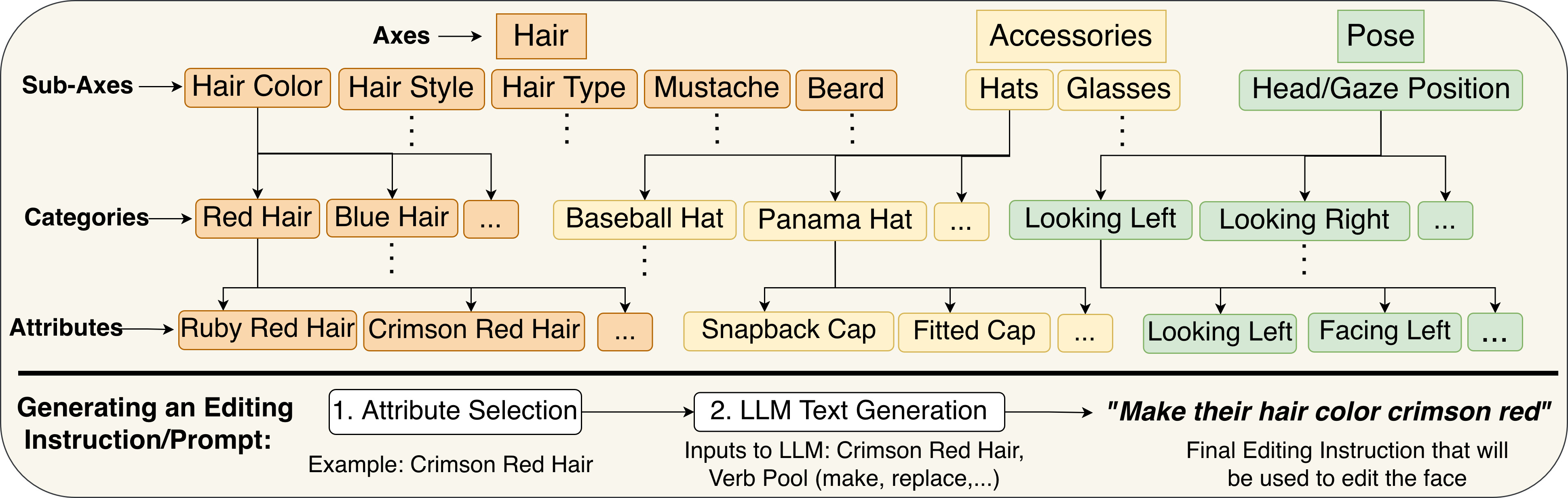}
  \vspace{-0.5cm}
  \caption{Face-Edit-Attributes: Leaf nodes (e.g., crimson red hair) are used to create the editing instructions.}
  \label{fig:ontology}
  \vspace{-0.7cm}
\end{figure}

We present the first large-scale evaluation of six state-of-the-art text-guided image editing models (Flux-Dev, Nano Banana, Bagel-Edit, Flux-Kontext-Pro, Qwen-Edit, and SeedEdit 3.0) for the facial appearance editing task. To perform facial edits, we propose \textbf{Face-Edit-Attributes} (Figure \ref{fig:ontology}), a systematic collection of $169$ manually curated facial editing attributes that focus on three types of edits: Hair (changing hair color, hair style, and adding facial hair), Accessories (adding different types of hats, glasses), and Pose/Gaze (making the person look left, right). Face-Edit-Attributes is an order of magnitude larger than prior facial editing datasets (e.g., CelebA-dialog \cite{jiang2021talk}, CHATEDIT \cite{cui2023chatedit}). 

We evaluated text-guided models on two popular celebrity face datasets: CelebA \cite{liu2015deep} and CelebSET \cite{raji2020saving}. To mimic practical facial editing use cases (e.g., retouching, virtual try-ons), we focused on multi-turn editing, where an image is edited using a sequence of editing instructions (Figure \ref{fig:teaser_fig}). For each image, we applied four editing instructions. We created separate evaluation sets for each edit type: Hair-Test, Accessories-Test, and Pose-Test. We also created a Multi-Axis-Test set, which is a mixture of these edit types. On each of these test sets, we compared overall model performance. We also evaluated whether models showed gender, skin-tone, and age bias.

We used $3$ metrics for evaluation: \textbf{$(1)$} facial similarity score (FSS), which evaluates how well a person's facial identity is preserved, \textbf{$(2)$} semantic consistency ($SC$), which evaluates whether the model executed the requested edit, and \textbf{$(3)$} we propose a new metric called the \textbf{overediting matrix}, which measures how many and which additional edits a model performed beyond the requested edit.

Our results show that most models correctly execute edits (high $SC$), while preserving $>70\%$ of a person's facial identity across the edit sequence (high FSS). However, performance is not the same across test sets --- most models executed hair and accessory-based edits well, but they struggled with pose edits. All models overedit --- for every $100$ editing instructions (25 sequences), models created $25$ overedits on average. The most common overedit patterns were associated with hair-edits --- for instance, several models changed hair color when they were asked to edit the person's hairstyle. Such overedit patterns are indicative of spurious correlations learned by models that future research should seek to mitigate.

In identity preservation (FSS) and in executing edits (SC), only Qwen-Edit and Flux-Dev showed consistent gender and skin-tone bias across test sets. We observed age bias in identity preservation --- several models better preserved the facial identity of young faces. We also observed bias in overediting performance --- all models created significantly more overedits when editing men (especially dark-skinned men), and when editing old people.

In summary, this paper makes the following contributions: $(1)$ We conducted the first large-scale evaluation of text-guided editing models (around $1$M images evaluated). $(2)$ We propose Face-Edit-Attributes, the largest collection of $169$ attributes to perform facial editing. $(3)$ We systematically evaluated four types of facial edits: hair, accessories, pose, and multi-axis edits. $(4)$ We propose a new metric, the Overediting Matrix, to count overedits. $(5)$ We conduct the first study of demographic biases (gender, skin-tone, and age bias) in facial editing, and show that models create more overedits when they edit male faces and old faces.

\paragraph{Summary of Related Work}: Current benchmarks for facial editing (e.g., MM-CelebA-HQ \cite{xia2021tedigan}, CHAT-EDIT \cite{cui2023chatedit}, CelebA-dialog \cite{jiang2021talk}) have only $15-35$ useful editing attributes. 

Well-known image editing benchmarks (e.g., Magic Brush \cite{zhang2023magicbrush}, Instruct-Pix-2-Pix \cite{brooks2023instructpix2pix}, Omni-Edit \cite{wei2024omniedit}) have evaluated text-guided models on synthetic images (generated by models like DALL-E 3 \cite{hurst2024gpt} and Stable Diffusion \cite{rombach2022high}), on scene datasets like COCO \cite{lin2014microsoft} and RefCOCO \cite{yu2016modeling}, or on online image repositories like Pexels \cite{pexels} and Flickr \cite{young2014image}. None of these benchmarks used face datasets (e.g., CelebA \cite{liu2015deep}, CelebA-HQ \cite{karras2019style}) for evaluation.  

Many metrics have been proposed to measure overediting in models \cite{ku2024viescore, li2025balancing, baraldi2025changed, liu2026human, qian2025gie}. All of these metrics report overediting with a numerical score. This does not provide information on what overedits occurred in the image. We can modify DICE \cite{baraldi2025changed} (one of these metrics) to detect what overedits occurred. However, DICE cannot reliably identify fine-grained overedits (e.g., did the model overedit a hat or a Panama hat?).

For faces, most studies have examined demographic biases in tasks such as facial recognition, classification, and generation \cite{buolamwini2018gender, cavazos2020accuracy, cascone2025framework, leyva2024demographic}. To our knowledge, no work has extensively evaluated demographic biases in facial editing. For facial editing, only stereotype-based biases have been studied (e.g., past work showed that if we edit an image of a woman with a prompt like ``present the person as a senior executive'', models often edit the woman into a man) \cite{seo2026evaluating}. We discussed related works in detail in Section \ref{sec:app_rel_work}.

\section{Evaluation Setup}

\subsection{Problem Statement}
\label{subsec:problem_setup}
We focus on multi-turn editing, where a model iteratively edits an image with prompts (or instructions) in sequence. We begin with an input image $I_{0}$ (also called the base image) and a prompt $P_{1}$, which is called an editing instruction. A text-guided image editing model $M$ takes the prompt $P_{1}$ and edits the base image $I_{0}$ to produce an edited image $I_{1}$; $I_1 = M(I_0)$. Next, the model takes the edited image $I_{1}$ and applies a new editing instruction $P_{2}$ to obtain $I_{2}$. This process continues for $k$ editing instructions, resulting in the final edited image $I_{K}$. This iterative sequence of edits from $I_{0}$ to $I_{K}$ is called an edit session. In this work, $K = 4$ (chosen based on computational constraints). Figure \ref{fig:teaser_fig} shows an example edit session. 

\subsection{Face Datasets (Base Images)}
\label{subsec:face_datasets}
To obtain the base images ($I_{0}$), we used two popular celebrity face datasets: CelebA: a dataset with over $200k$ images covering $10k$ celebrities \cite{liu2015deep} and CelebSET \cite{raji2020saving}: over $12$k images covering $80$ celebrities (we used face images from CelebSET's source dataset: IMDB-WIKI \cite{rothe2015dex}). In both CelebA and CelebSET, each image contains a headshot of a celebrity. We removed images with multiple faces. CelebA provides binary gender (male, female) and age annotations (young, old) for each celebrity. CelebSET provides annotations for gender (male, female), skin tone (light, dark), and age (numerical value). Using age values, we created two categories: young (faces in the age range $20 - 30$) and old (faces in the age range $50 - 60$). We only considered these age ranges to obtain two distinct age groups.  

Due to the computational cost of text-guided models, we limited our analysis to $\sim500$ images per dataset. To filter images from each dataset, we used a facial similarity score (FSS) threshold (refer to Section \ref{subsec:id_preserve} for the FSS metric). We used FSS $>0.95$ for CelebA and $>0.85$ for CelebSET. This resulted in $549$ images for CelebA and $517$ images for CelebSET.

CelebA data statistics for gender: $244$ images from $208$ males and $305$ images from $280$ females. Age statistics: $413$ images from $375$ young people and $136$ images from $116$ old people. 

CelebSET data statistics for gender-race: $222$ images from $20$ light-skinned males and $95$ images from $16$ dark-skinned males; $161$ images from $17$ light-skinned females and $39$ images from $15$ dark-skinned females. Age statistics: $98$ images from $28$ young people (aged $20 - 30$) and $41$ images from $8$ old people (aged $50 - 60$). 

\subsection{Editing Attributes}
\label{subsec:fec}

To generate editing instructions, we manually curated Face-Edit-Attributes, a dataset of $169$ facial editing attributes (e.g., stetson hat). We designed a hierarchical approach to curate this dataset (Figure \ref{fig:ontology}). At level $1$, we defined three broad editing types (or axes): Hair, Accessories, and Pose. Each axis is divided into sub-axes (level $2$). For example, in the axis ``Hair'', we defined the following sub-axes: hair style, hair color, hair type, mustache, and beard.

Each sub-axis has multiple categories (level $3$). For instance, within the sub-axis ``hair style'', we have categories like buzz cut, bob cut, and spiked cut. Finally, each category has multiple editing attributes (level $4$) closely related to the category. For instance, within the category ``spiked cut'', we have editing attributes like spiked hair, liberty spiked hair, mohawk hair, and gelled spiked hair. The editing attributes are used to create the editing instructions. Refer to Section \ref{app:face_edit_attributes} for a complete list of editing attributes. 

This hierarchical design has several advantages: (1) We can create a coarse-grained test set (e.g., Hair-Test), as well as a fine-grained test set (e.g., Mustache-Test). (2) This design forms the basis of our new overediting metric, which we discuss in Section \ref{subsec:overediting}.

We chose three axes (hair, accessories, and pose) to purely focus on the facial region. For the hair axis, we chose sub-axes focused on scalp and facial hair. For accessories, we chose glasses and hats as our sub-axes. We ignored non-facial sub-axes like hand or neck accessories. We did not consider earrings: if a woman's ears are occluded in the base image, adding an earring would require the model to generate an ear (or estimate ear position). This adds a layer of complexity to the model that can impact editing results. For pose, we used a single sub-axis focused on head/gaze movements.

In each sub-axis, we chose categories (level 3) with a commonly accepted shape and appearance definition. We excluded several categories that have a fluid shape definition. Consider the glass sub-axis. Glasses have several popular categories: aviators, cat-eyed, geometric, and browline. In most cases, these categories look similar, as each glass brand follows its own definition. This affects editing alignment, since we cannot be sure if the model executed the edit correctly. For the head/gaze sub-axis, we focused on three categories: looking left, right, and straight ahead. We excluded looking up and down movements as this can impact identity preservation, which is integral to our study. 

In each category, we chose attributes (level $4$) that encourage semantic and linguistic diversity in editing instructions. We introduced semantic diversity through synonyms (e.g., snapback cap for baseball hat). For linguistic diversity, we used alternative phrasings (e.g., looking ahead, frontal view for ``looking straight ahead'').

There are other editing attributes relevant to facial edits that we did not consider. We excluded attributes that modify the face (e.g., jawline and nose modifications) as this affects identity preservation. We excluded subtle edits such as facial expressions and makeup. Facial expressions are subtle and hard to evaluate (e.g., it is hard to distinguish a surprised reaction from a sad one). Makeup edits are hard to identify as the image hue can obscure makeup shades. 

\subsection{Editing Instructions and Sessions}
\label{subsec:edit_prompt_dataset}
To generate an editing instruction, we selected an attribute from the Face-Edit-Attributes dataset. For instance, to evaluate hair-based edits, we randomly select an attribute (say crimson red hair) from the Hair axis. We then prompt a multimodal large language model (MLLM) to generate an editing instruction (e.g., ``make the person's hair color crimson red'') by combining the attribute with an appropriate verb. We used Gemini-2.5-Flash \cite{gemini-flash} as our MLLM.

To create an edit session, the MLLM generates $4$ instructions in sequence (since each session has $4$ edits). After generating the first instruction, $P_1$ (e.g., ``make the person's hair color crimson red''), we provided the MLLM with metadata about $P_1$ (attribute: crimson red hair, sub-axis: hair-color, axis: hair) to give it context to generate the next instruction $P_{2}$. We provided our prompt in Section \ref{app:mllm_prompt}.

We created four evaluation sets: one set for each axis (Hair-Test, Accessories-Test, Pose-Test) and an overall set (Multi-Axis-Test). To create the evaluation set for a specific axis, we randomly selected editing attributes only from that axis. For the Multi-Axis-Test set, we randomly selected attributes from the complete set of $169$ attributes. For Hair-Test, Accessories-Test, and Multi-Axis-Test, we generated $10$ edit sessions per test set. For Pose-test, we generated $5$ sessions (since poses have only three variations: look left, look straight, and look right). This resulted in a total of $35$ edit sessions or $140$ editing instructions ($4$ instructions per session). Refer to Section \ref{app:edit_instructions} for the list of edit sessions in each test set.



We generated simple, one-line editing instructions consistent with the instruction template used by popular general scene editing benchmarks~\cite{zhang2023magicbrush, wei2024omniedit, brooks2023instructpix2pix}. These instructions test the effectiveness of text-guided models for real-world facial editing applications and enable larger-scale evaluations than detailed editing instructions.

\subsection{Evaluation Metrics}
\label{subsec:eval_metrics}

\subsubsection{Identity preservation} 
\label{subsec:id_preserve}

We verified whether the facial identity of a person is preserved across the edit session using a Facial Similarity Score (FSS), which ranges from $0$ to $1$. To compute FSS, we used Buffalo-Large \cite{buffalo}, a popular face recognition model, which uses RetinaFace \cite{deng2020retinaface} to detect a face in the image, and a ResNet-50 \cite{he2016deep} network to generate an embedding vector for the detected face. 

\begin{figure}[t]
  \centering
  \includegraphics[width=1.00\linewidth]{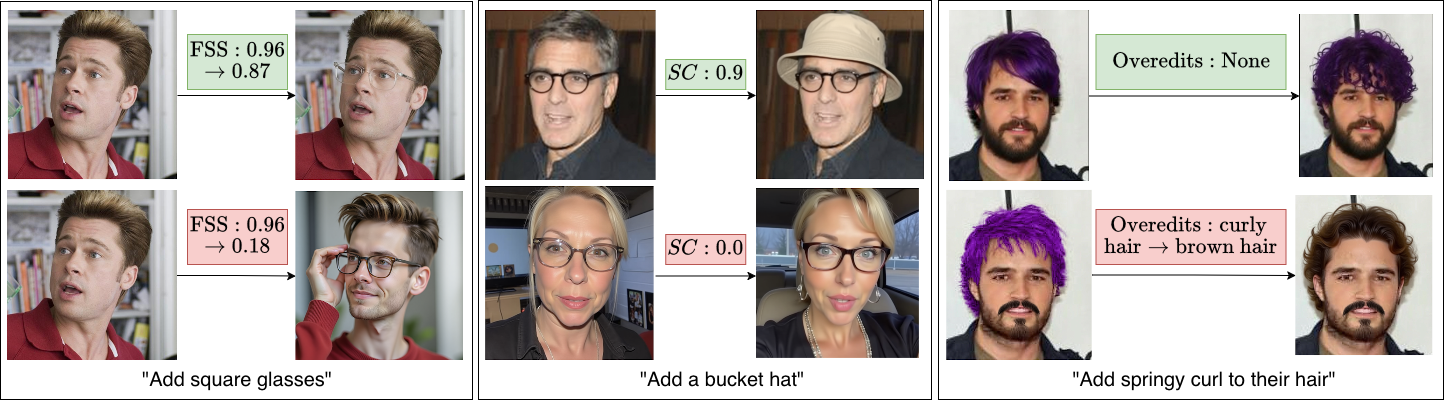}
  \vspace{-0.75cm}
  \caption{Visually comparing good and bad scenarios for our evaluation metrics. \textit{Left:} the top row (Nano-Banana) shows good identity preservation, while the bottom row (Flux-Dev) shows poor identity preservation. \textit{Center:} the top row (Nano Banana) shows a correctly executed edit, while the bottom row (Flux-Dev) shows a poorly executed edit. \textit{Right:} the top row (Nano Banana) shows a case for no overedits, while the bottom row (Qwen-Edit) indicates an overedit.}
  \label{fig:metric_comparisons}
  \vspace{-0.5cm}
\end{figure}

Assume we have an image $I_k$ ($k$ is the editing step) of a celebrity $C$. We computed an embedding vector $f(I_k)$. For celebrity $C$, we loaded a set of reference images $R_C$ from our reference database (the reference database has images for each celebrity, which are extracted from CelebA and CelebSET; they are not part of the base images we used for editing). We computed an embedding vector $f(r)$ for each reference image $r \in R_C$. We computed cosine similarity between $f(I_k)$ and each $f(r)$ from the reference set. $FSS_{k}$ is the maximum of these similarity scores. We computed FSS in each edit session, from $I_0$ to the final edited image $I_k$. An FSS$_k$ close to $1$ indicates that the person in $I_k$ closely resembles celebrity $C$. Figure \ref{fig:metric_comparisons} (left) shows examples of edits with high and low FSS. 

To compare model performance in an edit session, we also propose $\operatorname{Drop}_{\mathrm{FSS}}$, which is an extension to FSS. In $\operatorname{Drop}_{\mathrm{FSS}}$, we calculated the average $\%$ drop in FSS of a model across an edit session.


\vspace{-0.3cm}
\[
\operatorname{Drop}_{\mathrm{FSS}}
= \frac{1}{K}\sum_{k=1}^{K}\left(\frac{\mathrm{FSS}_0 - \mathrm{FSS}_k}{\mathrm{FSS}_0}\right) \times 100 \%.
\]
\vspace{-0.2cm}

$K = 4$ because we have four edits per session. $FSS_0$ is the facial similarity score for $I_0$ (base image), and $FSS_k$ is the score for $I_k$. A $\operatorname{Drop}_{\mathrm{FSS}}$ of $40\%$ indicates that across an edit session, the FSS dropped by $40\%$ (on average) with respect to the base image. A lower $\operatorname{Drop}_{\mathrm{FSS}}$ indicates better identity preservation. 


\subsubsection{Editing Alignment} 

To measure how well a model followed the editing instructions, we used Semantic Consistency ($SC$), a popular MLLM-based metric proposed by \cite{ku2024viescore} to measure alignment in scene editing benchmarks. The MLLM (we used Gemini-2.5-Flash) outputs two scores: $(1)$ editing success:  $0$ indicates the editing instruction was not executed at all, while $10$ indicates perfect execution, $(2)$ overediting degree: $0$ indicates that the image was overedited (e.g., unwanted background changes), and $10$ indicates that the image showed minimal changes. For a strict scoring, we reported $SC$ as the minimum of these two scores. We normalized $SC$ in the range of $0 - 1$. We computed $SC$ on images in $[I_1, I_k]$, as $SC$ can only be computed for an edited image. Refer to Figure \ref{fig:metric_comparisons} (center) for examples of edits with high and low $SC$. 

The overediting score in SC has limitations: $(1)$ In many cases, it does not capture overedits correctly. For example, in Figure \ref{fig:metric_comparisons}, the bottom row of the rightmost grid shows an overedit: the model changed hair color when it was prompted to change hair style. In this example, $SC = 0.8$, which means that both the ``editing success'' and ``overediting degree'' scores were high. While editing success should be high, the overediting score should be low. $(2)$ It cannot measure how many overedits occurred in an image and what these overedits were (this is critical to assess spurious correlations). To provide a more accurate and detailed measure of overediting, we propose a new metric called the Overediting Matrix.

\subsubsection{Overediting Matrix} 
\label{subsec:overediting}

To count overedits in a model, we propose an Overediting Matrix. This is a $35 \times 35$ matrix where the rows and columns represent $35$ categories of the Face-Edit-Attributes dataset (level 3). Each row represents the category that the user intended to edit, and each column represents the category that the model edited in the image. For example, if the matrix entry [red hair][stubble] equals $645$, it means that in $645$ instances, the prompt requested a change in hair color to red, but the model added stubble.

To create this matrix, we used a state dictionary. Assume we have a base image ($I_0$) of a person with black-colored straight hair, no facial hair, looking ahead. For $I_0$, the state dictionary will be: \{hair-color: black, beard: clean shaven, hair-type: straight, head position: looking forward, hair-style: None, mustache: None, hats: None, glasses: None\}. The keys in this dictionary are the $8$ sub-axes from the Face-Edit-Attributes dataset. The values are the categories in these sub-axes (e.g., red hair, blue hair). 

Suppose a model edited $I_0$ using the instruction $P_{1}$: ``make the person's hair color crimson red''. Assume the model changed the hair color to crimson red, but also added stubble (overedit). The updated dictionary would be: \{hair-color: black $\rightarrow$ red, {beard: clean shaven $\rightarrow$ stubble}, ... \} (other sub-axes stay unchanged). The hair color changing from black $\rightarrow$ red is an intended edit. However, beard changing from clean shaven $\rightarrow$ stubble is an overedit. In our overediting matrix, we increment $[\text{red hair}][\text{stubble}]$ by $1$. 

To track the state dictionary after each edit, we used Gemini-2.5-Flash. Given an image $I_0$, the MLLM predicts the most probable category for each sub-axis. It returns None for sub-axes not present in the image. If the MLLM selects multiple top categories for a sub-axis (e.g., black hair, brown hair), we record all of them.

We selected Gemini-2.5-Flash as our state tracker as it showed the highest alignment with human annotators. We compared state dictionaries generated by four models (Gemini-2.5-Flash \cite{gemini-flash}, Gemini-2.5-Pro \cite{gemini-pro}, GPT-4o \cite{hurst2024gpt}, and  CLIP \cite{radford2021learning}) against the state dictionary created by human annotators. Alignment accuracies: $79.65\%$ for Gemini-2.5-Flash, $79.26\%$ for Gemini-2.5-Pro, $70.31\%$ for GPT-4o, and $35.47\%$ for CLIP. Refer to Section \ref{app:human_exp} for details.

We tracked coarse-grained categories rather than the $169$ fine-grained attributes, as MLLMs more reliably distinguish categories (e.g., red vs.\ blue hair) than closely related attributes (e.g., crimson red vs.\ ruby red hair). Coarse-grained categories are sufficient for counting overedits.

To compare the amount of overediting across models, we also computed an overediting rate (OER) using the state dictionary. In a session (of four edits), if a model performs a total of two overedits, the OER of that session is $2/4$ or $0.5$. 

\subsubsection{Wasserstein Distance}
\label{subsec:wd}

To quantify performance disparities across demographic groups, we used Wasserstein Distance (WD), a popular fairness metric \cite{miroshnikov2022wasserstein}. WD measures the distance between two distributions. Consider two distributions: ($1$) $\operatorname{Drop}_{\mathrm{FSS}}$ values for male faces, and $(2)$ $\operatorname{Drop}_{\mathrm{FSS}}$ for female faces. If $WD = 2$, this indicates that the male and female $\operatorname{Drop}_{\mathrm{FSS}}$ values are $2\%$ apart. WD does not give the direction of bias (here, we do not know if $\operatorname{Drop}_{\mathrm{FSS}}$ is higher for males or females). To get the direction of bias, we must examine the mean values of the demographic groups. WD ranges from $0$ to the largest value of the chosen metric. 

For each of the following metrics: $\operatorname{Drop}_{\mathrm{FSS}}$, $SC$, and OER, we reported three WD scores: $\operatorname{WD}_{\mathrm{g}}$ (gender disparity between male and female groups), $\operatorname{WD}_{\mathrm{s}}$ (skin-tone disparity between light and dark groups), and $\operatorname{WD}_{\mathrm{age}}$ (age disparity between young and old groups). To assess whether WD values are statistically significant, we reported the p-values. Our null hypothesis was that two groups (e.g., male and female) are drawn from the same distribution (no disparity). We tested our null hypothesis using the standard Monte Carlo permutation test. If p $< 0.05$, we reject the null hypothesis and say that the WD score is statistically significant.



\subsection{Editing Models}
We evaluated five popular text-guided image editing models on CelebA: Flux.1-Dev (Flux-Dev) \cite{flux-dev}, Flux.1 Kontext-Pro (Kontext-Pro) \cite{labs2025flux}, Gemini-2.5-Flash-Image (Nano-Banana) \cite{nano}, Bagel-Edit \cite{deng2025emerging}, and Qwen-Edit \cite{wu2025qwen}. On CelebSET, we also evaluated SeedEdit 3.0 \cite{wang2025seededit}. We limited SeedEdit 3.0 to a single dataset due to cost constraints. If a model did not generate all four edited images in an edit session, we excluded that session. Across all models, we evaluated about $200$K sessions ($\sim1$M images) across both datasets. Refer to Section \ref{app:edit_sessions} for the number of edit sessions per model.

\section{Results}

\begin{table}[t]
\centering
\caption{$\operatorname{Drop}_{\mathrm{FSS}}$ (lower is better) across all test sets on CelebA and CelebSET. For each model, \textcolor{blue}{blue} number indicates its best performance, while \textcolor{red}{red} indicates its worst performance across the four test sets. The Avg. column shows the average performance of a model across test sets. The bolded number indicates the overall best-performing model. Subscripts denote $95\%$ confidence intervals.}
\vspace{-0.4cm}
\label{tab:face_id_drop_overall}

\setlength{\tabcolsep}{2.6pt}
\renewcommand{\arraystretch}{0.95}

\noindent
\begin{minipage}[t]{0.485\textwidth}
\centering
{\small \textbf{CelebA}} 

\vspace{0.1em}
\small
\begin{tabular}{lccccc}
\toprule
Model &
Hair$\downarrow$ &
Acc.$\downarrow$ &
Pose$\downarrow$ &
Multi$\downarrow$ &
Avg.$\downarrow$ \\
\midrule

Kontext-Pro &
\textcolor{blue}{24.60$_{0.16}$} &
25.37$_{0.14}$ &
\textcolor{red}{37.48$_{0.28}$} &
32.57$_{0.17}$ &
\textbf{28.68}$_{0.10}$ \\

Nano-Banana &
\textcolor{blue}{23.08$_{0.12}$} &
\textcolor{red}{33.51$_{0.14}$} &
32.43$_{0.15}$ &
32.06$_{0.16}$ &
29.93$_{0.08}$ \\

Bagel-Edit &
\textcolor{blue}{18.79$_{0.09}$} &
27.37$_{0.10}$ &
\textcolor{red}{52.98$_{0.28}$} &
33.51$_{0.20}$ &
30.33$_{0.11}$ \\

Qwen-Edit &
28.04$_{0.20}$ &
\textcolor{blue}{27.90$_{0.14}$} &
\textcolor{red}{47.59$_{0.22}$} &
34.82$_{0.17}$ &
32.73$_{0.10}$ \\

Flux-Dev &
\textcolor{blue}{89.13$_{0.06}$} &
89.58$_{0.06}$ &
\textcolor{red}{89.68$_{0.09}$} &
89.40$_{0.06}$ &
89.41$_{0.04}$ \\

\bottomrule
\end{tabular}
\end{minipage}
\hfill
\begin{minipage}[t]{0.485\textwidth}
\centering
{\small \textbf{CelebSET}} 

\vspace{0.1em}
\small
\begin{tabular}{lccccc}
\toprule
Model &
Hair$\downarrow$ &
Acc.$\downarrow$ &
Pose$\downarrow$ &
Multi$\downarrow$ &
Avg.$\downarrow$ \\
\midrule

Kontext-Pro &
\textcolor{blue}{21.47$_{0.16}$} &
26.58$_{0.12}$ &
\textcolor{red}{29.39$_{0.30}$} &
29.32$_{0.18}$ &
26.21$_{0.09}$ \\

Nano-Banana &
\textcolor{blue}{16.56$_{0.10}$} &
\textcolor{red}{25.97$_{0.10}$} &
23.02$_{0.16}$ &
23.85$_{0.14}$ &
\textbf{22.25}$_{0.06}$ \\

Bagel-Edit &
\textcolor{blue}{18.20$_{0.10}$} &
27.73$_{0.09}$ &
\textcolor{red}{56.25$_{0.29}$} &
33.79$_{0.22}$ &
30.81$_{0.12}$ \\

Qwen-Edit &
\textcolor{red}{46.56$_{0.24}$} &
46.19$_{0.22}$ &
\textcolor{blue}{39.40$_{0.23}$} &
44.40$_{0.23}$ &
44.82$_{0.12}$ \\

Flux-Dev &
\textcolor{blue}{83.53$_{0.07}$} &
\textcolor{red}{83.96$_{0.07}$} &
83.58$_{0.11}$ &
83.65$_{0.07}$ &
83.69$_{0.04}$ \\

SeedEdit 3.0 &
\textcolor{blue}{18.64$_{0.11}$} &
25.50$_{0.10}$ &
\textcolor{red}{45.54$_{0.20}$} &
31.68$_{0.14}$ &
28.15$_{0.09}$ \\

\bottomrule
\end{tabular}
\end{minipage}

\vspace{-0.3cm}
\end{table}

\begin{table}[t]
\centering
\caption{$SC$ (higher is better) across all test sets on CelebA and CelebSET. For each model, \textcolor{blue}{blue} number indicates its best performance, while \textcolor{red}{red} number indicates its worst performance across the four test sets. In the Avg. column, the bolded number indicates the best-performing model. Values scaled by $100$.}
\vspace{-0.5cm}
\label{tab:sc_overall}

\setlength{\tabcolsep}{2.6pt}
\renewcommand{\arraystretch}{0.95}

\noindent
\begin{minipage}[t]{0.485\textwidth}
\centering
{\small \textbf{CelebA}}

\vspace{0.2em}
\small
\begin{tabular}{lccccc}
\toprule
Model &
Hair$\uparrow$ &
Acc.$\uparrow$ &
Pose$\uparrow$ &
Multi$\uparrow$ &
Avg.$\uparrow$ \\
\midrule

Kontext-Pro &
\textcolor{blue}{76.10$_{0.50}$} &
73.40$_{0.60}$ &
\textcolor{red}{47.40$_{0.90}$} &
64.00$_{0.50}$ &
68.40$_{0.30}$ \\

Nano Banana &
\textcolor{blue}{87.90$_{0.30}$} &
84.70$_{0.40}$ &
\textcolor{red}{62.60$_{0.80}$} &
75.10$_{0.50}$ &
\textbf{79.60}$_{0.30}$ \\

Bagel-Edit &
80.30$_{0.40}$ &
\textcolor{blue}{87.80$_{0.30}$} &
\textcolor{red}{59.60$_{0.70}$} &
68.90$_{0.50}$ &
76.20$_{0.30}$ \\

Qwen-Edit &
76.60$_{0.50}$ &
\textcolor{blue}{79.90$_{0.40}$} &
67.50$_{0.80}$ &
\textcolor{red}{64.70$_{0.50}$} &
72.90$_{0.30}$ \\

Flux-Dev &
30.30$_{0.50}$ &
\textcolor{blue}{33.70$_{0.50}$} &
31.40$_{0.80}$ &
\textcolor{red}{30.20$_{0.50}$} &
31.40$_{0.30}$ \\

\bottomrule
\end{tabular}
\end{minipage}
\hfill
\begin{minipage}[t]{0.485\textwidth}
\centering
\vspace{0.05cm}
{\small \textbf{CelebSET}}

\vspace{0.2em}
\small
\begin{tabular}{lccccc}
\toprule
Model &
Hair$\uparrow$ &
Acc.$\uparrow$ &
Pose$\uparrow$ &
Multi$\uparrow$ &
Avg.$\uparrow$ \\
\midrule

Kontext-Pro &
\textcolor{blue}{75.60$_{0.40}$} &
73.10$_{0.60}$ &
\textcolor{red}{37.00$_{1.00}$} &
61.20$_{0.60}$ &
66.10$_{0.40}$ \\

Nano Banana &
\textcolor{blue}{85.90$_{0.30}$} &
82.30$_{0.40}$ &
\textcolor{red}{59.40$_{0.80}$} &
73.00$_{0.60}$ &
\textbf{77.40}$_{0.30}$ \\

Bagel-Edit &
78.30$_{0.40}$ &
\textcolor{blue}{86.80$_{0.40}$} &
\textcolor{red}{60.50$_{0.70}$} &
67.70$_{0.60}$ &
75.10$_{0.30}$ \\

Qwen-Edit &
78.90$_{0.40}$ &
\textcolor{blue}{82.50$_{0.40}$} &
\textcolor{red}{63.00$_{0.80}$} &
66.30$_{0.50}$ &
74.00$_{0.30}$ \\

Flux-Dev &
28.80$_{0.50}$ &
28.70$_{0.50}$ &
29.30$_{0.80}$ &
28.20$_{0.50}$ &
28.70$_{0.30}$ \\

SeedEdit 3.0 &
82.10$_{0.40}$ &
\textcolor{blue}{86.30$_{0.40}$} &
\textcolor{red}{63.80$_{0.80}$} &
68.90$_{0.50}$ &
77.00$_{0.30}$ \\

\bottomrule
\end{tabular}
\end{minipage}

\vspace{-0.3cm}
\end{table}

\begin{table}[t]
\centering
\caption{Frequent overediting patterns across models. Nano-Banana (NB), SeedEdit (SE), Bagel-Edit (BE), Qwen-Edit (QE), Flux-Dev (F-Dev), Kontext-Pro (K-Pro).}
\vspace{-0.4cm}
\label{tab:overedit_patterns}
\setlength{\tabcolsep}{4pt}
\renewcommand{\arraystretch}{1.05}
\footnotesize

\begin{minipage}[t]{0.48\textwidth}
\centering
\textbf{CelebA}\par\vspace{0.3em}
\begin{tabular}{@{} l p{1.5cm} @{}}
\toprule
\textbf{Intended Edit $\rightarrow$ Overedit} & \textbf{Models} \\
\midrule
bob cut $\rightarrow$ straight hair & BE, K-Pro, QE \\
sideburn mutton chops $\rightarrow$ handlebar mustache & BE, K-Pro, QE \\
bob cut $\rightarrow$ brown hair & F-Dev, QE \\
sideburn mutton chops $\rightarrow$ pencil mustache & K-Pro, QE \\
straight hair $\rightarrow$ brown hair & BE, QE \\
\bottomrule
\end{tabular}
\end{minipage}
\hfill
\begin{minipage}[t]{0.48\textwidth}
\centering
\vspace{0.05cm}
\textbf{CelebSET}\par\vspace{0.3em}
\begin{tabular}{@{} l p{2.0cm} @{}}
\toprule
\textbf{Intended Edit $\rightarrow$ Overedit} & \textbf{Models} \\
\midrule
sideburn mutton chops $\rightarrow$ handlebar mustache & BE, K-Pro, QE, SE \\
bob cut $\rightarrow$ brown hair & F-Dev, K-Pro, QE \\
curly hair $\rightarrow$ None & BE, NB, SE \\
bob cut $\rightarrow$ straight hair & K-Pro, QE \\
curly hair $\rightarrow$ brown hair & F-Dev, QE \\
\bottomrule
\end{tabular}
\end{minipage}
\vspace{-0.4cm}
\end{table}

\begin{table}[t]
\centering
\caption{OER (lower is better) across all test sets on CelebA and CelebSet. For each model, \textcolor{blue}{blue} number indicates its best performance, while \textcolor{red}{red} number indicates its worst performance across the test sets. The bolded number indicates the best-performing model. Values scaled by $100$.}
\vspace{-0.3cm}
\label{tab:oer_overall}

\setlength{\tabcolsep}{2.6pt}
\renewcommand{\arraystretch}{0.95}

\noindent
\begin{minipage}[t]{0.485\textwidth}
\centering
{\small \textbf{CelebA}}

\vspace{0.2em}
\small
\begin{tabular}{lccccc}
\toprule
Model &
Hair$\downarrow$ &
Acc.$\downarrow$ &
Pose$\downarrow$ &
Multi$\downarrow$ &
Avg.$\downarrow$ \\
\midrule

Kontext-Pro &
19.54$_{0.67}$ &
\textcolor{blue}{11.38$_{0.64}$} &
16.47$_{1.10}$ &
\textcolor{red}{22.83$_{0.77}$} &
17.75$_{0.39}$ \\

Nano Banana &
15.02$_{0.59}$ &
10.42$_{0.71}$ &
\textcolor{blue}{10.18$_{0.83}$} &
\textcolor{red}{17.62$_{0.69}$} &
\textbf{13.79}$_{0.35}$ \\

Bagel-Edit &
18.47$_{0.64}$ &
\textcolor{blue}{8.48$_{0.56}$} &
13.82$_{0.95}$ &
\textcolor{red}{21.03$_{0.73}$} &
15.68$_{0.35}$ \\

Qwen-Edit &
24.87$_{0.76}$ &
\textcolor{blue}{13.66$_{0.74}$} &
19.87$_{1.05}$ &
\textcolor{red}{27.25$_{0.90}$} &
21.63$_{0.43}$ \\

Flux-Dev &
66.42$_{1.33}$ &
\textcolor{red}{68.38$_{1.31}$} &
\textcolor{blue}{47.76$_{1.58}$} &
65.31$_{1.29}$ &
64.00$_{0.70}$ \\

\bottomrule
\end{tabular}
\end{minipage}
\hfill
\begin{minipage}[t]{0.485\textwidth}
\centering
\vspace{0.05cm}
{\small \textbf{CelebSET}}

\vspace{0.2em}
\small
\begin{tabular}{lccccc}
\toprule
Model &
Hair$\downarrow$ &
Acc.$\downarrow$ &
Pose$\downarrow$ &
Multi$\downarrow$ &
Avg.$\downarrow$ \\
\midrule

Kontext-Pro &
23.53$_{0.74}$ &
15.63$_{0.81}$ &
\textcolor{blue}{14.95$_{1.14}$} &
\textcolor{red}{25.90$_{0.84}$} &
20.90$_{0.44}$ \\

Nano Banana &
16.99$_{0.63}$ &
21.72$_{1.41}$ &
\textcolor{blue}{9.71$_{0.81}$} &
\textcolor{red}{24.15$_{1.05}$} &
19.34$_{0.57}$ \\

Bagel-Edit &
24.10$_{0.88}$ &
\textcolor{blue}{12.77$_{0.99}$} &
13.81$_{0.94}$ &
\textcolor{red}{28.36$_{1.08}$} &
20.61$_{0.51}$ \\

Qwen-Edit &
32.08$_{0.84}$ &
22.88$_{0.92}$ &
\textcolor{blue}{18.64$_{1.05}$} &
\textcolor{red}{33.17$_{0.97}$} &
27.85$_{0.49}$ \\

Flux-Dev &
72.23$_{1.40}$ &
68.07$_{1.37}$ &
\textcolor{blue}{52.13$_{1.73}$} &
\textcolor{red}{72.95$_{1.46}$} &
68.38$_{0.75}$ \\

SeedEdit 3.0 &
18.84$_{0.69}$ &
\textcolor{blue}{9.29$_{0.62}$} &
18.34$_{1.08}$ &
\textcolor{red}{25.12$_{1.04}$} &
\textbf{17.80}$_{0.43}$ \\

\bottomrule
\end{tabular}
\end{minipage}

\vspace{-0.4cm}
\end{table}

\begin{table*}[t]
\centering
\caption{$\operatorname{Drop}_{\mathrm{FSS}}$ \% (lower is better) disparity on CelebA. For each model and test set, we reported mean $\operatorname{Drop}_{\mathrm{FSS}}$ for Male (M) and Female (F). In each mean column, the left value is the mean of the worst performing group, and the right value is the mean of the best group. $WD_g$ denotes the Wasserstein distance between the male and female $\operatorname{Drop}_{\mathrm{FSS}}$ distributions. A superscript $*$ marks non-significant values ($p > 0.05$). WD > 3 is highlighted as it indicates gender bias.}
\vspace{-0.35cm}
\label{tab:dropfss_celeba}

\setlength{\tabcolsep}{4.2pt}
\renewcommand{\arraystretch}{1.05}

\definecolor{wdlight}{RGB}{252,228,214}
\definecolor{wddark}{RGB}{244,177,131}

\newcommand{\wdlightcell}[1]{\cellcolor{wdlight}{#1}}
\newcommand{\wddarkcell}[1]{\cellcolor{wddark}{#1}}

\scriptsize
\begin{tabular}{lcccccccc}
\toprule
&
\multicolumn{2}{c}{Accessories$\downarrow$} &
\multicolumn{2}{c}{Hair$\downarrow$} &
\multicolumn{2}{c}{Pose$\downarrow$} &
\multicolumn{2}{c}{Multi-axis$\downarrow$} \\
\cmidrule(lr){2-3}
\cmidrule(lr){4-5}
\cmidrule(lr){6-7}
\cmidrule(lr){8-9}
Model &
Mean & $WD_g$ &
Mean & $WD_g$ &
Mean & $WD_g$ &
Mean & $WD_g$ \\
\midrule

Flux-Dev &
F:89.7 \quad M:89.4 & 0.4 &
M:89.2 \quad F:89.1 & 0.4 &
F:90.0 \quad M:89.2 & 0.8 &
F:89.5 \quad M:89.3 & 0.5 \\

Kontext-Pro &
M:25.8 \quad F:25.0 & 1.6 &
F:24.7 \quad M:24.4 & 1.2 &
F:39.1 \quad M:35.5 & \textbf{3.4} &
F:33.6 \quad M:31.2 & 2.4 \\

Nano Banana &
M:34.5 \quad F:32.8 & 1.8 &
M:24.3 \quad F:22.1 & 2.3 &
M:33.0 \quad F:32.0 & 1.0 &
M:33.0 \quad F:31.3 & 2.2 \\

Bagel-Edit &
M:28.1 \quad F:26.8 & 1.4 &
M:19.5 \quad F:18.3 & 1.2 &
F:55.2 \quad M:50.2 & \textbf{5.0} &
F:33.8 \quad M:33.1 & 1.4 \\

Qwen-Edit &
M:29.1 \quad F:27.0 & 2.5 &
F:28.8 \quad M:27.1 & 2.0 &
F:50.8 \quad M:43.6 & \textbf{7.3} &
F:35.3 \quad M:34.2 & 1.8 \\

\bottomrule
\end{tabular}

\vspace{-0.25cm}
\end{table*}

\begin{table*}[t]
\centering
\caption{$\operatorname{Drop}_{\mathrm{FSS}}$ \% (lower is better)  disparity on CelebSET. We reported the mean $\operatorname{Drop}_{\mathrm{FSS}}$ for the two worst intersectional groups. In each mean column, left value: worst; right value: second-worst mean group. $WD_g$ and $WD_s$ denote Wasserstein distances between gender and skin-tone groups, respectively. A superscript $*$ marks non-significant values ($p > 0.05$). ML: Male-Light, MD: Male-Dark, FL: Female-Light, FD: Female-Dark. WD $> 3$ is highlighted.}
\vspace{-0.35cm}
\label{tab:dropfss_celebset}

\setlength{\tabcolsep}{4.2pt}
\renewcommand{\arraystretch}{1.05}

\scriptsize
\begin{tabular}{lcccccccccccc}
\toprule
&
\multicolumn{3}{c}{Accessories$\downarrow$} &
\multicolumn{3}{c}{Hair$\downarrow$} &
\multicolumn{3}{c}{Pose$\downarrow$} &
\multicolumn{3}{c}{Multi-axis$\downarrow$} \\
\cmidrule(lr){2-4}
\cmidrule(lr){5-7}
\cmidrule(lr){8-10}
\cmidrule(lr){11-13}
Model &
Mean: 2 worst groups & $WD_g$ & $WD_s$ &
Mean: 2 worst groups & $WD_g$ & $WD_s$ &
Mean: 2 worst groups & $WD_g$ & $WD_s$ &
Mean: 2 worst groups & $WD_g$ & $WD_s$ \\
\midrule

Flux-Dev &
FD:89.6 \quad MD:85.5 & 0.9 & \textbf{3.7} &
FD:88.7 \quad MD:85.8 & 0.5 & \textbf{4.2} &
FD:89.9 \quad MD:85.2 & 1.3 & \textbf{4.0} &
FD:89.3 \quad MD:85.1 & 0.8 & \textbf{3.6} \\

Kontext-Pro &
ML:28.1 \quad FL:26.3 & 2.2 & 2.9 &
FL:22.0 \quad ML:21.7 & 1.7 & 1.3 &
FD:33.5 \quad FL:32.6 & \textbf{5.6} & 1.1$^{*}$ &
FL:31.0 \quad ML:29.0 & 2.0 & 2.0 \\

Nano-Banana &
ML:27.2 \quad MD:26.0 & 2.2 & 1.6 &
MD:17.5 \quad FL:16.8 & 0.4$^{*}$ & 1.1 &
MD:25.1 \quad FD:23.8 & 0.7$^{*}$ & 2.2 &
FL:24.1 \quad ML:24.0 & 1.6 & 1.2 \\

Bagel-Edit &
ML:29.2 \quad MD:27.2 & 2.4 & 1.9 &
MD:19.4 \quad FD:18.8 & 1.9 & 1.6 &
MD:58.0 \quad FD:57.6 & 1.2$^{*}$ & 2.2 &
MD:34.8 \quad ML:33.8 & 1.3 & 0.9$^{*}$ \\

Qwen-Edit &
FL:50.6 \quad FD:47.3 & \textbf{6.4} & \textbf{4.8} &
FL:52.0 \quad FD:47.8 & \textbf{7.6} & \textbf{3.1} &
FD:45.4 \quad FL:43.8 & \textbf{7.7} & 1.5 &
FL:49.8 \quad FD:45.3 & \textbf{7.4} & 2.9 \\

SeedEdit 3.0 &
ML:27.3 \quad MD:25.7 & \textbf{3.5} & 1.5 &
MD:20.6 \quad ML:19.2 & \textbf{3.1} & 1.4 &
FL:46.0 \quad MD:46.0 & 0.5$^{*}$ & 0.9$^{*}$ &
MD:32.3 \quad ML:31.9 & 2.0 & 0.9 \\

\bottomrule
\end{tabular}

\vspace{-0.25cm}
\end{table*}

\begin{table*}[t]
\centering
\caption{$SC$ (higher is better) disparity on CelebA. We reported mean $SC$ values for Male (M) and Female (F). In each mean column, left: worst, right: best group. $WD_g$ denotes the Wasserstein distance between the male and female $SC$ distributions. A superscript $*$ marks non-significant values ($p > 0.05$). WD > 3 is highlighted. Values are scaled by $100$.}
\vspace{-0.35cm}
\label{tab:sc_celeba}

\setlength{\tabcolsep}{4.2pt}
\renewcommand{\arraystretch}{1.05}

\scriptsize
\begin{tabular}{lcccccccc}
\toprule
&
\multicolumn{2}{c}{Accessories$\uparrow$} &
\multicolumn{2}{c}{Hair$\uparrow$} &
\multicolumn{2}{c}{Pose$\uparrow$} &
\multicolumn{2}{c}{Multi-axis$\uparrow$} \\
\cmidrule(lr){2-3}
\cmidrule(lr){4-5}
\cmidrule(lr){6-7}
\cmidrule(lr){8-9}
Model &
Mean & $WD_g$ &
Mean & $WD_g$ &
Mean & $WD_g$ &
Mean & $WD_g$ \\
\midrule

Flux-Dev &
M:33.2 \quad F:34.1 & 0.9$^{*}$ &
F:28.2 \quad M:32.9 & \textbf{4.6} &
F:30.1 \quad M:33.0 & \textbf{3.0} &
F:29.0 \quad M:31.8 & 2.9 \\

Kontext-Pro &
F:73.4 \quad M:73.5 & 1.8 &
M:75.5 \quad F:76.7 & 2.0 &
F:46.5 \quad M:48.6 & 2.6 &
F:63.5 \quad M:64.6 & 1.2 \\

Nano Banana &
M:83.7 \quad F:85.6 & 1.9 &
M:87.0 \quad F:88.6 & 1.7 &
F:61.9 \quad M:63.4 & 1.8 &
M:74.0 \quad F:75.9 & 2.0 \\

Bagel-Edit &
M:87.5 \quad F:88.0 & 0.6 &
M:79.0 \quad F:81.2 & 2.3 &
M:58.8 \quad F:60.2 & 1.6 &
M:68.8 \quad F:69.0 & 1.4 \\

Qwen-Edit &
M:78.6 \quad F:80.9 & 2.3 &
F:76.4 \quad M:76.9 & 1.1 &
M:66.1 \quad F:68.7 & 2.6 &
M:63.4 \quad F:65.8 & 2.5 \\

\bottomrule
\end{tabular}

\vspace{-0.25cm}
\end{table*}

\begin{table*}[t]
\centering
\caption{$SC$ (higher is better) disparity on CelebSET. We reported the mean $SC$ values for the two worst intersectional groups. In each mean column, left: worst, right: second-worst group. $WD_g$ and $WD_s$ denote Wasserstein distances between gender and skin-tone groups, respectively. $*$ marks non-significant values ($p > 0.05$). WD $> 3$ is highlighted. Values scaled by $100$.}
\vspace{-0.35cm}
\label{tab:sc_celebset}

\setlength{\tabcolsep}{4.2pt}
\renewcommand{\arraystretch}{1.05}

\scriptsize
\begin{tabular}{lcccccccccccc}
\toprule
&
\multicolumn{3}{c}{Accessories$\uparrow$} &
\multicolumn{3}{c}{Hair$\uparrow$} &
\multicolumn{3}{c}{Pose$\uparrow$} &
\multicolumn{3}{c}{Multi-axis$\uparrow$} \\
\cmidrule(lr){2-4}
\cmidrule(lr){5-7}
\cmidrule(lr){8-10}
\cmidrule(lr){11-13}
Model &
Mean: 2 worst groups & $WD_g$ & $WD_s$ &
Mean: 2 worst groups & $WD_g$ & $WD_s$ &
Mean: 2 worst groups & $WD_g$ & $WD_s$ &
Mean: 2 worst groups & $WD_g$ & $WD_s$ \\
\midrule

Flux-Dev &
ML:28.1 \quad FL:28.9 & 0.8$^{*}$ & 1.0$^{*}$ &
FL:25.1 \quad FD:26.6 & \textbf{5.5} & 1.3 &
FD:26.2 \quad FL:27.8 & 2.9 & 1.2 &
FL:25.3 \quad FD:27.3 & \textbf{4.1} & 0.3$^{*}$ \\

Kontext-Pro &
FL:72.2 \quad ML:73.5 & 2.8 & 0.8$^{*}$ &
FL:74.6 \quad ML:75.2 & 0.6$^{*}$ & 2.6 &
MD:35.1 \quad FD:35.4 & \textbf{3.3} & 2.5 &
FD:60.4 \quad FL:60.5 & 1.2$^{*}$ & 1.4 \\

Nano Banana &
FD:74.8 \quad MD:80.9 & 1.5 & \textbf{4.4} &
ML:84.2 \quad FL:86.8 & 2.4 & 2.4 &
FD:58.2 \quad FL:58.8 & 2.1 & 1.0$^{*}$ &
MD:71.6 \quad FL:73.1 & 1.1$^{*}$ & 1.1$^{*}$ \\

Bagel-Edit &
FL:85.7 \quad FD:86.9 & 1.5 & 0.7$^{*}$ &
MD:76.7 \quad FL:77.8 & 1.3 & 0.5$^{*}$ &
FL:59.6 \quad FD:60.8 & 1.1$^{*}$ & 0.8$^{*}$ &
ML:67.2 \quad FL:67.5 & 0.7$^{*}$ & 1.4 \\

Qwen-Edit &
ML:82.0 \quad FL:82.4 & 1.2 & 1.2 &
FL:76.2 \quad ML:79.7 & 2.9 & 2.7 &
ML:60.7 \quad MD:61.7 & \textbf{5.1} & 1.0 &
ML:64.9 \quad FL:66.2 & 1.3 & \textbf{3.3} \\

SeedEdit 3.0 &
FL:85.7 \quad FD:86.0 & 1.0 & 0.8 &
ML:80.9 \quad MD:82.1 & 2.2 & 1.1 &
FD:62.5 \quad ML:63.5 & 0.9$^{*}$ & 1.1$^{*}$ &
FD:67.1 \quad MD:68.2 & 0.8$^{*}$ & 1.3 \\

\bottomrule
\end{tabular}

\vspace{-0.4cm}
\end{table*}



\subsubsection{Overall comparsions}



We compared identity preservation across test sets. In Table \ref{tab:face_id_drop_overall}, for each model, we computed a mean $\operatorname{Drop}_{\mathrm{FSS}}$ on each test set by averaging the $\operatorname{Drop}_{\mathrm{FSS}}$ percentages across all sessions in the test set.  A low $\operatorname{Drop}_{\mathrm{FSS}}$ is better, as it indicates that the model better preserved facial identity.


\textbf{For most models, FSS dropped in the range of 22 - 32\% across four edits}. In Table \ref{tab:face_id_drop_overall}, (Avg. column), for CelebA, all models (except Flux-Dev) showed average $\operatorname{Drop}_{\mathrm{FSS}}$ in the range of $28.7 - 32.7\%$. In CelebSET, we saw similar average $\operatorname{Drop}_{\mathrm{FSS}}$ in $4$ out of $6$ models ($22.3 - 30.8\%$). Across CelebA and CelebSET, most models best preserved facial identity for hair edits (blue numbers), and worst preserved identity for pose edits (red numbers).

\textbf{Nano-Banana and Kontext-Pro performed best at preserving facial identity, while Flux-Dev performed the worst}. In Table \ref{tab:face_id_drop_overall}, for CelebA, Kontext-Pro showed the best aggregated $\operatorname{Drop}_{\mathrm{FSS}}$ of $28.68\%$. For CelebSET, Nano-Banana showed the best aggregated $\operatorname{Drop}_{\mathrm{FSS}}$ of $22.25\%$. For Flux-Dev, the aggregated $\operatorname{Drop}_{\mathrm{FSS}}$ was very high: $>80\%$ in both CelebA and CelebSET.

To compare editing alignment, for each model, we computed an average $SC$ score per session by averaging the $SC$ scores across the edits in that session. In Table \ref{tab:sc_overall}, for each model, we calculated the mean of the averaged $SC$ scores for all sessions in each test set. We scaled the $SC$ values by $100$. A higher $SC$ is better, as it indicates that the model correctly executed the editing instruction.

\textbf{Editing models best execute accessory and hair edits, and worst execute pose edits.} In Table \ref{tab:sc_overall}, for CelebA and CelebSET, models showed the best (highest) $SC$ either on accessory or on hair edits (blue numbers). Across CelebA and CelebSET, most models reported their worst $SC$ on pose-edits (red numbers).

\textbf{Nano-Banana and Seed-Edit are the best models to execute edits, while Flux-Dev is the worst}. In Table \ref{tab:sc_overall}, for CelebA, Nano-Banana showed the best average $SC$ at $79.6$. For CelebSET, both Nano-Banana and Seed-Edit showed the best average $SC$ at $77$. Flux-Dev $SC$ was much lower than other models in both datasets.

We examined overediting patterns in models. For each model, across all test sets, we computed one Overediting Matrix each for CelebA and CelebSET. We used this matrix to identify frequent overedits. For example, assume we ask a model to add a ``handlebar mustache'' to $100$ images. For $15$ or more of these images, say our model did the same overedit --- changing the hairstyle to curly hair. In this case, we call ``handlebar mustache $\rightarrow$ curly hair'' as an overedit pattern. In Table \ref{tab:overedit_patterns}, we showed the most common overedit patterns for all models in CelebA and CelebSET. 

\textbf{Models frequently made overedits when performing hair-based edits.} In Table \ref{tab:overedit_patterns}, for CelebA and CelebSET, we observed common overedit patterns for hair-based edits (e.g, mutton chops $\rightarrow$ handlebar mustache, bob cut $\rightarrow$ straight hair). These overedit patterns were prevalent in Bagel-Edit, Kontext-Pro, and Qwen-Edit.

To compare the amount of overediting across models, we used the OER metric (Section \ref{subsec:overediting}). In Table \ref{tab:oer_overall}, for each model, we reported the mean OER for a test set by averaging the OER values across all sessions in that set. We scaled OER values by $100$. A lower OER is better, as it indicates fewer overedits. 


\textbf{Editing models created the fewest overedits for accessory or pose edits, and the most overedits for multi-axis edits}. In Table \ref{tab:oer_overall}, for CelebA and CelebSET, almost all models achieved their best (lowest) OER either on accessory or on pose edits (blue numbers). Almost all models achieved their worst OER on multi-axis edits (red numbers).

\textbf{Nano-Banana and SeedEdit created the fewest overedits in a session.} In Table \ref{tab:oer_overall} (Avg. column), for CelebA, Nano-Banana showed the lowest average OER at $13.8$. For CelebSET, SeedEdit showed the lowest average OER at $17.49$.

\begin{table*}[t]
\centering
\caption{OER (lower is better) disparity on CelebA. We reported mean OER values for Male (M) and Female (F). In each mean column, left: worst, right: best group. $WD_g$ denotes the Wasserstein distance between the male and female OER distributions. $*$ marks non-significant values ($p > 0.05$). WD $> 3$ is highlighted. Values are scaled by $100$.}
\vspace{-0.35cm}
\label{tab:oer_celeba}

\setlength{\tabcolsep}{4.2pt}
\renewcommand{\arraystretch}{1.05}

\scriptsize
\begin{tabular}{lcccccccc}
\toprule
&
\multicolumn{2}{c}{Accessories$\downarrow$} &
\multicolumn{2}{c}{Hair$\downarrow$} &
\multicolumn{2}{c}{Pose$\downarrow$} &
\multicolumn{2}{c}{Multi-axis$\downarrow$} \\
\cmidrule(lr){2-3}
\cmidrule(lr){4-5}
\cmidrule(lr){6-7}
\cmidrule(lr){8-9}
Model &
Mean Values & $WD_g$ &
Mean Values & $WD_g$ &
Mean Values & $WD_g$ &
Mean Values & $WD_g$ \\
\midrule

Flux-Dev &
M:82.5 \quad F:57.1 & \textbf{25.3} &
M:85.0 \quad F:51.6 & \textbf{33.5} &
M:65.3 \quad F:33.8 & \textbf{31.5} &
M:84.4 \quad F:50.0 & \textbf{34.4} \\

Kontext-Pro &
M:14.9 \quad F:8.6 & \textbf{6.3} &
M:25.5 \quad F:14.7 & \textbf{10.8} &
M:21.9 \quad F:12.1 & \textbf{10.0} &
M:28.3 \quad F:18.5 & \textbf{9.8} \\

Nano Banana &
M:15.0 \quad F:6.8 & \textbf{8.3} &
M:18.0 \quad F:12.6 & \textbf{5.3} &
F:10.5 \quad M:9.8 & 1.0$^{*}$ &
M:21.9 \quad F:14.2 & \textbf{7.7} \\

Bagel-Edit &
M:11.7 \quad F:6.0 & \textbf{5.9} &
M:23.5 \quad F:14.4 & \textbf{9.1} &
M:16.1 \quad F:12.0 & \textbf{4.5} &
M:26.9 \quad F:16.3 & \textbf{10.6} \\

Qwen-Edit &
M:16.2 \quad F:11.7 & \textbf{4.5} &
M:30.4 \quad F:20.4 & \textbf{10.0} &
M:26.4 \quad F:14.7 & \textbf{11.7} &
M:33.3 \quad F:22.4 & \textbf{11.1} \\

\bottomrule
\end{tabular}

\vspace{-0.2cm}
\end{table*}

\begin{table*}[t]
\centering
\caption{OER (lower is better) disparity on CelebSET. We reported the mean OER values for the two worst intersectional groups. In each mean column, left: worst, right: second-worst group. $WD_g$ and $WD_s$ denote Wasserstein distances between gender and skin-tone groups, respectively. $*$ marks non-significant values ($p > 0.05$). WD $> 3$ is highlighted. Values scaled by $100$.}
\vspace{-0.35cm}
\label{tab:oer_celebset}

\setlength{\tabcolsep}{4.2pt}
\renewcommand{\arraystretch}{1.05}

\scriptsize
\begin{tabular}{lcccccccccccc}
\toprule
&
\multicolumn{3}{c}{Accessories$\downarrow$} &
\multicolumn{3}{c}{Hair$\downarrow$} &
\multicolumn{3}{c}{Pose$\downarrow$} &
\multicolumn{3}{c}{Multi-axis$\downarrow$} \\
\cmidrule(lr){2-4}
\cmidrule(lr){5-7}
\cmidrule(lr){8-10}
\cmidrule(lr){11-13}
Model &
Mean: 2 worst groups & $WD_g$ & $WD_s$ &
Mean: 2 worst groups & $WD_g$ & $WD_s$ &
Mean: 2 worst groups & $WD_g$ & $WD_s$ &
Mean: 2 worst groups & $WD_g$ & $WD_s$ \\
\midrule

Flux-Dev &
MD:93.7 \quad ML:69.9 & \textbf{23.1} & \textbf{19.9} &
MD:106.7 \quad ML:76.9 & \textbf{35.1} & \textbf{24.5} &
MD:84.5 \quad ML:56.0 & \textbf{32.4} & \textbf{24.4} &
MD:101.3 \quad ML:81.1 & \textbf{36.8} & \textbf{22.3} \\

Kontext-Pro &
MD:25.5 \quad ML:16.4 & \textbf{8.9} & \textbf{7.1} &
MD:38.5 \quad FD:20.9 & \textbf{5.7} & \textbf{13.3} &
MD:19.0 \quad ML:16.6 & \textbf{6.4} & 2.3$^{*}$ &
MD:41.4 \quad ML:24.3 & \textbf{9.2} & \textbf{13.1} \\

Nano Banana &
FD:50.3 \quad MD:32.9 & \textbf{8.5} & \textbf{22.0} &
MD:32.1 \quad FD:15.3 & \textbf{6.6} & \textbf{13.8} &
MD:15.6 \quad ML:9.2 & \textbf{3.9} & \textbf{3.7} &
MD:39.0 \quad ML:21.5 & \textbf{9.8} & \textbf{11.4} \\

Bagel-Edit &
MD:19.9 \quad FL:15.0 & \textbf{4.9} & \textbf{4.7} &
MD:38.7 \quad FL:23.2 & \textbf{7.1} & \textbf{13.2} &
MD:17.3 \quad ML:14.4 & \textbf{3.8} & 1.4$^{*}$ &
MD:42.8 \quad ML:29.1 & \textbf{12.5} & \textbf{11.5} \\

Qwen-Edit &
MD:36.2 \quad ML:22.2 & \textbf{9.0} & \textbf{9.2} &
MD:49.9 \quad FL:28.5 & \textbf{6.5} & \textbf{14.8} &
MD:25.3 \quad ML:21.5 & \textbf{10.3} & 2.4$^{*}$ &
MD:52.1 \quad ML:31.6 & \textbf{11.8} & \textbf{15.3} \\

SeedEdit 3.0 &
MD:17.8 \quad ML:8.1 & \textbf{4.5} & \textbf{6.7} &
MD:33.3 \quad ML:17.0 & \textbf{8.0} & \textbf{12.4} &
MD:23.5 \quad ML:18.7 & \textbf{4.7} & 2.6 &
MD:46.4 \quad FD:35.0 & \textbf{6.4} & \textbf{24.5} \\

\bottomrule
\end{tabular}

\vspace{-0.3cm}
\end{table*}

\subsubsection{Demographic comparisons}
\label{subsec: demographic_results}
We examined demographic biases in identity preservation using Wasserstein Distance or WD (Section \ref{subsec:wd}). If a model has WD $>3$ (and p $<0.05$), we consider the model to have a prominent bias (refer to Section \ref{app:wd_thresh_selection} to understand why we chose WD $>3$). In Table \ref{tab:dropfss_celeba}, we reported $\operatorname{WD}_{\mathrm{g}}$ (gender disparity between male and female $\operatorname{Drop}_{\mathrm{FSS}}$ values) for CelebA. We also reported mean values for both gender groups. In the ``mean'' column, the left value is the mean of the worst performing group, and the right value is the mean of the best group. 

For CelebSET (Table \ref{tab:dropfss_celebset}), we reported $\operatorname{WD}_{\mathrm{g}}$ and $\operatorname{WD}_{\mathrm{s}}$ (skin-tone disparity between light and dark groups). CelebSET has four intersectional groups: male-light (ML), male-dark (MD), female-light (FL), and female-dark (FD). For conciseness, we only reported the mean values of the two worst intersectional groups. In the ``mean'' column, the left value is the mean of the worst-performing group, and the right value is the mean of the second-worst group. The two-worst mean values, combined with $\operatorname{WD}_{\mathrm{g}}$ and $\operatorname{WD}_{\mathrm{s}}$ are sufficient to examine bias trends.

\textbf{Qwen-Edit is poor at retaining the facial identity of females.} In CelebA (Table \ref{tab:dropfss_celeba}), for Qwen-Edit we observed a gender bias for pose-edits ($\operatorname{WD}_{\mathrm{g}}$ = $7.3$). The worst performing group was female (F). In CelebSET (Table \ref{tab:dropfss_celebset}), we observed a gender bias in Qwen-Edit for all test sets. For each test set, the worst two groups were females (FL and FD). 

\textbf{Flux-dev is poor at retaining the facial identity of people with darker skin tones.} In CelebSET (Table \ref{tab:dropfss_celebset}), for Flux-Dev, we observed skin-tone bias in all test sets ($\operatorname{WD}_{\mathrm{s}}$ $> 3$). The two worst groups were always dark-skinned people (FD and MD).

We showed age bias results for identity preservation in Section \ref{app:age_bias}. With CelebA, we did not observe clear age bias trends (we saw age bias in a few models only for pose-edits). With CelebSET, we observed age bias across all test sets. Most models (except Qwen-Edit) showed worse $\operatorname{Drop}_{\mathrm{FSS}}$ on old faces.

To examine demographic bias in editing alignment, we computed WD scores using $SC$ values. We used the same reporting format as $\operatorname{Drop}_{\mathrm{FSS}}$. CelebA results are shown in Table \ref{tab:sc_celeba}, and CelebSET results are shown in Table \ref{tab:sc_celebset} (for $SC$, lower score is worse).





\textbf{When executing edits, most models showed marginal (or no) gender and skin-tone bias for most edit types.} In most models, we did not observe clear trends for gender and skin-tone bias across both datasets (Tables \ref{tab:sc_celeba} and \ref{tab:sc_celebset}). With Flux-Dev, we observed gender bias ($WD_g >3$) in some test sets across both datasets. In these cases, the worst performing group was always female. 

We showed our age bias analysis for editing alignment in Section \ref{app:age_bias}. On CelebA, only Flux-Dev exhibited age bias, with lower $SC$ for older faces. On CelebSET, we observed no age bias trends.

To examine demographic bias in overediting, we computed WD using OER values. CelebA results are shown in Table \ref{tab:oer_celeba}, and CelebSET results are shown in Table \ref{tab:oer_celebset}.

\textbf{Models created more overedits when editing males, especially dark-skinned males.} In CelebA (Table \ref{tab:oer_celeba}) across all test sets, we observed a significant gender bias for all models. The worst group (highest OER) was Male (M). In CelebSET (Table \ref{tab:oer_celebset}), across test sets, we observed significant gender or skin-tone bias. In all models, the worst group was dark-skinned males (MD).

In Section \ref{app:age_bias}, we showed age bias results for overediting. For CelebA and CelebSET, all models showed age bias in all test sets. In most cases, models showed worse OER on old faces. For all our demographic bias analyses (with $\operatorname{Drop}_{\mathrm{FSS}}$, $SC$, and OER), we provided mean values for all CelebA and CelebSET groups (with confidence intervals) in Section \ref{sec:all_mean}. 

We conducted an additional analysis to examine which specific editing attributes commonly led to demographic biases in overediting. Refer to our analysis in Section \ref{app:per_operation_bias}.

To ensure our bias insights are reliable, we conducted additional experiments: $(1)$ In Section \ref{app:demographic_bias_fss_base_images}, we evaluated if FSS showed demographic bias in base images. We found no gender, age, or skin tone bias. $(2)$ In Section \ref{app:occlusion_sensitivity}, we conducted an experiment to check if FSS is sensitive to occlusions (if a face in the base image is occluded by glasses or hats, does FSS drop w.r.t a clean face?). We found that FSS was similar for faces with or without occlusions. These experiments proved that any bias in identity preservation came from the editing models. $(3)$ In Section \ref{app:gendered_representation_analysis}, we conducted a controlled experiment to check if text-guided models were biased toward stereotypical edits (do models perform better on males if instructions are male-centric? --- e.g., add a handlebar mustache). We found that models do not show such stereotypical biases.

\section{Discussion}

Our results show that Nano-Banana is the best model for facial editing and Flux-Dev is the worst. While Flux-Dev supports image editing, it was optimized for image generation. Flux-Dev often produces cartoonish edits, resulting in poor evaluation scores.

Models are reliable for short facial editing sequences ($3 - 4$ edits). We do not recommend using these models for longer sequences as their performance consistently drops with every edit (refer to our analysis in Section \ref{app:face_id_progression}).

All models overedit: on average, for every $100$ instructions (25 sequences), they make about $25$ overedits. These overedits often come from spurious correlations learned during training. For example, when asked to add a bob cut, models often change the hair type to straight because they have learned to associate bob cuts with straight hair. To mitigate this issue, models should be trained on diverse data (e.g., bob cuts with curly hair). If retraining is difficult, we can use methods to help models distinguish between concepts that frequently co-occur but are not related. 


\section{Limitations}

In our state tracking approach (Section \ref{subsec:overediting}), we can only identify overedits that fall within the $35$ facial editing categories in Face-Edit-Attributes. For instance, Qwen-Edit often creates a toothbrush when asked to add a ``toothbrush mustache''. Such overedits cannot be tracked using our method. Concept activation vectors would be an important future step to make our state tracking approach generalizable to all types of overedits.

To analyze skin-tone bias, we used only binary labels (light and dark), since CelebSET provided only binary skin-tone annotations. Numerical skin-tone labels (e.g., the Fitzpatrick scale, which ranges from $1$: lightest to $6$: darkest) could have made our insights more fine-grained. For example, with Fitzpatrick, we could have identified whether models produced more overedits on dark-skinned men ($4 - 5$ Fitzpatrick score) or on very dark-skinned men ($5 - 6$ score).

Our Gemini-2.5-Flash state-tracker reliably tracks edits in most cases. However, there are cases where it is poor at identifying edits (e.g., looking left). Refer to Section \ref{app:human_exp} for details. 

\section{Conclusion}

We conducted the first large-scale evaluation of six text-guided image editing models for facial appearance editing. We proposed Face-Edit-Attributes, the largest collection of $169$ facial attributes that enable hair, accessory, and pose-based edits. We also released our repository of $\sim$ $1$M edited images, which can be used to advance research in facial editing. To our knowledge, we conducted the first systematic analysis of demographic bias (gender, skin tone, and age) in facial editing. To count how many and which overedits occurred in a model, we proposed a novel overediting matrix. Our results show that several text-guided models can be used off-the-shelf for facial appearance editing applications. However, it is important to note that all models overedit. Future research on text-guided models should focus on strategies to mitigate overedits, especially when they edit dark-skinned and older males.

\section{Acknowledgements}

We would like to thank 2nd SET AI Corp. for providing GPU resources for our study. We would also like to thank Jeff Smith for his valuable suggestions throughout the project.

\bibliographystyle{ACM-Reference-Format}

\balance

\bibliography{references}

\appendix

\clearpage
\appendix

\section{Related Work}
\label{sec:app_rel_work}

\paragraph{Datasets and benchmarks for facial appearance editing} Popular 3DMMs \cite{khan2025instaface, ding2023diffusionrig} have been trained on FFHQ \cite{karras2019style}, a popular face dataset. These models can be tested only on three attributes: pose, expression, and lighting. Popular GAN-based editing models \cite{choi2018stargan, kwak2020cafe, chu2020sscgan} have been evaluated on CelebA \cite{liu2015deep}, a popular face dataset with $40$ facial attribute annotations (e.g., bangs, smiling, young). To perform facial edits, these models simply used $5 - 13$ attribute names from CelebA. Recent works have built benchmarks that allow us to perform facial edits using captions (e.g., this person has bangs) rather than using attribute names (e.g., bangs). Popular benchmarks are FFLEBench \cite{xian2025edit} (uses images from FFHQ), Multimodal-CelebA-HQ or MM-CelebA-HQ \cite{xia2021tedigan} (uses images from CelebA-HQ \cite{karras2017progressive} --- a subset of CelebA with high-quality face images), CelebA-dialog \cite{jiang2021talk} (images from CelebA), and CHAT-EDIT \cite{cui2023chatedit} (images from CelebA-HQ). In FFLEBench, captions are generated by an MLLM (the MLLM is given a face image and a semantic segment of the facial regions). They do not provide the LLM with an attribute list. As a result, the LLM generates vague editing captions (e.g., sparse eyebrows, tall nose, smooth face). This makes it hard to evaluate if the model executed the edit correctly. In MM-CelebA-HQ, captions are generated using a template-based approach (using a list of $40$ attributes). The captions often target multiple attributes (e.g., ``This old woman has big lips, pale skin, and gray hair''). This does not allow us to check how the model performed on each edit. MM-CelebA-HQ and FFLEBench have several editing captions that can significantly impact identity preservation (e.g., tall nose, big eyes, big lips). CHAT-EDIT and CelebA-dialog are robust benchmarks: $(1)$ they provide clear and actionable editing captions (e.g., she should have black hair), $(2)$ they avoid attributes that modify facial structure (e.g., editing lips, nose), and $(3)$ they only edit one attribute at a time. However, they support limited attributes --- CHAT-EDIT only supports $21$ attributes, while CelebA-dialog only supports $5$ attributes. 

\paragraph{Benchmarks for Text-Guided Image Editing} Several benchmarks have been proposed to evaluate text-guided image editing models (e.g., Nano-Banana \cite{nano}) on generic scenes (e.g., color the cars pink, have the girl in the image ride a dragon). Well-known benchmarks, such as InsturctPix2Pix \cite{brooks2023instructpix2pix} and HQ-Edit \cite{hui2024hq}, evaluate editing models on synthetically generated base images. In InstructPix2Pix, base images are generated from human-written captions (e.g., "a girl riding a horse") using Stable Diffusion \cite{rombach2022high} and Prompt-2-Prompt \cite{hertz2022prompt} (editing instruction is generated using GPT-3 \cite{brown2020language}). In HQ-Edit, the images are generated from human-written captions using DALLE3 \cite{betker2023improving} (editing instructions are human-written). Other popular benchmarks (e.g., GIE-Bench \cite{qian2025gie}, Human-MLLM-Bench \cite{liu2026human}, Magic-Brush \cite{zhang2023magicbrush}, Ref-Edit \cite{pathiraja2025refedit}, and Omni-Edit \cite{wei2024omniedit}) use real-world images from scene datasets (e.g., COCO \cite{lin2014microsoft}, RefCOCO \cite{yu2016modeling}) or from online image repositories like Pexels \cite{pexels}. None of these benchmarks evaluated editing models on face datasets like CelebA, CelebA-HQ, and FFHQ. 

\paragraph{Overediting Metrics} Several metrics have been proposed to evaluate overediting in models. All metrics report overediting with a numerical score. Rubric-based metrics like VIEScore \cite{ku2024viescore} and Human-MLLM \cite{liu2026human} give an overediting score between $0 - 10$ (score generated by an MLLM). A score of $0$ indicates significant overediting in the image, while $10$ indicates minimal overediting. Metrics like GIE-Bench \cite{qian2025gie} and BPM \cite{li2025balancing} first locate regions that should not be affected during an edit. For these regions, they compute a mean-squared (MSE) error between the pixels in the original and in the edited image. A higher MSE error indicates more overediting. Newer metrics like DICE \cite{baraldi2025changed} instruct an MLLM to identify what objects were changed between the original and the edited image (e.g., dog was added, frisbee was removed). They also instruct the MLLM to provide region co-ordinates for the changes. They quantify overediting as the percentage of the total image area where unwanted edits occurred. DICE cannot always identify fine-grained overedits since change detection in the MLLM is open-ended (i.e., the MLLM may say ``a hat was added'', when we need a specific description like ``a beret hat was added''). To identify spurious correlations in facial editing, identifying fine-grained overedits is critical.

\paragraph{Bias Studies for Faces} 

Most bias studies for faces have focused on facial recognition, localization, and generation. Popular works have studied biases in face recognition. \cite{buolamwini2018gender} showed that commercial face recognition models (e.g., Microsoft \cite{ms_api} and IBM \cite{ibm_api} models) showed demographic biases in gender classification: all models classified gender more accurately for male faces (compared to females) and for lighter faces (compared to darker ones). \cite{wang2019racial} showed that both commercial (e.g., Microsoft \cite{azure_api} and Amazon \cite{amazon_api} APIs) and open-source (e.g., Center-loss \cite{wen2016discriminative}, ArcFace \cite{deng2019arcface}) face recognition models have ethnicity biases: models showed highest recognition accuracy on Caucasian faces and lowest accuracy on African faces. \cite{karkkainen2021fairface} showed that face classification models have ethnicity biases because they are trained on datasets where some ethnicities (e.g., white faces) are over-represented. They proposed FairFace, a dataset with a balanced number of samples for $7$ ethnicities (e.g., White, Black, South-East Asian, Latino). Compared to other face datasets, performance disparities between ethnicities were less when models were trained on FairFace. Some recent works have studied biases in face localization models. \cite{mittal2023face, xiang2025fair} showed that face detection models like MTCNN \cite{zhang2016joint} and Retinaface \cite{deng2020retinaface} showed better face detection accuracy for females compared to males. Several works have studied biases in face generation. \cite{perera2023analyzing, maluleke2022studying} showed that StyleGAN2-ADA \cite{karras2020training} generated more white faces when it was trained on FFHQ (a face dataset dominated by white faces). They showed that the bias (toward generating more white faces) disappeared when StyleGAN2-ADA was trained on a balanced FairFace dataset (equal samples of white and black faces). \cite{perera2023analyzing} showed that DDPM \cite{choi2022perception} (a diffusion-based generation model) showed biases in face generation (generated more female faces compared to males, and more white faces compared to black) even though it was trained on a balanced FairFace dataset. For facial editing, current works have evaluated whether models conform to social stereotypes. \cite{seo2026evaluating} showed that when editing models are asked to edit a female face to look like a ``senior executive'' or a ``CEO'', the model can edit the image to look like a male face (since models strongly stereotype executive positions with men). To our knowledge, current works have not studied demographic biases in editing models for facial appearance edits (e.g., adding a hat, changing hair color).

\section{Face-Edit-Attributes}
\label{app:face_edit_attributes}

\begin{table*}[htbp]
\centering
\footnotesize
\caption{Hair Attributes}
\label{tab:hair_attr_names}
\begin{tabularx}{0.95\textwidth}{P{1cm}P{1.5cm}Y}
\textbf{Sub-axes} & \textbf{Categories} & \textbf{Attributes} \\
\midrule

 & red hair & bright red hair; ruby red hair; intense scarlet red hair; crimson red hair; cherry red hair; rose red hair \\
\addlinespace
 & brown hair & light brown hair; light golden brown hair; honey brown hair; caramel brown hair; toffee brown hair; sandy brown hair; amber brown hair; ginger hazel brown hair; light hazel brown hair \\
\addlinespace
\multirow{9}{*}{color} & black hair & black hair; jet black hair; raven black hair; ebony hair; charcoal black hair \\
\addlinespace
 & white hair & stark white hair; snow white hair; icy white hair; frost white hair; arctic white hair; nordic white hair \\
\addlinespace
 & blue hair & sky blue hair; aqua blue hair; electric blue hair; turquoise blue hair; teal blue hair; cobalt blue hair \\
\addlinespace
 & green hair & lime green hair; neon green hair; bright green hair; apple green hair; emerald green hair; spring green hair \\
\addlinespace
 & pink hair & magenta pink hair; cotton candy pink hair; bubblegum pink hair; candy pink hair; fuchsia pink hair; neon pink hair \\
\addlinespace
 & purple hair & orchid purple hair; bright purple hair; violet purple hair; lavender hair; amethyst hair \\
 & orange hair & bright orange hair; neon peach hair \\

\addlinespace\midrule
\multirow{2}{*}{type} & straight hair & sleek straight hair; silky straight hair; pin-straight hair \\
\addlinespace
 & curly hair & spiral curl; kinky hair; coily hair; tight curl; afro-textured hair; springy curl \\

\addlinespace\midrule
\multirow{3}{*}{style} & buzz cut & buzz cut; military cut \\
\addlinespace
 & bob cut & bob cut; lob cut; chin-length bob; classic bob \\
\addlinespace
 & spiked cut & spiked hair; liberty spike hair; mohawk; gelled spike hair \\

\addlinespace\midrule
 & pencil mustache & thin pencil mustache; classic pencil mustache; parted pencil mustache \\
\addlinespace
\multirow{5}{*}{mustache} & walrus mustache & thick walrus mustache; bushy walrus mustache \\
\addlinespace
 & toothbrush mustache & small toothbrush mustache; classic toothbrush mustache; petite toothbrush mustache; chaplin mustache; mini toothbrush mustache \\
\addlinespace
 & handlebar mustache & curled mustache; classic handlebar mustache; twisted handlebar mustache \\
\addlinespace
 & fumanchu mustache & fumanchu mustache; long fumanchu mustache; traditional fumanchu mustache \\
\addlinespace\midrule
 & clean shaven & clean shaven; no facial hair; shaved; fully shaved; clean shaved \\
\addlinespace
\multirow{5}{*}{beard} & stubble & five o'clock shadow; light stubble; short stubble \\
\addlinespace
 & goatee & classic goatee; chin goatee; goatee; soul patch \\
\addlinespace
 & mutton chops & mutton chops; thick mutton chops; sideburn mutton chops \\
\addlinespace
 & full beard & full beard; thick full beard; long full beard; bushy full beard; dense full beard; grown out full beard; full-face beard \\

\bottomrule
\end{tabularx}
\end{table*}

\begin{table*}[htbp]
\centering
\footnotesize
\caption{Accessory Attributes}
\label{tab:accessory_attr_names}
\begin{tabularx}{0.95\textwidth}{P{1cm}P{2.5cm}Y}
\textbf{Sub-axes} & \textbf{Categories} & \textbf{Attributes} \\
\midrule

\multirow{2}{*}{glasses} & round glasses & round glasses; round sunglasses; circular glasses; circular sunglasses; John Lennon glasses; oval glasses; oval sunglasses \\
\addlinespace
 & rectangle-shaped glasses & square glasses; square sunglasses; rectangular glasses; rectangular sunglasses; rectangle-shaped glasses; rectangle-shaped sunglasses \\

\addlinespace\midrule
 & baseball hat & baseball cap; baseball hat; snapback cap; snapback hat; fitted cap; fitted hat; dad hat \\
\addlinespace
\multirow{6}{*}{hats} & beanie hat & beanie hat; knit hat; wool hat; slouchy beanie hat; winter beanie hat \\
\addlinespace
 & beret hat & beret hat; french beret hat \\
\addlinespace
 & panama hat & panama hat; straw panama hat; classic panama hat \\
\addlinespace
 & bucket hat & bucket hat \\
\addlinespace
 & cowboy hat & cowboy hat; stetson hat; western cowboy hat \\

\bottomrule
\end{tabularx}
\end{table*}

\begin{table*}[htbp]
\centering
\footnotesize
\caption{Pose Attributes}
\label{tab:pose_attr_names}
\begin{tabularx}{0.95\textwidth}{P{1cm}P{3cm}Y}
\textbf{Sub-axes} & \textbf{Categories} & \textbf{Attributes} \\
\midrule

 & looking straight ahead & looking straight ahead; frontal view; looking forward; head straight; front facing; looking ahead; head facing forward \\
\addlinespace
\multirow{3}{*}{head position} & looking left & looking left; looking to the left; facing left; looking at the left; head facing left; head looking left; facing to the left; looking leftward; looking towards the left; looking left side \\
\addlinespace
 & looking right & looking right; looking to the right; facing right; looking at the right; head facing right; head looking right; facing to the right; looking rightward; looking towards the right; looking right side \\

\bottomrule
\end{tabularx}
\vspace{1cm}
\end{table*}

We showed our hair attributes in Table \ref{tab:hair_attr_names}, accessory attributes in Table \ref{tab:accessory_attr_names}, and pose attributes in Table \ref{tab:pose_attr_names}.

\section{MLLM Prompt for an Editing Instruction}
\label{app:mllm_prompt}

\begin{tcolorbox}[
  colback=gray!5,
  colframe=black!30,
  boxrule=0.5pt,
  arc=2pt,
  left=6pt,
  right=6pt,
  top=6pt,
  bottom=6pt,
]
\footnotesize
You are an expert photo editing instructor. Generate natural, human-like photo editing instructions based on the current state of the image.

AVAILABLE VERB POOL:
\begin{itemize}
    \item[-] CHANGE verbs: change, modify, alter, switch, transform, adjust, convert, shift, update, revise, transition, swap, replace, morph
    \item[-] ADD verbs: add, put on, include, wear, apply, give them, have them wear, place on them, attach, equip them with, provide them with, introduce, incorporate, install, mount, adorn with
    \item[-] REMOVE verbs: remove, take off, eliminate, get rid of, strip away, discard, delete, clear, exclude, subtract
    \item[-] KEEP verbs: keep, maintain, preserve, retain, continue with, sustain, hold, stay with, leave
    \item[-] MAKE verbs: make, have them, get them to, pose them, make the person, have the subject, position them to, render, cause, direct them to, compose, set
\end{itemize}

STATE-AWARE INSTRUCTION GENERATION:
You will receive:
\begin{enumerate}
    \item Current image state (what features are present in the image)
    \item Editing Attribute
    \item Axis and sub-axis context
\end{enumerate}

Based on the current state, you MUST choose the appropriate verb:
\begin{itemize}
    \item[-] If the feature is ALREADY PRESENT and matches the edit candidate → use KEEP verbs
    \item[-] If the feature is PRESENT but DIFFERENT from edit candidate → use CHANGE verbs
    \item[-] If the feature is NOT PRESENT → use ADD verbs
    \item[-] If you need to remove something first → use REMOVE verbs (rare)
    \item[-] For poses/expressions → use MAKE verbs
\end{itemize}

CRITICAL RULES:
\begin{enumerate}
    \item Analyze the current state carefully before selecting a verb
    \item If person already has red hair and edit candidate is "red hair" → use "keep the hair color as red"
    \item If person has brown hair and edit candidate is "red hair" → use "change the hair color to red"
    \item If person has no glasses and edit candidate is "round glasses" → use "add round glasses"
    \item If person has square glasses and edit candidate is "round glasses" $\rightarrow$ use "change to round glasses"
    \item Generate natural, conversational instructions (5-8 words typically)
    \item Include appropriate articles (a/an/the) automatically
    \item Vary your verb choice within the appropriate category
    \item **MUST use the EXACT edit candidate phrase verbatim - do NOT paraphrase, simplify, or shorten it**
    \item Return ONLY the instruction, no additional text
\end{enumerate}

\end{tcolorbox}

\section{Editing Instructions}
\label{app:edit_instructions}

We showed our editing instructions for accessory edits in Table \ref{tab:accessory_instructions}, for hair edits in Table \ref{tab:hair_instructions}, and for multi-axis edits in Table \ref{tab:multi_axis_instructions}, for pose edits in Table \ref{tab:pose_instructions}.

\begin{table}[h]
\centering
\footnotesize
\caption{Editing sessions for accessory-based edits}
\label{tab:accessory_instructions}
\begin{tabularx}{0.5\textwidth}{P{1.2cm}P{0.4cm}Y}
\textbf{Session} & \textbf{Step} & \textbf{Prompt} \\
\midrule

\multirow{4}{*}{accessories\_001} & 1 & Add square glasses \\
 & 2 & Add a wool hat \\
 & 3 & Replace their wool hat with a knit hat. \\
 & 4 & Change to rectangular sunglasses \\
\addlinespace\midrule

\multirow{4}{*}{accessories\_002} & 1 & Add a french beret hat. \\
 & 2 & Add circular sunglasses. \\
 & 3 & Change to oval sunglasses \\
 & 4 & Replace the french beret hat with a snapback cap. \\
\addlinespace\midrule

\multirow{4}{*}{accessories\_003} & 1 & Add a classic panama hat. \\
 & 2 & Replace their classic panama hat with a dad hat. \\
 & 3 & Add John Lennon glasses. \\
 & 4 & Change to rectangle-shaped glasses. \\
\addlinespace\midrule

\multirow{4}{*}{accessories\_004} & 1 & Add rectangular glasses. \\
 & 2 & Add a knit hat. \\
 & 3 & Replace their knit hat with a beanie hat. \\
 & 4 & Change to rectangular sunglasses. \\
\addlinespace\midrule

\multirow{4}{*}{accessories\_005} & 1 & Add a winter beanie hat \\
 & 2 & Add round glasses. \\
 & 3 & Change their winter beanie hat to a knit hat. \\
 & 4 & Replace their knit hat with a snapback cap \\
\addlinespace\midrule

\multirow{4}{*}{accessories\_006} & 1 & Add round glasses \\
 & 2 & Add a bucket hat. \\
 & 3 & Replace their bucket hat with a dad hat. \\
 & 4 & Replace their dad hat with a beanie hat. \\
\addlinespace\midrule

\multirow{4}{*}{accessories\_007} & 1 & Add a baseball hat. \\
 & 2 & Keep the baseball hat. \\
 & 3 & Replace their baseball hat with a knit hat. \\
 & 4 & Replace their knit hat with a snapback cap. \\
\addlinespace\midrule

\multirow{4}{*}{accessories\_008} & 1 & Add a bucket hat. \\
 & 2 & Replace their bucket hat with a straw panama hat. \\
 & 3 & Add oval glasses \\
 & 4 & Change to square glasses \\
\addlinespace\midrule

\multirow{4}{*}{accessories\_009} & 1 & Add rectangular glasses \\
 & 2 & Change to square sunglasses. \\
 & 3 & Change their square sunglasses to rectangle-shaped glasses. \\
 & 4 & Change the glasses to oval glasses. \\
\addlinespace\midrule

\multirow{4}{*}{accessories\_010} & 1 & Add a knit hat. \\
 & 2 & Add circular glasses. \\
 & 3 & Keep their circular glasses. \\
 & 4 & Change to a cowboy hat. \\

\bottomrule
\end{tabularx}
\end{table}

\begin{table}[h]
\centering
\footnotesize
\caption{Editing sessions for hair-based edits}
\label{tab:hair_instructions}
\begin{tabularx}{0.5\textwidth}{P{0.7cm}P{0.3cm}Y}
\textbf{Session} & \textbf{Step} & \textbf{Prompt} \\
\midrule

\multirow{4}{*}{hair\_001} & 1 & Add caramel brown hair \\
 & 2 & Add pin-straight hair. \\
 & 3 & Change the hair color to lavender hair. \\
 & 4 & Add a bob cut. \\
\addlinespace\midrule

\multirow{4}{*}{hair\_002} & 1 & Add light hazel brown hair \\
 & 2 & Change the hair color to neon peach hair. \\
 & 3 & Change the hair color to fuchsia pink hair. \\
 & 4 & Change the hair color to raven black hair. \\
\addlinespace\midrule

\multirow{4}{*}{hair\_003} & 1 & Add spiked hair. \\
 & 2 & Add ruby red hair \\
 & 3 & Change the hair color to bright orange hair. \\
 & 4 & Change the hair color to frost white hair. \\
\addlinespace\midrule

\multirow{4}{*}{hair\_004} & 1 & Add sleek straight hair \\
 & 2 & Add rose red hair \\
 & 3 & Change the hair color to bright green hair \\
 & 4 & Add a bushy full beard. \\
\addlinespace\midrule

\multirow{4}{*}{hair\_005} & 1 & Add a bob cut. \\
 & 2 & Add bright red hair \\
 & 3 & Change the hair color to raven black hair. \\
 & 4 & Change the hair color to light brown hair. \\
\addlinespace\midrule

\multirow{4}{*}{hair\_006} & 1 & Add bright purple hair. \\
 & 2 & Add a twisted handlebar mustache. \\
 & 3 & Change the hair to violet purple hair. \\
 & 4 & Change the hair color to ginger hazel brown hair. \\
\addlinespace\midrule

\multirow{4}{*}{hair\_007} & 1 & Add emerald green hair. \\
 & 2 & Add pin-straight hair \\
 & 3 & Add spiked hair. \\
 & 4 & Change the hair color to neon green hair. \\
\addlinespace\midrule

\multirow{4}{*}{hair\_008} & 1 & Add emerald green hair \\
 & 2 & Add a mini toothbrush mustache \\
 & 3 & Change the hair color to lavender hair. \\
 & 4 & Add a bushy full beard. \\
\addlinespace\midrule

\multirow{4}{*}{hair\_009} & 1 & Add bright purple hair \\
 & 2 & Add sideburn mutton chops. \\
 & 3 & Add springy curl to their hair. \\
 & 4 & Change their hair color to crimson red hair. \\
\addlinespace\midrule

\multirow{4}{*}{hair\_010} & 1 & Add a fumanchu mustache. \\
 & 2 & Add lime green hair \\
 & 3 & Change their fumanchu mustache to a curled mustache. \\
 & 4 & Change their hair color to lavender hair. \\

\bottomrule
\end{tabularx}
\end{table}

\begin{table}[h]
\centering
\footnotesize
\caption{Editing sessions for multi-axis edits}
\label{tab:multi_axis_instructions}
\begin{tabularx}{0.48\textwidth}{P{1.2cm}P{0.3cm}Y}
\textbf{Session} & \textbf{Step} & \textbf{Prompt} \\
\midrule

\multirow{4}{*}{multi\_axis\_001} & 1 & Make the person looking ahead. \\
 & 2 & Add emerald green hair \\
 & 3 & Add spiral curl. \\
 & 4 & Have them looking straight ahead. \\
\addlinespace\midrule

\multirow{4}{*}{multi\_axis\_002} & 1 & Add spiked hair. \\
 & 2 & Add square glasses. \\
 & 3 & Add stark white hair \\
 & 4 & Add a thin pencil mustache. \\
\addlinespace\midrule

\multirow{4}{*}{multi\_axis\_003} & 1 & Add a slouchy beanie hat. \\
 & 2 & Make the head facing right. \\
 & 3 & Pose them looking right \\
 & 4 & Adjust them to facing right. \\
\addlinespace\midrule

\multirow{4}{*}{multi\_axis\_004} & 1 & Add rectangle-shaped glasses. \\
 & 2 & Add a full beard. \\
 & 3 & Add a fitted cap. \\
 & 4 & Change the full beard to a full-face beard. \\
\addlinespace\midrule

\multirow{4}{*}{multi\_axis\_005} & 1 & Add coily hair \\
 & 2 & Add amethyst hair. \\
 & 3 & Add sideburn mutton chops. \\
 & 4 & Change their hair type to silky straight hair. \\
\addlinespace\midrule

\multirow{4}{*}{multi\_axis\_006} & 1 & Add ebony hair. \\
 & 2 & Add a panama hat \\
 & 3 & Add a chaplin mustache. \\
 & 4 & Change the mustache to a thin pencil mustache. \\
\addlinespace\midrule

\multirow{4}{*}{multi\_axis\_007} & 1 & Add sky blue hair \\
 & 2 & Change the hair color to cobalt blue hair \\
 & 3 & Make the head looking left. \\
 & 4 & Add pin-straight hair. \\
\addlinespace\midrule

\multirow{4}{*}{multi\_axis\_008} & 1 & Add a petite toothbrush mustache \\
 & 2 & Make them looking straight ahead. \\
 & 3 & Pose them front facing. \\
 & 4 & Shift their head to looking right \\
\addlinespace\midrule

\multirow{4}{*}{multi\_axis\_009} & 1 & Add a petite toothbrush mustache \\
 & 2 & Add a snapback cap. \\
 & 3 & Add honey brown hair. \\
 & 4 & Change to a mini toothbrush mustache. \\
\addlinespace\midrule

\multirow{4}{*}{multi\_axis\_010} & 1 & Make the person looking right side \\
 & 2 & Add cherry red hair \\
 & 3 & Change the hair color to caramel brown hair. \\
 & 4 & Shift the head to a frontal view. \\

\bottomrule
\end{tabularx}
\end{table}

\clearpage

\begin{table}[h]
\centering
\footnotesize
\caption{Editing sessions for pose-based edits}
\label{tab:pose_instructions}
\begin{tabularx}{0.5\textwidth}{P{1cm}P{0.5cm}Y}
\textbf{Session} & \textbf{Step} & \textbf{Prompt} \\
\midrule

\multirow{4}{*}{poses\_001} & 1 & Make them looking right. \\
 & 2 & Change the head position to looking rightward. \\
 & 3 & Change to head facing right \\
 & 4 & Keep them facing right. \\
\addlinespace
\midrule
\multirow{4}{*}{poses\_002} & 1 & Make them looking forward \\
 & 2 & Make them looking to the right. \\
 & 3 & Make the person head straight. \\
 & 4 & Adjust to head facing forward. \\
\addlinespace
\midrule
\multirow{4}{*}{poses\_003} & 1 & Make them have a frontal view \\
 & 2 & Adjust the head position to looking straight ahead. \\
 & 3 & Set their head looking left. \\
 & 4 & Make them looking straight ahead. \\
\addlinespace
\midrule
\multirow{4}{*}{poses\_004} & 1 & Make them looking right side \\
 & 2 & Change to head facing right. \\
 & 3 & Adjust the head position to front facing. \\
 & 4 & Make them looking right side \\
\addlinespace
\midrule
\multirow{4}{*}{poses\_005} & 1 & Make them looking towards the right \\
 & 2 & Keep them looking towards the right. \\
 & 3 & Make them looking to the left. \\
 & 4 & Change the head position to looking left. \\
\bottomrule
\end{tabularx}
\end{table}

\begin{figure*}[h]
  \centering
  \quad
   \par
  \vspace{0.4em}
  \captionsetup[subfigure]{margin={2.5em,0pt}}
  \begin{subfigure}[t]{0.35\textwidth}
    \centering
    \includegraphics[width=\linewidth]{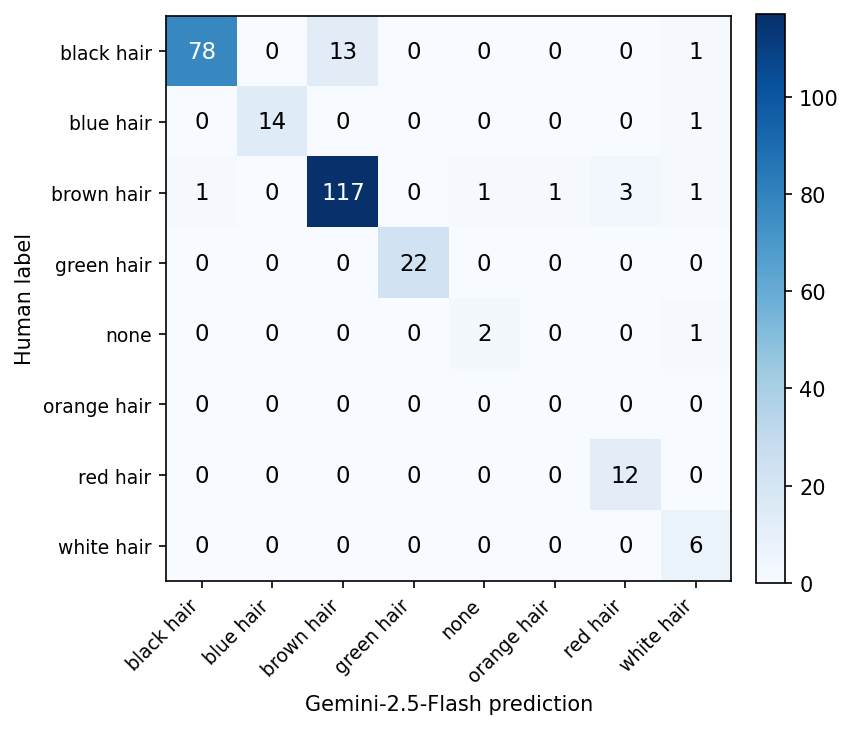}
    \caption{Hair Color}
    \label{fig:cm_hair_color}
  \end{subfigure}\hfill
  \begin{subfigure}[t]{0.35\textwidth}
    \centering
    \includegraphics[width=\linewidth]{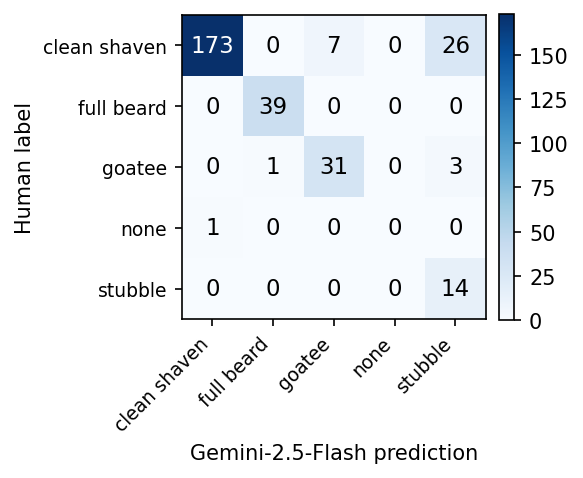}
    \caption{Beard}
    \label{fig:Beard}
  \end{subfigure}
  \vspace{0.10em}
  \begin{subfigure}[t]{0.35\textwidth}
    \centering
    \includegraphics[width=\linewidth]{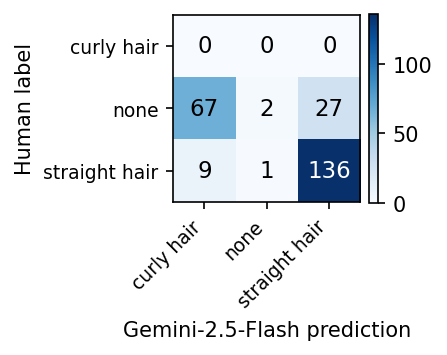}
    \caption{Hair Type}
    \label{fig:cm_hair_type}
  \end{subfigure}\hfill
  \begin{subfigure}[t]{0.35\textwidth}
    \centering
    \includegraphics[width=\linewidth]{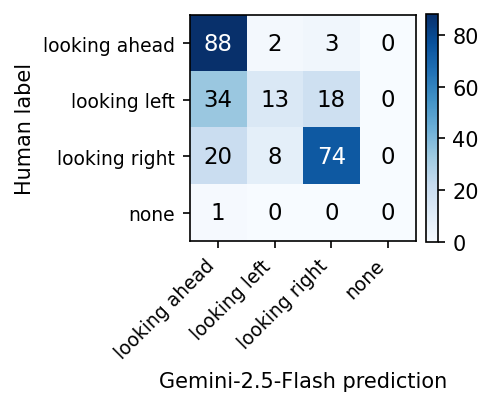}
    \caption{Head Position}
    \label{fig:cm_head_pos}
  \end{subfigure}
  \vspace{0.10em}
  \begin{subfigure}[t]{0.35\textwidth}
    \centering
    \includegraphics[width=\linewidth]{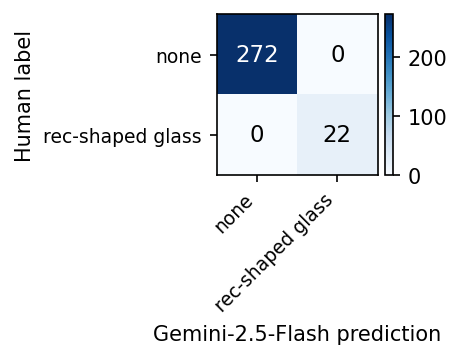}
    \caption{Eyegless}
    \label{fig:cm_eyeglasses}
  \end{subfigure}\hfill
  \begin{subfigure}[t]{0.35\textwidth}
    \centering
    \includegraphics[width=\linewidth]{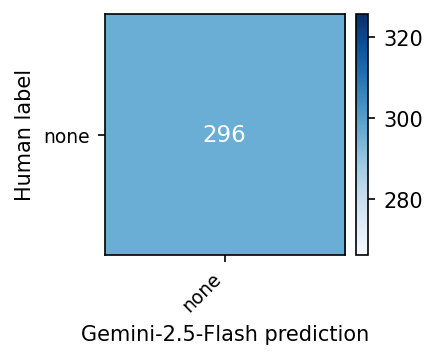}
    \caption{Hats}
    \label{fig:cm_hats}
  \end{subfigure}
  \vspace{0.10em}
  \begin{subfigure}[t]{0.35\textwidth}
    \centering
    \includegraphics[width=\linewidth]{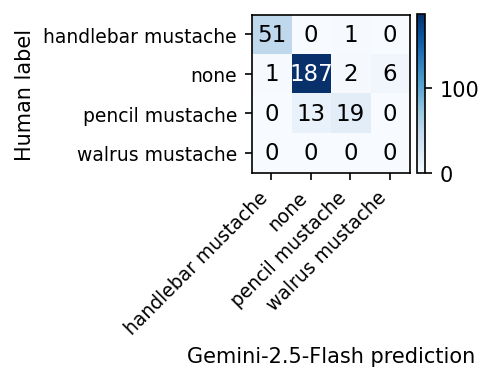}
    \caption{Mustache}
    \label{fig:cm_mustache}
  \end{subfigure}\hfill
  \begin{subfigure}[t]{0.35\textwidth}
    \centering
    \includegraphics[width=\linewidth]{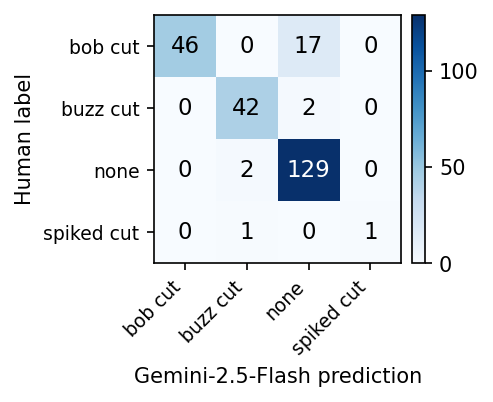}
    \caption{Hair Style}
    \label{fig:cm_hair_style}
  \end{subfigure}
  \vspace{-0.2cm}
  \caption{Confusion matrices showing the alignment between human annotators and the Gemini-2.5-Flash MLLM.}
  \label{fig:confusion_matrix_gemini}
  \vspace{-0.1cm}
\end{figure*}

\section{Human Experiment for State Tracking}
\label{app:human_exp}

To identify which model is suitable for our state dictionary, we compared three MLLMs: Gemini-2.5-Flash \cite{gemini-flash}, Gemini-2.5-Pro \cite{gemini-pro}, and GPT-4o \cite{hurst2024gpt} with the responses recorded by human annotators. We also compared CLIP (a popular text-to-image similarity model) \cite{radford2021learning}. With CLIP, for each sub-axis (e.g., hair-color), we calculated a cosine similarity score between the CLIP embeddings of the image $I_k$ and the CLIP embeddings of each category text from that sub-axis (e.g., red hair, blue hair). The category with the highest similarity score was chosen as the top candidate for that sub-axis. 

We randomly selected $12$ base images from CelebSET ($3$ each of light-skinned male, light-skinned female, dark-skinned male, and dark-skinned female faces). For each image, we ran $5$ multi-axis edit sessions using SeedEdit 3.0 \cite{wang2025seededit}, a state-of-the-art editing model from ByteDance \cite{bytedance}. We evaluated $12$ (base images) $\times$ $5$ (edit sessions) $\times$ $5$ (base image $+$ 4 edits in a session) $= 300$ images. 

For each of the $300$ images, we created a state dictionary using human annotators. We asked four annotators to mark the visible category for each of the $8$ sub-axes. For instance, if an annotator saw ``red hair'' in an image, they marked red hair in the hair-color sub-axis. Annotators marked ``None'' for a sub-axis if they did not see any category from that sub-axis. For each sub-axis, we took the majority of the four annotator votes. For instance, in an image, if three annotators marked the person's hair color as ``red hair'' and one annotator marked ``blue hair'', we assigned --- hair-color: red hair. When annotators tied on a sub-axis (e.g., $2$ votes for red hair and $2$ votes for blue hair), we considered all the top-tied candidates for that sub-axis. Overall annotators provided category annotations for $300$ (images) $\times$ $8$ (sub-axes per image) = $2400$ sub-axes.

We generated state dictionaries using four models (three MLLMs and CLIP). We reported the overall accuracy of the state dictionaries computed by each model compared with the state dictionary created via human voting. An accuracy of $50\%$ means that out of $2400$ sub-axes, the human and the model aligned on $1200$ sub-axes. We consider the human and the model aligned on a sub-axis only if all categories in that sub-axis match. 

Overall accuracy scores (in $\%$): CLIP (ViT-Base): $35.47$, Gemini-2.5-Flash: $79.65$, Gemini-2.5-Pro: $79.26$, GPT-4o: $70.31$. We used Gemini-2.5-Flash as our MLLM for state tracking because it showed the best alignment with humans. In Figure \ref{fig:confusion_matrix_gemini}, we showed the human and Gemini-2.5-Flash alignment at the attribute level using confusion matrices. In each confusion matrix, the rows represent human labels, while the columns represent Gemini-2.5-Flash predictions. For each confusion matrix, we only selected cases where both the model and humans had only one top candidate. Tied top candidates cannot be faithfully represented in a confusion matrix.

In Figure \ref{fig:confusion_matrix_gemini}, Gemini-2.5-Flash showed high human alignment ($>85\%$) in most cases. There are certain cases where alignment was low. For hair type (Figure \ref{fig:cm_hair_type}), humans marked ``none'' in $96$ cases. The model aligned in only $2$ out of these $96$ cases. This discrepancy arose from how annotators and the model perceived images. Human annotators were strict: when hair types were not strictly straight or curly (e.g., buzz cut, slightly wavy hair), humans marked ``none''. In such cases, Gemini-2.5-Flash was lenient: it preferred choosing a hair type rather than marking none. Alignment was low even for ``looking left'' (Figure \ref{fig:cm_head_pos}): $13$ out of $65$ cases. 

\clearpage

\begin{table*}[htbp]
\centering
\footnotesize
\caption{CelebA — number of edit sessions by gender across different edit types for each model}
\label{tab:celeba_num_edit_sessions}

\begin{minipage}[t]{0.48\textwidth}
\centering
\subcaption{Poses}
\vspace{0.3em}
\begin{tabular}{lcc}
\toprule
\textbf{Model} & \textbf{Male} & \textbf{Female} \\
\midrule
Flux-Dev           & 1220 & 1525 \\
Flux-Kontext-Pro   &  977 & 1220 \\
Nano Banana   & 1210 & 1500 \\
Bagel-Edit         & 1220 & 1525 \\
Qwen-Edit          & 1220 & 1520 \\
\bottomrule
\end{tabular}
\end{minipage}
\hfill
\begin{minipage}[t]{0.48\textwidth}
\centering
\subcaption{Accessories}
\vspace{0.3em}
\begin{tabular}{lcc}
\toprule
\textbf{Model} & \textbf{Male} & \textbf{Female} \\
\midrule
Flux-Dev           & 2440 & 3050 \\
Flux-Kontext-Pro   & 2440 & 3050 \\
Nano Banana   & 2234 & 2811 \\
Bagel-Edit         & 2440 & 3050 \\
Qwen-Edit          & 2440 & 3039 \\
\bottomrule
\end{tabular}
\end{minipage}

\vspace{1em}

\begin{minipage}[t]{0.48\textwidth}
\centering
\subcaption{Multi-Axis}
\vspace{0.3em}
\begin{tabular}{lcc}
\toprule
\textbf{Model} & \textbf{Male} & \textbf{Female} \\
\midrule
Flux-Dev           & 2440 & 3048 \\
Flux-Kontext-Pro   & 2440 & 3050 \\
Nano Banana   & 2349 & 2951 \\
Bagel-Edit         & 2440 & 3050 \\
Qwen-Edit          & 2440 & 3038 \\
\bottomrule
\end{tabular}
\end{minipage}
\hfill
\begin{minipage}[t]{0.48\textwidth}
\centering
\subcaption{Hair}
\vspace{0.3em}
\begin{tabular}{lcc}
\toprule
\textbf{Model} & \textbf{Male} & \textbf{Female} \\
\midrule
Flux-Dev           & 2440 & 3050 \\
Flux-Kontext-Pro   & 2440 & 3050 \\
Nano Banana   & 2363 & 2908 \\
Bagel-Edit         & 2440 & 3050 \\
Qwen-Edit          & 2440 & 3039 \\
\bottomrule
\end{tabular}
\end{minipage}
\vspace{1cm}

\end{table*}

\begin{table*}[htbp]
\centering
\footnotesize
\caption{CelebA — number of edit sessions by age across different edit types for each model}
\label{tab:celeba_num_edit_sessions_age}

\begin{minipage}[t]{0.48\textwidth}
\centering
\subcaption{Poses}
\vspace{0.3em}
\begin{tabular}{lcc}
\toprule
\textbf{Model} & \textbf{Young } & \textbf{Old} \\
\midrule
Flux-Dev           & 2065 & 680 \\
Flux-Kontext-Pro   &  1652 & 545 \\
Nano Banana   & 2037 & 673 \\
Bagel-Edit         & 2065 & 680 \\
Qwen-Edit          & 2060 & 680 \\
\bottomrule
\end{tabular}
\end{minipage}
\hfill
\begin{minipage}[t]{0.48\textwidth}
\centering
\subcaption{Accessories}
\vspace{0.3em}
\begin{tabular}{lcc}
\toprule
\textbf{Model} & \textbf{Young} & \textbf{Old} \\
\midrule
Flux-Dev           & 4130 & 1360 \\
Flux-Kontext-Pro   & 4130 & 1360 \\
Nano Banana   & 3795 & 1250 \\
Bagel-Edit         & 4130 & 1360 \\
Qwen-Edit          & 4119 & 1360 \\
\bottomrule
\end{tabular}
\end{minipage}

\vspace{1em}

\begin{minipage}[t]{0.48\textwidth}
\centering
\subcaption{Multi-Axis}
\vspace{0.3em}
\begin{tabular}{lcc}
\toprule
\textbf{Model} & \textbf{Young} & \textbf{Old} \\
\midrule
Flux-Dev           & 4128 & 1360 \\
Flux-Kontext-Pro   & 4130 & 1360 \\
Nano Banana   & 3981 & 1319 \\
Bagel-Edit         & 4130 & 1360 \\
Qwen-Edit          & 4119 & 1359 \\
\bottomrule
\end{tabular}
\end{minipage}
\hfill
\begin{minipage}[t]{0.48\textwidth}
\centering
\subcaption{Hair}
\vspace{0.3em}
\begin{tabular}{lcc}
\toprule
\textbf{Model} & \textbf{Young} & \textbf{Old} \\
\midrule
Flux-Dev           & 4130 & 1360 \\
Flux-Kontext-Pro   & 4130 & 1360 \\
Nano Banana   & 3960 & 1311 \\
Bagel-Edit         & 4130 & 1360 \\
Qwen-Edit          & 4120 & 1359 \\
\bottomrule
\end{tabular}
\end{minipage}
\vspace{1cm}

\end{table*}

\begin{table*}[htbp]
\centering
\footnotesize
\caption{CelebSET — number of edit sessions by gender and skin-tone across different edit types for each model.}
\label{tab:celebset_num_edit_sessions}

\begin{subtable}[t]{0.48\textwidth}
\centering
\caption{Poses}
\begin{tabular}{lcccc}
\toprule
\textbf{Model} &
\textbf{\makecell{Male\\Light}} &
\textbf{\makecell{Male\\Dark}} &
\textbf{\makecell{Female\\Light}} &
\textbf{\makecell{Female\\Dark}} \\
\midrule
Flux-Dev           & 1110 & 475 & 805 & 195 \\
Flux-Kontext-Pro   &  890 & 380 & 644 & 156 \\
Nano Banana   & 1110 & 475 & 805 & 195 \\
Bagel-Edit         & 1110 & 475 & 805 & 195 \\
Qwen-Edit          & 1110 & 475 & 803 & 195 \\
SeedEdit 3.0        & 1110 & 475 & 798 & 195 \\
\bottomrule
\end{tabular}
\end{subtable}
\hfill
\begin{subtable}[t]{0.48\textwidth}
\centering
\caption{Hair}
\begin{tabular}{lcccc}
\toprule
\textbf{Model} &
\textbf{\makecell{Male\\Light}} &
\textbf{\makecell{Male\\Dark}} &
\textbf{\makecell{Female\\Light}} &
\textbf{\makecell{Female\\Dark}} \\
\midrule
Flux-Dev           & 2220 & 950 & 1610 & 390 \\
Flux-Kontext-Pro   & 2220 & 950 & 1610 & 390 \\
Nano Banana   & 2220 & 949 & 1607 & 390 \\
Bagel-Edit         & 2220 & 950 & 1610 & 390 \\
Qwen-Edit          & 2220 & 950 & 1609 & 390 \\
SeedEdit 3.0       & 2220 & 950 & 1606 & 390 \\
\bottomrule
\end{tabular}
\end{subtable}

\vspace{1em}

\begin{subtable}[t]{0.48\textwidth}
\centering
\caption{Accessories}
\begin{tabular}{lcccc}
\toprule
\textbf{Model} &
\textbf{\makecell{Male\\Light}} &
\textbf{\makecell{Male\\Dark}} &
\textbf{\makecell{Female\\Light}} &
\textbf{\makecell{Female\\Dark}} \\
\midrule
Flux-Dev           & 2220 & 950 & 1610 & 390 \\
Flux-Kontext-Pro   & 2220 & 950 & 1610 & 390 \\
Nano Banana   & 2207 & 945 & 1603 & 389 \\
Bagel-Edit         & 2220 & 950 & 1610 & 390 \\
Qwen-Edit          & 2220 & 950 & 1610 & 390 \\
SeedEdit 3.0        & 2220 & 950 & 1608 & 390 \\
\bottomrule
\end{tabular}
\end{subtable}
\hfill
\begin{subtable}[t]{0.48\textwidth}
\centering
\caption{Multi-Axis}
\begin{tabular}{lcccc}
\toprule
\textbf{Model} &
\textbf{\makecell{Male\\Light}} &
\textbf{\makecell{Male\\Dark}} &
\textbf{\makecell{Female\\Light}} &
\textbf{\makecell{Female\\Dark}} \\
\midrule
Flux-Dev           & 2220 & 950 & 1610 & 390 \\
Flux-Kontext-Pro   & 2220 & 950 & 1610 & 390 \\
Nano Banana   & 2201 & 949 & 1607 & 390 \\
Bagel-Edit         & 2220 & 950 & 1610 & 390 \\
Qwen-Edit          & 2220 & 950 & 1610 & 389 \\
SeedEdit 3.0       & 2191 & 950 & 1605 & 339 \\
\bottomrule
\end{tabular}
\end{subtable}

\end{table*}

\begin{table*}[htbp]
\centering
\footnotesize
\caption{CelebSET — number of edit sessions by age across different edit types for each model}
\label{tab:celebset_num_edit_sessions_age}

\begin{minipage}[t]{0.48\textwidth}
\centering
\subcaption{Poses}
\vspace{0.3em}
\begin{tabular}{lcc}
\toprule
\textbf{Model} & \textbf{Young } & \textbf{Old} \\
\midrule
Flux-Dev           & 490 & 205 \\
Flux-Kontext-Pro   &  392 & 165 \\
Nano Banana   & 490 & 205 \\
Bagel-Edit         & 490 & 205 \\
Qwen-Edit          & 488 & 205 \\
SeedEdit 3.0 & 483 & 205 \\
\bottomrule
\end{tabular}
\end{minipage}
\hfill
\begin{minipage}[t]{0.48\textwidth}
\centering
\subcaption{Accessories}
\vspace{0.3em}
\begin{tabular}{lcc}
\toprule
\textbf{Model} & \textbf{Young} & \textbf{Old} \\
\midrule
Flux-Dev           & 980 & 410 \\
Flux-Kontext-Pro   & 980 & 410 \\
Nano Banana   & 974 & 410 \\
Bagel-Edit         & 980 & 410 \\
Qwen-Edit          & 980 & 410 \\
SeedEdit 3.0 & 978 & 410 \\
\bottomrule
\end{tabular}
\end{minipage}

\vspace{1em}

\begin{minipage}[t]{0.48\textwidth}
\centering
\subcaption{Multi-Axis}
\vspace{0.3em}
\begin{tabular}{lcc}
\toprule
\textbf{Model} & \textbf{Young} & \textbf{Old} \\
\midrule
Flux-Dev           & 980 & 410 \\
Flux-Kontext-Pro   & 980 & 410 \\
Nano Banana   & 979 & 406 \\
Bagel-Edit         & 980 & 410 \\
Qwen-Edit          & 980 & 410 \\
SeedEdit 3.0 & 956 & 410 \\
\bottomrule
\end{tabular}
\end{minipage}
\hfill
\begin{minipage}[t]{0.48\textwidth}
\centering
\subcaption{Hair}
\vspace{0.3em}
\begin{tabular}{lcc}
\toprule
\textbf{Model} & \textbf{Young} & \textbf{Old} \\
\midrule
Flux-Dev           & 980 & 410 \\
Flux-Kontext-Pro   & 980 & 410 \\
Nano Banana   & 979 & 410 \\
Bagel-Edit         & 980 & 410 \\
Qwen-Edit          & 979 & 410 \\
SeedEdit 3.0 & 977 & 410 \\
\bottomrule
\end{tabular}
\end{minipage}
\vspace{1cm}

\end{table*}

\clearpage

\section{Number of Edit Sessions}
\label{app:edit_sessions}

We reported the number of editing sessions per model with CelebA by gender in Table \ref{tab:celeba_num_edit_sessions} and by age in Table \ref{tab:celeba_num_edit_sessions_age}. For CelebSET, we reported the number of editing sessions by gender and skin-tone in Table \ref{tab:celebset_num_edit_sessions} and by age in Table \ref{tab:celebset_num_edit_sessions_age}.

\section{Explanation for WD threshold}
\label{app:wd_thresh_selection}
\begin{figure}[h]
  \centering
  \vspace{0.2cm}
  \includegraphics[width=\linewidth]{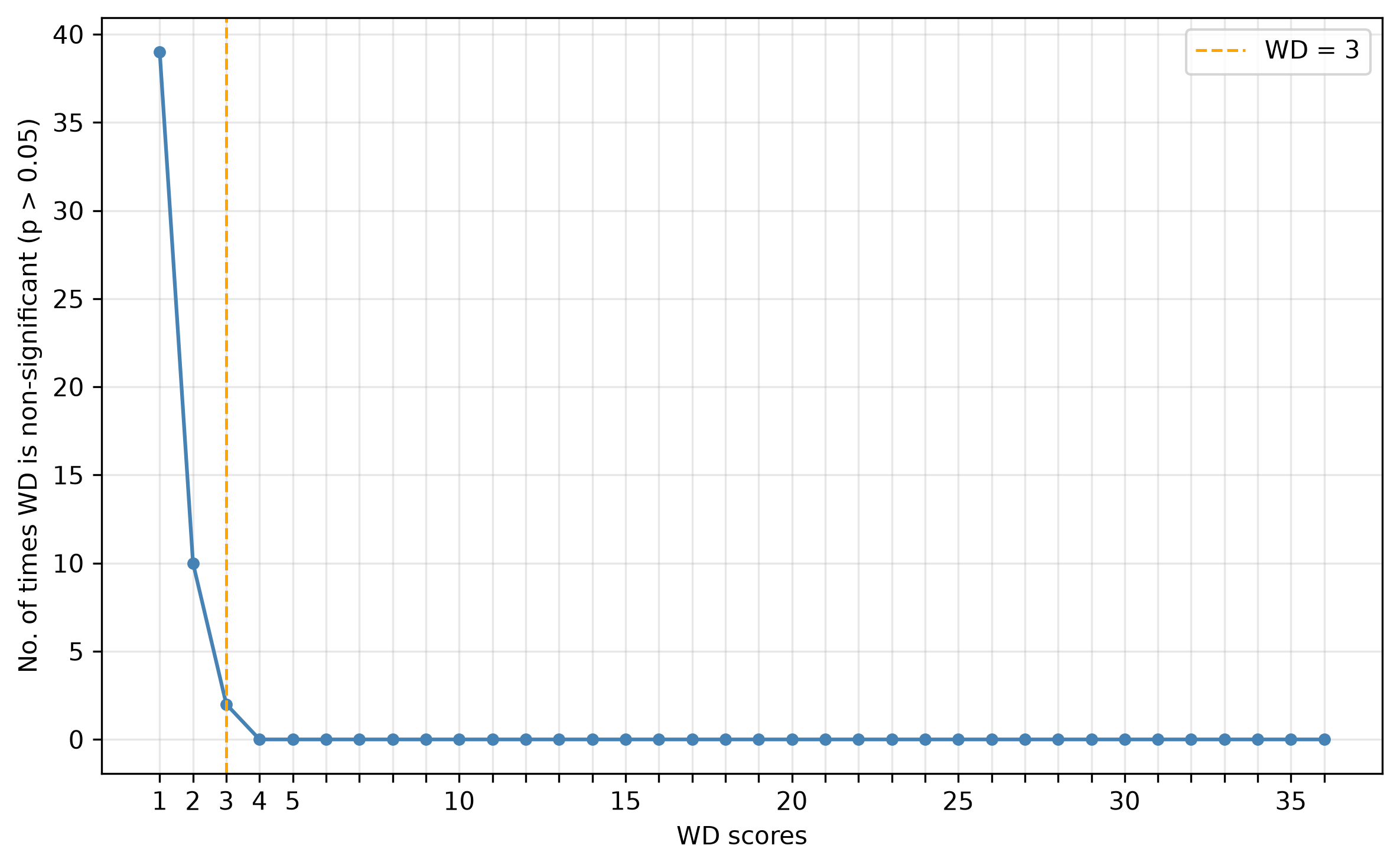}
  \vspace{-0.5cm}
  \caption{Non-significant p-val vs. WD scores: once WD > 3, almost all p-values are statistically significant.}
  \label{fig:wd_threshold}
  \vspace{-0.7cm}
\end{figure}

We chose WD $>3$ as our threshold for prominent bias, based on statistical significance. In Figure \ref{fig:wd_threshold}, the x-axis represents the WD scores across all our demographic bias analyses. The y-axis represents the number of times a WD score was not statistically significant (p > 0.05). At WD $>3$, almost all WD values were statistically significant, making WD $> 3$ an ideal bias threshold.

\section{Age Bias}
\label{app:age_bias}

For identity preservation, we showed age bias results for CelebA in Table \ref{tab:dropfss_celeba_age} and for CelebSET in Table \ref{tab:dropfss_celebset_age}. For Tables  \ref{tab:dropfss_celeba_age} and \ref{tab:dropfss_celebset_age}, in each ``Mean'' column, the left value indicates the worst performing group (higher $\operatorname{Drop}_{\mathrm{FSS}}$) and the right value indicates the best group (note that we have only two groups for age in both datasets).

\textbf{Most models (except Qwen-Edit) showed poor identity preservation on older faces}. In CelebSET (Table \ref{tab:dropfss_celebset_age}), we observed an age bias ($WD_{age}$ $>3$) in several models across different test sets. In all these models (except Qwen-Edit), the worst-performing group was older faces. In Qwen-Edit, the worst-performing group was always younger faces. In CelebA (Table \ref{tab:dropfss_celeba_age}), we did not observe many instances of age bias. 

For editing alignment, we showed age bias results for CelebA in Table \ref{tab:sc_celeba_age} and for CelebSET in Table \ref{tab:sc_celebset_age}. 

\textbf{Most models showed minimal (or no) age bias in both datasets}. In CelebA (Table \ref{tab:sc_celeba_age}), for Flux-Dev, we observed an age bias across three test-sets --- in all these cases, the worst performing group was older faces. We did not observe a similar trend with Flux-Dev for CelebSET (Table \ref{tab:sc_celebset_age}). For other models, we did not observe clear age bias trends across both datasets.

For Overediting, we showed age bias results for CelebA in Table \ref{tab:oer_celeba_age} and for CelebSET in Table \ref{tab:oer_celebset_age}. 

\textbf{Several models created more overedits on older faces.} In both CelebA and CelebSET, we observed age bias in all models for most test sets. In almost all these cases, the worst performing group (higher OER) was older faces.

\section{All mean values}
\label{sec:all_mean}
For CelebA and CelebSET, we reported mean values (with $95\%$ confidence intervals) for all demographic groups using bar plots. For identity preservation ($\operatorname{Drop}_{\mathrm{FSS}}$), we showed mean values for male and female gender groups in CelebA in Figure \ref{fig:face_id_drop_celeba}. We showed mean values for all four gender-skin-tone intersectional groups (male-light, male-dark, female-light, female-dark) in CelebSET in Figure \ref{fig:face_id_drop_celebset}. We reported mean values on our age groups (young and old) for CelebA in Figure \ref{fig:face_id_drop_celeba_age} and for CelebSET in Figure \ref{fig:face_id_drop_celebset_age}. In each figure, we showed separate bar plots for each test set.

For alignment ($SC$), we showed mean values for male and female gender groups in CelebA in Figure \ref{fig:sem_con_celeba}. We showed mean values for all four gender-skin-tone intersectional groups in CelebSET in Figure \ref{fig:sem_con_celebset}. We reported mean values on our age groups (young and old) for CelebA in Figure \ref{fig:sem_con_celeba_age} and for CelebSET in Figure \ref{fig:sem_con_celebset_age}. 

For overediting (OER), we showed mean values for male and female gender groups in CelebA in Figure \ref{fig:oer_celeba}. We showed mean values for all four gender-skin-tone intersectional groups in CelebSET in Figure \ref{fig:oer_celebset}. We reported mean values on our age groups (young and old) for CelebA in Figure \ref{fig:oer_celeba_age} and for CelebSET in Figure \ref{fig:oer_celebset_age}.

\begin{table*}[t]
\centering
\caption{$\operatorname{Drop}_{\mathrm{FSS}}$ \% (lower is better) disparity on CelebA. For each model and test set, we reported mean $\operatorname{Drop}_{\mathrm{FSS}}$ for Young (Y) and Old (O) groups. In each mean column, the left value is the mean of the worst-performing group, and the right value is the mean of the best-performing group. $WD_{age}$ denotes the Wasserstein distance between the young and old $\operatorname{Drop}_{\mathrm{FSS}}$ distributions. $*$ marks non-significant values ($p > 0.05$). WD $> 3$ is highlighted as it indicates gender bias.}
\vspace{-0.35cm}
\label{tab:dropfss_celeba_age}

\setlength{\tabcolsep}{4.2pt}
\renewcommand{\arraystretch}{1.05}

\definecolor{wdlight}{RGB}{252,228,214}
\definecolor{wddark}{RGB}{244,177,131}

\newcommand{\wdlightcell}[1]{\cellcolor{wdlight}{#1}}
\newcommand{\wddarkcell}[1]{\cellcolor{wddark}{#1}}

\scriptsize
\begin{tabular}{lcccccccc}
\toprule
&
\multicolumn{2}{c}{Accessories$\downarrow$} &
\multicolumn{2}{c}{Hair$\downarrow$} &
\multicolumn{2}{c}{Pose$\downarrow$} &
\multicolumn{2}{c}{Multi-axis$\downarrow$} \\
\cmidrule(lr){2-3}
\cmidrule(lr){4-5}
\cmidrule(lr){6-7}
\cmidrule(lr){8-9}
Model &
Mean & $WD_{age}$ &
Mean & $WD_{age}$ &
Mean & $WD_{age}$ &
Mean & $WD_{age}$ \\
\midrule

Flux-Dev &
O:90.2 \quad Y:89.4 & 0.8 &
O:90.3 \quad Y:88.7 & 1.6 &
O:90.3 \quad Y:89.4 & 0.9 &
O:90.3 \quad Y:89.1 & 1.2 \\

Kontext-Pro &
Y:26.0 \quad O:23.5 & 2.4 &
O:25.6 \quad Y:24.2 & 1.6 &
Y:38.3 \quad O:34.9 & \textbf{3.4} &
Y:32.7 \quad O:32.3 & 0.8 \\

Nano Banana &
Y:34.0 \quad O:32.0 & 2.0 &
O:23.3 \quad Y:23.0 & 0.5 &
Y:32.5 \quad O:32.1 & 0.7$^{*}$ &
Y:32.0 \quad O:32.0 & 0.5 \\

Bagel-Edit &
Y:27.9 \quad O:25.7 & 2.3 &
O:19.3 \quad Y:18.6 & 1.0 &
Y:54.1 \quad O:49.6 & \textbf{4.5} &
Y:34.0 \quad O:32.1 & 2.0 \\

Qwen-Edit &
Y:28.1 \quad O:27.4 & 0.7 &
Y:28.0 \quad O:28.0 & 1.7 &
Y:48.4 \quad O:45.1 & \textbf{3.4} &
O:35.5 \quad Y:35.0 & 1.4 \\

\bottomrule
\end{tabular}

\vspace{-0.25cm}
\end{table*}

\begin{table*}[t]
\centering
\caption{$\operatorname{Drop}_{\mathrm{FSS}}$ \% (lower is better) disparity on CelebSET. We reported the mean $\operatorname{Drop}_{\mathrm{FSS}}$ for Young (Y) and Old (O) groups (left mean value: worst group; right mean value: best group). $WD_{age}$ denotes the Wasserstein distance between the young and old $\operatorname{Drop}_{\mathrm{FSS}}$ distributions. $*$ marks non-significant values ($p > 0.05$). WD $> 3$ is highlighted.}
\vspace{-0.35cm}
\label{tab:dropfss_celebset_age}

\setlength{\tabcolsep}{4.2pt}
\renewcommand{\arraystretch}{1.05}

\definecolor{wdlight}{RGB}{252,228,214}
\definecolor{wddark}{RGB}{244,177,131}

\newcommand{\wdlightcell}[1]{\cellcolor{wdlight}{#1}}
\newcommand{\wddarkcell}[1]{\cellcolor{wddark}{#1}}

\scriptsize
\begin{tabular}{lcccccccc}
\toprule
&
\multicolumn{2}{c}{Accessories$\downarrow$} &
\multicolumn{2}{c}{Hair$\downarrow$} &
\multicolumn{2}{c}{Pose$\downarrow$} &
\multicolumn{2}{c}{Multi-axis$\downarrow$} \\
\cmidrule(lr){2-3}
\cmidrule(lr){4-5}
\cmidrule(lr){6-7}
\cmidrule(lr){8-9}
Model &
Mean & $WD_{age}$ &
Mean & $WD_{age}$ &
Mean & $WD_{age}$ &
Mean & $WD_{age}$ \\
\midrule

Flux-Dev &
O:84.3 \quad Y:83.4 & 1.5 &
O:83.8 \quad Y:82.5 & 2.0 &
Y:83.2 \quad O:82.8 & 1.9 &
O:83.3 \quad Y:82.8 & 1.2 \\

Kontext-Pro &
O:27.9 \quad Y:26.4 & 2.4 &
O:21.3 \quad Y:21.3 & 2.7 &
Y:31.6 \quad O:29.2 & 1.7 &
O:30.7 \quad Y:29.5 & \textbf{3.4} \\

Nano Banana &
O:27.1 \quad Y:25.3 & \textbf{3.8} &
O:17.7 \quad Y:15.6 & 2.3 &
O:24.4 \quad Y:21.6 & \textbf{3.2} &
O:25.1 \quad Y:23.1 & \textbf{3.0} \\

Bagel-Edit &
O:28.9 \quad Y:26.8 & 2.3 &
O:19.9 \quad Y:17.3 & 2.8 &
Y:56.9 \quad O:55.3 & 2.7 &
O:34.8 \quad Y:33.8 & 2.7 \\

Qwen-Edit &
Y:48.7 \quad O:44.0 & \textbf{5.1} &
Y:49.1 \quad O:46.6 & \textbf{3.1} &
Y:41.8 \quad O:37.7 & \textbf{3.4} &
Y:46.5 \quad O:43.6 & \textbf{4.2} \\

SeedEdit 3.0 &
O:27.1 \quad Y:24.6 & \textbf{3.3} &
O:21.9 \quad Y:17.1 & \textbf{4.9} &
Y:46.7 \quad O:45.2 & 2.7 &
O:33.6 \quad Y:31.2 & 1.5 \\

\bottomrule
\end{tabular}
\end{table*}

\begin{table*}[t]
\centering
\caption{$SC$ (higher is better) disparity on CelebA. We reported the mean $SC$ for Young (Y) and Old (O) groups (left mean value: worst group; right mean value: best group). $WD_{age}$ denotes the Wasserstein distance between the young and old $SC$ distributions. $*$ marks non-significant values ($p > 0.05$). WD $> 3$ is highlighted. Values scaled by $100$.}
\vspace{-0.35cm}
\label{tab:sc_celeba_age}

\setlength{\tabcolsep}{4.2pt}
\renewcommand{\arraystretch}{1.05}

\definecolor{wdlight}{RGB}{252,228,214}
\definecolor{wddark}{RGB}{244,177,131}

\newcommand{\wdlightcell}[1]{\cellcolor{wdlight}{#1}}
\newcommand{\wddarkcell}[1]{\cellcolor{wddark}{#1}}

\scriptsize
\begin{tabular}{lcccccccc}
\toprule
&
\multicolumn{2}{c}{Accessories$\uparrow$} &
\multicolumn{2}{c}{Hair$\uparrow$} &
\multicolumn{2}{c}{Pose$\uparrow$} &
\multicolumn{2}{c}{Multi-axis$\uparrow$} \\
\cmidrule(lr){2-3}
\cmidrule(lr){4-5}
\cmidrule(lr){6-7}
\cmidrule(lr){8-9}
Model &
Mean & $WD_{age}$ &
Mean & $WD_{age}$ &
Mean & $WD_{age}$ &
Mean & $WD_{age}$ \\
\midrule

Flux-Dev &
O:29.7 \quad Y:35.1 & \textbf{5.4} &
O:26.9 \quad Y:31.4 & \textbf{4.5} &
O:31.0 \quad Y:31.5 & 0.9$^{*}$ &
O:27.4 \quad Y:31.2 & \textbf{3.8} \\

Kontext-Pro &
Y:73.4 \quad O:73.5 & 1.0$^{*}$ &
Y:75.9 \quad O:76.9 & 1.1 &
O:46.6 \quad Y:47.7 & 1.5$^{*}$ &
Y:63.9 \quad O:64.3 & 0.6$^{*}$ \\

Nano Banana &
O:82.8 \quad Y:85.4 & 2.6 &
O:86.9 \quad Y:88.2 & 1.3 &
O:62.1 \quad Y:62.8 & 0.9$^{*}$ &
O:73.8 \quad Y:75.5 & 1.7 \\

Bagel-Edit &
O:87.1 \quad Y:88.0 & 0.9 &
O:80.2 \quad Y:80.3 & 0.3$^{*}$ &
O:56.8 \quad Y:60.5 & \textbf{3.7} &
O:68.6 \quad Y:69.0 & 0.8$^{*}$ \\

Qwen-Edit &
O:79.6 \quad Y:80.0 & 0.7$^{*}$ &
Y:76.5 \quad O:77.0 & 0.7$^{*}$ &
O:65.8 \quad Y:68.1 & 2.3 &
O:63.8 \quad Y:65.1 & 1.2 \\

\bottomrule
\end{tabular}
\end{table*}

\begin{table*}[t]
\centering
\caption{$SC$ (higher is better) disparity on CelebSET. We reported the mean $SC$ for Young (Y) and Old (O) groups (left mean value: worst group; right mean value: best group). $WD_{age}$ denotes the Wasserstein distance between the young and old $SC$ distributions. $*$ marks non-significant values ($p > 0.05$). WD $> 3$ is highlighted. Values scaled by $100$.}
\vspace{-0.35cm}
\label{tab:sc_celebset_age}

\setlength{\tabcolsep}{4.2pt}
\renewcommand{\arraystretch}{1.05}

\definecolor{wdlight}{RGB}{252,228,214}
\definecolor{wddark}{RGB}{244,177,131}

\newcommand{\wdlightcell}[1]{\cellcolor{wdlight}{#1}}
\newcommand{\wddarkcell}[1]{\cellcolor{wddark}{#1}}

\scriptsize
\begin{tabular}{lcccccccc}
\toprule
&
\multicolumn{2}{c}{Accessories$\uparrow$} &
\multicolumn{2}{c}{Hair$\uparrow$} &
\multicolumn{2}{c}{Pose$\uparrow$} &
\multicolumn{2}{c}{Multi-axis$\uparrow$} \\
\cmidrule(lr){2-3}
\cmidrule(lr){4-5}
\cmidrule(lr){6-7}
\cmidrule(lr){8-9}
Model &
Mean & $WD_{age}$ &
Mean & $WD_{age}$ &
Mean & $WD_{age}$ &
Mean & $WD_{age}$ \\
\midrule

Flux-Dev &
O:26.4 \quad Y:30.3 & \textbf{3.9} &
O:27.3 \quad Y:28.8 & 1.6$^{*}$ &
O:28.5 \quad Y:29.0 & 1.4$^{*}$ &
O:26.4 \quad Y:27.4 & 1.3$^{*}$ \\

Kontext-Pro &
Y:72.5 \quad O:73.5 & 1.6$^{*}$ &
O:74.5 \quad Y:75.0 & 0.9$^{*}$ &
O:35.4 \quad Y:40.5 & \textbf{5.1} &
O:59.9 \quad Y:61.2 & 1.5$^{*}$ \\

Nano Banana &
Y:82.1 \quad O:84.7 & 2.6 &
O:84.6 \quad Y:86.8 & 2.2 &
O:58.7 \quad Y:59.5 & 1.1$^{*}$ &
O:72.1 \quad Y:73.4 & 1.5$^{*}$ \\

Bagel-Edit &
Y:86.2 \quad O:88.5 & 2.2 &
Y:77.4 \quad O:78.3 & 1.1$^{*}$ &
Y:59.9 \quad O:61.1 & 1.5$^{*}$ &
O:67.5 \quad Y:68.4 & 1.3$^{*}$ \\

Qwen-Edit &
O:81.9 \quad Y:82.1 & 0.8$^{*}$ &
Y:77.2 \quad O:78.7 & 1.5$^{*}$ &
O:60.8 \quad Y:66.1 & \textbf{5.4} &
O:65.0 \quad Y:66.4 & 1.7$^{*}$ \\

SeedEdit 3.0 &
Y:86.0 \quad O:87.1 & 1.1$^{*}$ &
O:80.9 \quad Y:82.2 & 1.5$^{*}$ &
O:63.7 \quad Y:64.2 & 2.9$^{*}$ &
Y:69.1 \quad O:69.5 & 1.3$^{*}$ \\

\bottomrule
\end{tabular}
\end{table*}

\begin{table*}[t]
\centering
\caption{OER (lower is better) disparity on CelebA. We reported the mean OER for the Young (Y) and Old (O) groups (left mean value: worst group; right mean value: best group). $WD_{age}$ denotes the Wasserstein distance between the young and old OER distributions. $*$ marks non-significant values ($p > 0.05$). WD $> 3$ is highlighted. Values scaled by $100$.}
\vspace{-0.35cm}
\label{tab:oer_celeba_age}

\setlength{\tabcolsep}{4.2pt}
\renewcommand{\arraystretch}{1.05}

\definecolor{wdlight}{RGB}{252,228,214}
\definecolor{wddark}{RGB}{244,177,131}

\newcommand{\wdlightcell}[1]{\cellcolor{wdlight}{#1}}
\newcommand{\wddarkcell}[1]{\cellcolor{wddark}{#1}}

\scriptsize
\begin{tabular}{lcccccccc}
\toprule
&
\multicolumn{2}{c}{Accessories$\downarrow$} &
\multicolumn{2}{c}{Hair$\downarrow$} &
\multicolumn{2}{c}{Pose$\downarrow$} &
\multicolumn{2}{c}{Multi-axis$\downarrow$} \\
\cmidrule(lr){2-3}
\cmidrule(lr){4-5}
\cmidrule(lr){6-7}
\cmidrule(lr){8-9}
Model &
Mean & $WD_{age}$ &
Mean & $WD_{age}$ &
Mean & $WD_{age}$ &
Mean & $WD_{age}$ \\
\midrule

Flux-Dev &
O:78.6 \quad Y:65.0 & \textbf{13.6} &
O:78.8 \quad Y:62.3 & \textbf{16.5} &
O:59.6 \quad Y:43.9 & \textbf{15.7} &
O:79.0 \quad Y:60.8 & \textbf{18.3} \\

Kontext-Pro &
O:14.6 \quad Y:10.3 & \textbf{4.3} &
O:24.8 \quad Y:17.8 & \textbf{7.0} &
O:18.9 \quad Y:15.7 & \textbf{3.4} &
O:28.0 \quad Y:21.1 & \textbf{6.9} \\

Nano Banana &
O:14.1 \quad Y:9.2 & \textbf{4.9} &
O:20.0 \quad Y:13.4 & \textbf{6.7} &
O:13.5 \quad Y:9.1 & \textbf{4.5} &
O:23.6 \quad Y:15.6 & \textbf{8.0} \\

Bagel-Edit &
O:11.1 \quad Y:7.6 & \textbf{3.7} &
O:24.4 \quad Y:16.5 & \textbf{7.9} &
O:16.3 \quad Y:13.0 & \textbf{3.6} &
O:25.9 \quad Y:19.4 & \textbf{6.5} \\

Qwen-Edit &
O:16.6 \quad Y:12.7 & \textbf{3.9} &
O:30.0 \quad Y:23.2 & \textbf{6.9} &
O:20.4 \quad Y:19.7 & \textbf{1.0$^{*}$} &
O:30.9 \quad Y:26.1 & \textbf{4.8} \\

\bottomrule
\end{tabular}
\end{table*}

\begin{table*}[t]
\centering
\caption{OER (lower is better) disparity on CelebSET. We reported the mean OER for Young (age: $20-30$) and Old (age: $50-60$) groups (left mean value: worst group; right mean value: best group). $WD_{age}$ denotes the Wasserstein distance between the young and old OER distributions. $*$ marks non-significant values ($p > 0.05$). WD $> 3$ is highlighted. Values scaled by $100$.}
\vspace{-0.35cm}
\label{tab:oer_celebset_age}

\setlength{\tabcolsep}{4.2pt}
\renewcommand{\arraystretch}{1.05}

\definecolor{wdlight}{RGB}{252,228,214}
\definecolor{wddark}{RGB}{244,177,131}

\newcommand{\wdlightcell}[1]{\cellcolor{wdlight}{#1}}
\newcommand{\wddarkcell}[1]{\cellcolor{wddark}{#1}}

\scriptsize
\begin{tabular}{lcccccccc}
\toprule
&
\multicolumn{2}{c}{Accessories$\downarrow$} &
\multicolumn{2}{c}{Hair$\downarrow$} &
\multicolumn{2}{c}{Pose$\downarrow$} &
\multicolumn{2}{c}{Multi-axis$\downarrow$} \\
\cmidrule(lr){2-3}
\cmidrule(lr){4-5}
\cmidrule(lr){6-7}
\cmidrule(lr){8-9}
Model &
Mean & $WD_{age}$ &
Mean & $WD_{age}$ &
Mean & $WD_{age}$ &
Mean & $WD_{age}$ \\
\midrule

Flux-Dev &
O:81.3 \quad Y:58.6 & \textbf{22.9} &
O:80.6 \quad Y:58.5 & \textbf{22.1} &
O:62.6 \quad Y:37.3 & \textbf{25.3} &
O:80.1 \quad Y:56.5 & \textbf{23.8} \\

Kontext-Pro &
O:14.8 \quad Y:13.0 & \textbf{3.1} &
O:26.3 \quad Y:20.9 & \textbf{5.7} &
O:14.7 \quad Y:13.7 & 2.4$^{*}$ &
O:29.9 \quad Y:22.8 & \textbf{7.3} \\

Nano Banana &
Y:21.6 \quad O:15.4 & \textbf{7.4} &
O:21.8 \quad Y:12.5 & \textbf{9.2} &
O:9.2 \quad Y:6.9 & 2.5$^{*}$ &
O:29.5 \quad Y:24.1 & \textbf{10.1} \\

Bagel-Edit &
Y:12.7 \quad O:11.8 & \textbf{5.1} &
O:26.2 \quad Y:25.9 & \textbf{7.7} &
O:13.1 \quad Y:12.5 & 2.3$^{*}$ &
O:34.5 \quad Y:20.1 & \textbf{14.1} \\

Qwen-Edit &
O:25.0 \quad Y:21.9 & 3.1$^{*}$ &
O:36.0 \quad Y:29.9 & \textbf{6.4} &
O:23.4 \quad Y:13.5 & \textbf{9.9} &
O:39.3 \quad Y:29.6 & \textbf{9.8} \\

SeedEdit 3.0 &
O:10.9 \quad Y:7.5 & \textbf{3.4} &
O:25.7 \quad Y:15.8 & \textbf{9.9} &
O:15.4 \quad Y:14.3 & 2.9$^{*}$ &
O:25.3 \quad Y:26.9 & \textbf{5.7} \\

\bottomrule
\end{tabular}
\end{table*}

\newcommand{\legbox}[1]{%
  \raisebox{0.2ex}{\textcolor{#1}{\rule{0.55em}{0.55em}}}%
}

\newcommand{\legendmale}{\legbox{blue}}
\newcommand{\legendfemale}{\legbox{orange}}

\newcommand{\legendml}{\legbox{blue!40}}    
\newcommand{\legendmd}{\legbox{blue!80}}    
\newcommand{\legendfl}{\legbox{orange!40}}  
\newcommand{\legendfd}{\legbox{orange!80}}  

\newcommand{\legendyoung}{\legbox{green}}
\newcommand{\legendold}{\legbox{yellow}}

\begin{figure*}[h]
  \centering
  \quad \legendmale\ Male\;\; \legendfemale\ Female \par
  \vspace{0.4em}

  \captionsetup[subfigure]{margin={2.5em,0pt}}

  \begin{subfigure}[t]{0.49\textwidth}
    \centering
    \includegraphics[width=\linewidth]{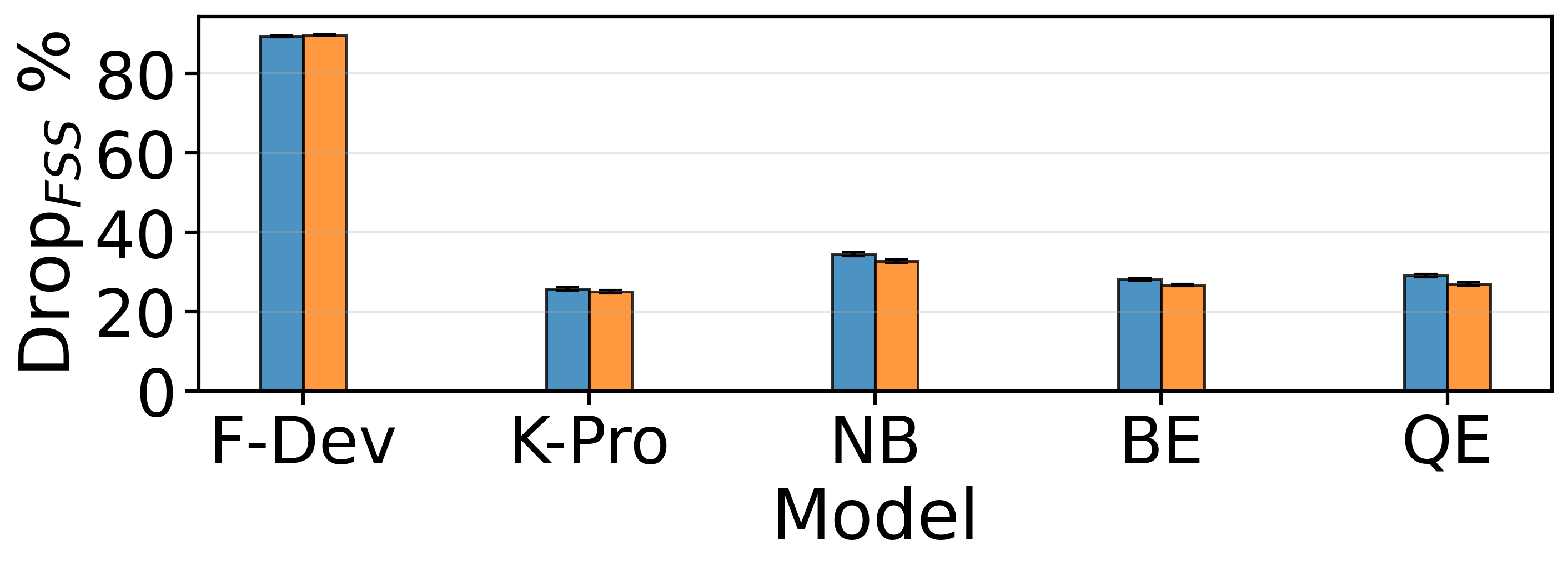}
    \caption{Accessories}
  \end{subfigure}\hfill
  \begin{subfigure}[t]{0.49\textwidth}
    \centering
    \includegraphics[width=\linewidth]{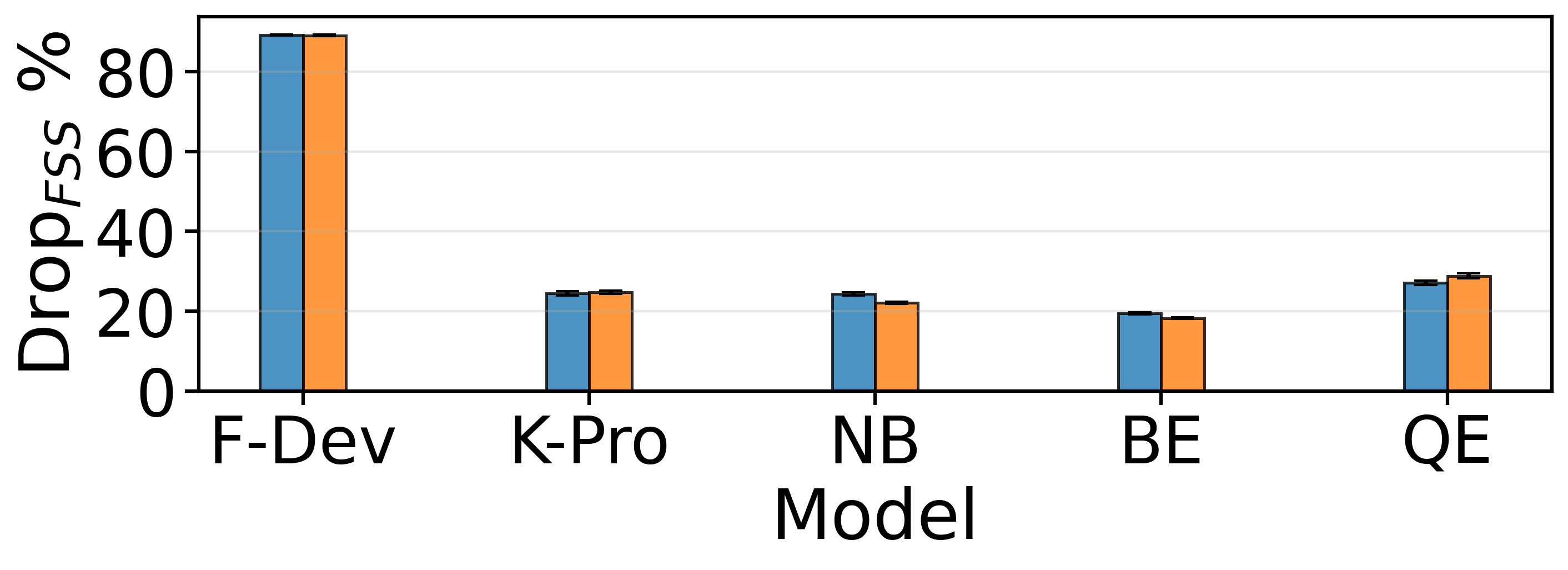}
    \caption{Hair}
  \end{subfigure}

  \vspace{0.10em}

  \begin{subfigure}[t]{0.49\textwidth}
    \centering
    \includegraphics[width=\linewidth]{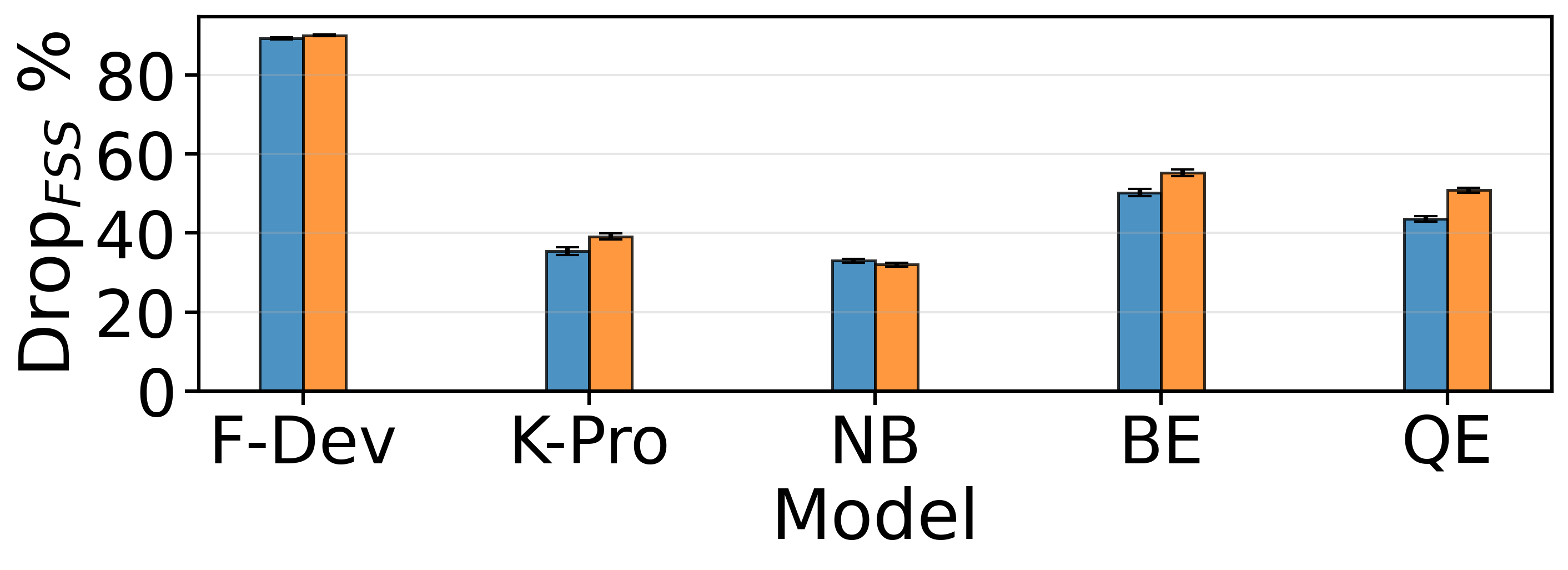}
    \caption{Pose}
  \end{subfigure}\hfill
  \begin{subfigure}[t]{0.49\textwidth}
    \centering
    \includegraphics[width=\linewidth]{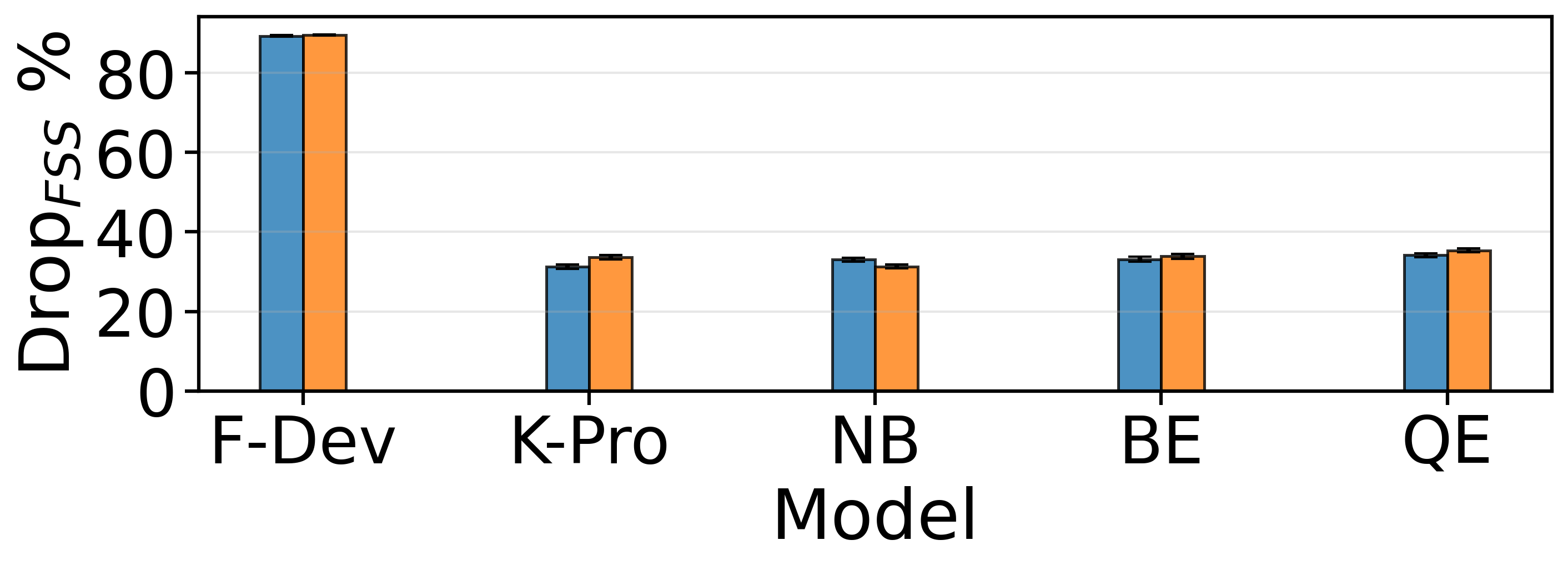}
    \caption{Multi-axis}
  \end{subfigure}

  \vspace{-0.2cm}
  \caption{$\operatorname{Drop}_{\mathrm{FSS}}$ $\%$ (lower is better) across gender subgroups in CelebA. Flux-Dev (F-Dev), Kontext-Pro (K-Pro), Nano Banana (NB), Bagel-Edit (BE), and Qwen-Edit (QE).}
  \label{fig:face_id_drop_celeba}
  \vspace{-0.1cm}
\end{figure*}

\begin{figure*}[h]
  \centering
  \quad
      \legendml\ Male-Light\;\;
      \legendmd\ Male-Dark\;\;
      \legendfl\ Female-Light\;\;
      \legendfd\ Female-Dark\par
  \vspace{0.4em}

  \captionsetup[subfigure]{margin={2.5em,0pt}}

  \begin{subfigure}[t]{0.49\textwidth}
    \centering
    \includegraphics[width=\linewidth]{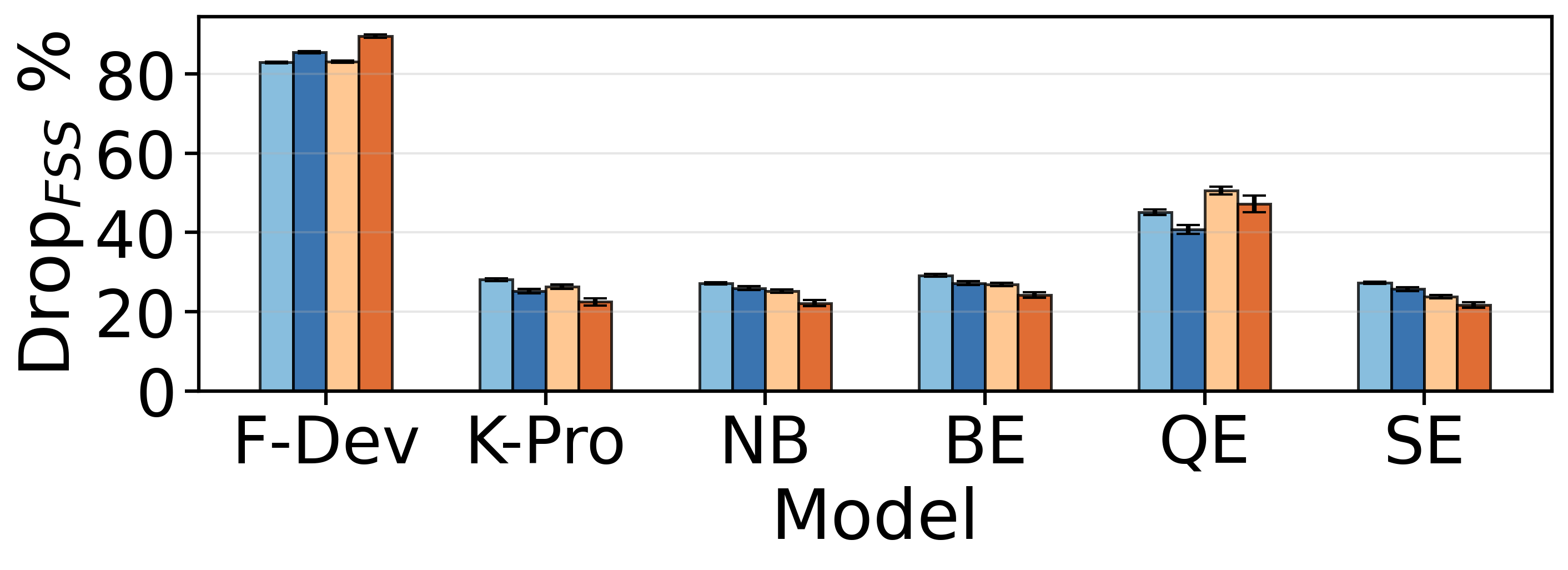}
    \caption{Accessories}
  \end{subfigure}\hfill
  \begin{subfigure}[t]{0.49\textwidth}
    \centering
    \includegraphics[width=\linewidth]{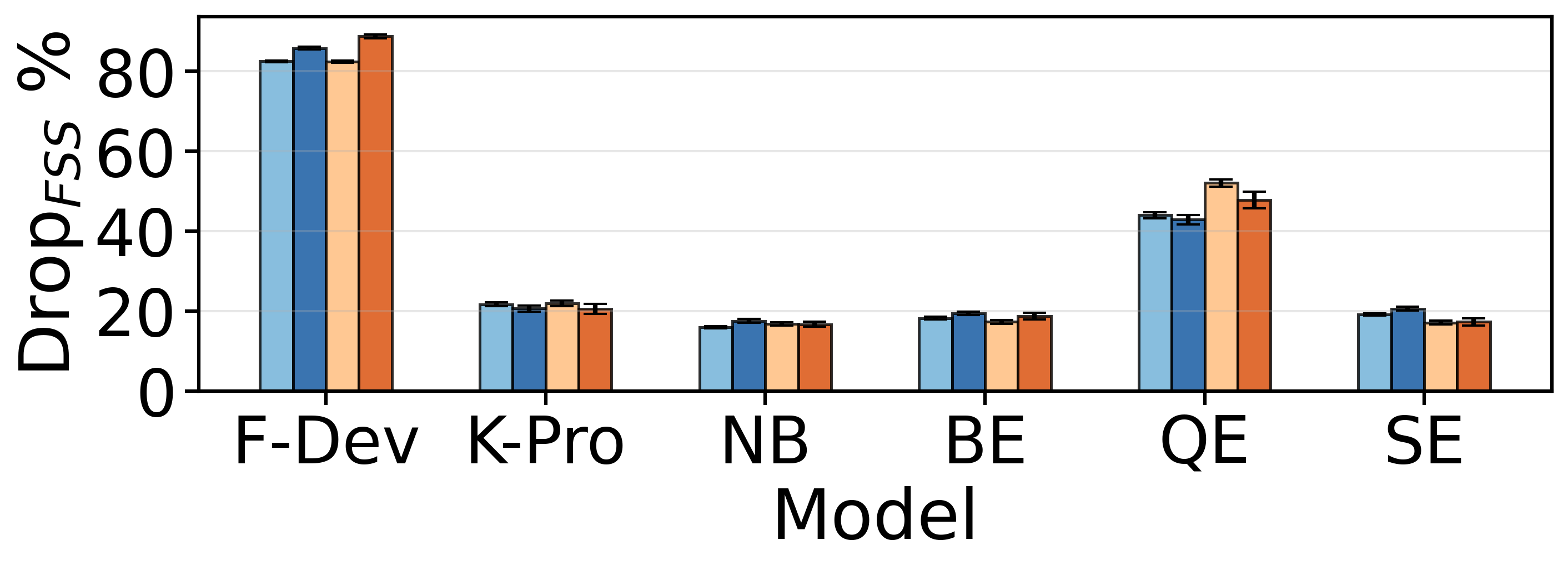}
    \caption{Hair}
  \end{subfigure}

  \vspace{0.10em}

  \begin{subfigure}[t]{0.49\textwidth}
    \centering
    \includegraphics[width=\linewidth]{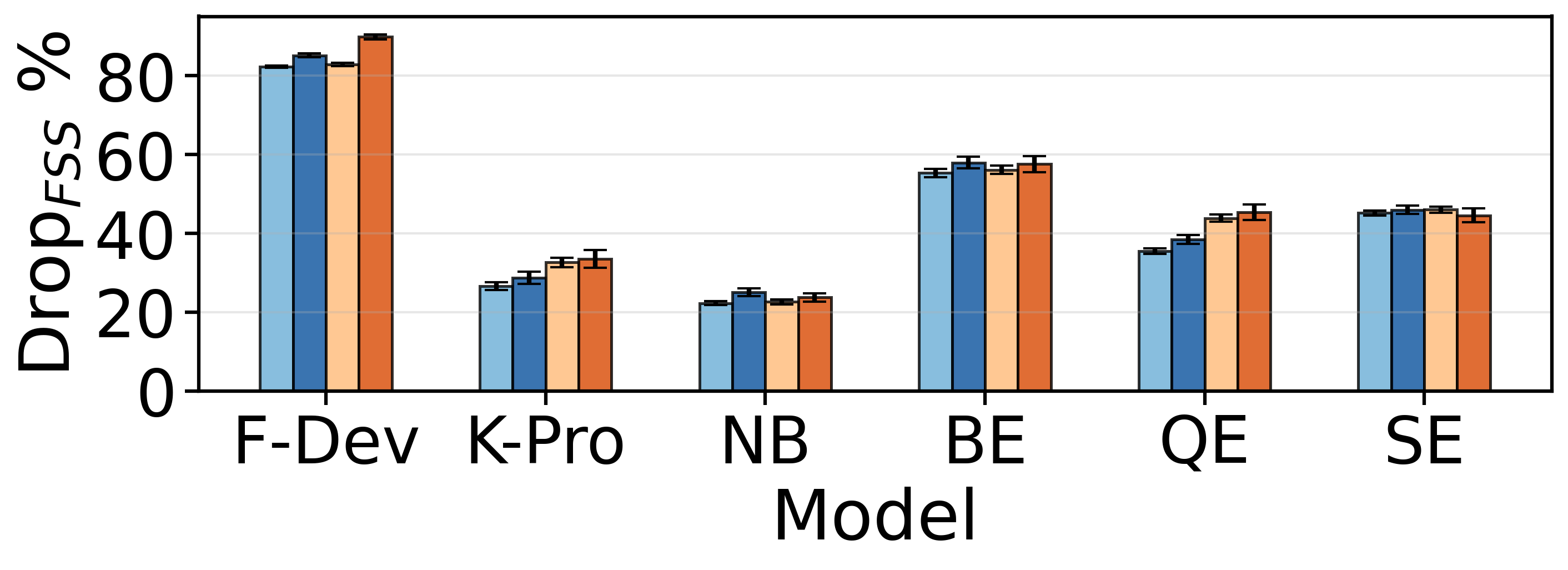}
    \caption{Pose}
  \end{subfigure}\hfill
  \begin{subfigure}[t]{0.49\textwidth}
    \centering
    \includegraphics[width=\linewidth]{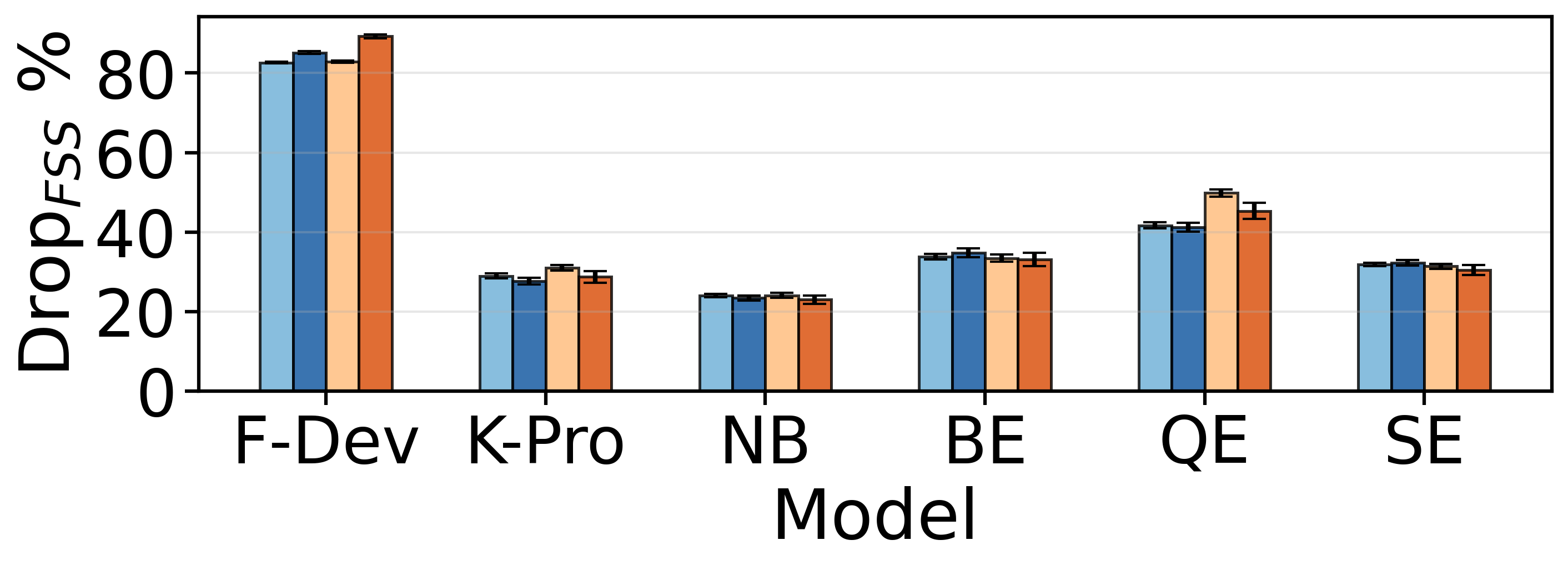}
    \caption{Multi-axis}
  \end{subfigure}

  \vspace{-0.2cm}
  \caption{$\operatorname{Drop}_{\mathrm{FSS}}$ $\%$ (lower is better) across intersectional demographic subgroups in CelebSET.}
  \label{fig:face_id_drop_celebset}
  \vspace{-0.1cm}
\end{figure*}

\begin{figure*}[h]
  \centering
  \quad \legendyoung\ Young\;\; \legendold\ Old \par
  \vspace{0.4em}

  \captionsetup[subfigure]{margin={2.5em,0pt}}

  \begin{subfigure}[t]{0.49\textwidth}
    \centering
    \includegraphics[width=\linewidth]{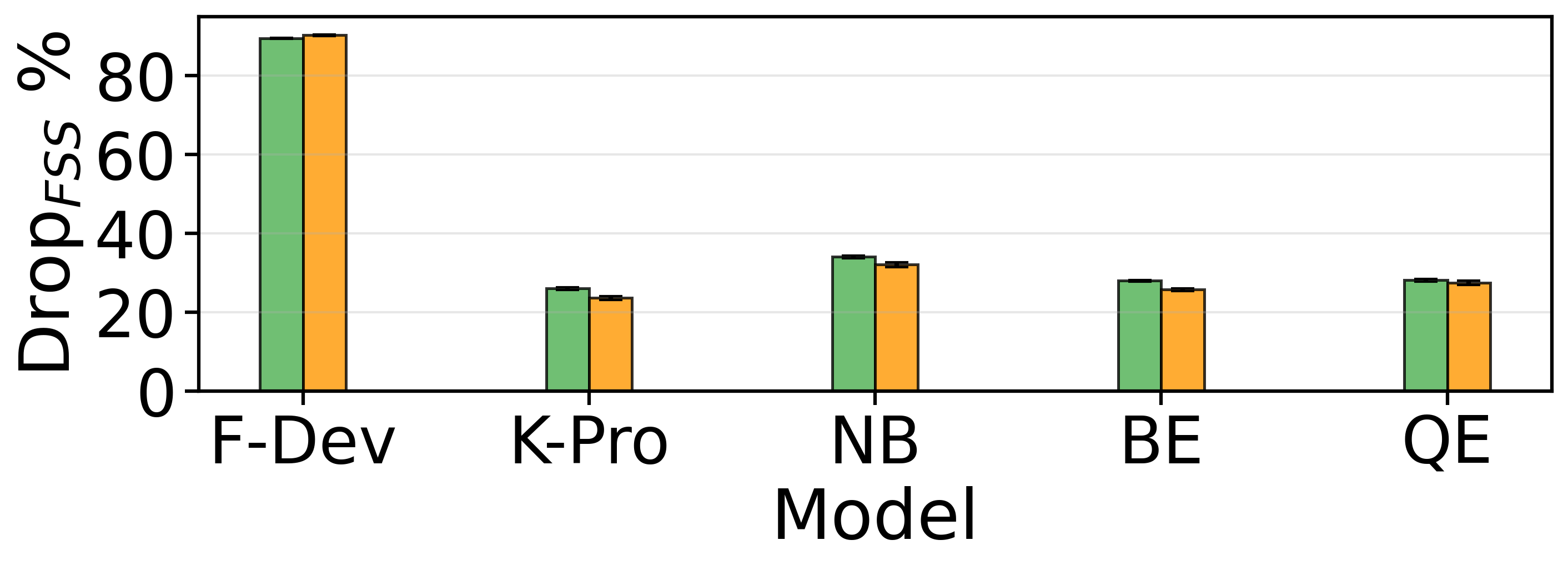}
    \caption{Accessories}
  \end{subfigure}\hfill
  \begin{subfigure}[t]{0.49\textwidth}
    \centering
    \includegraphics[width=\linewidth]{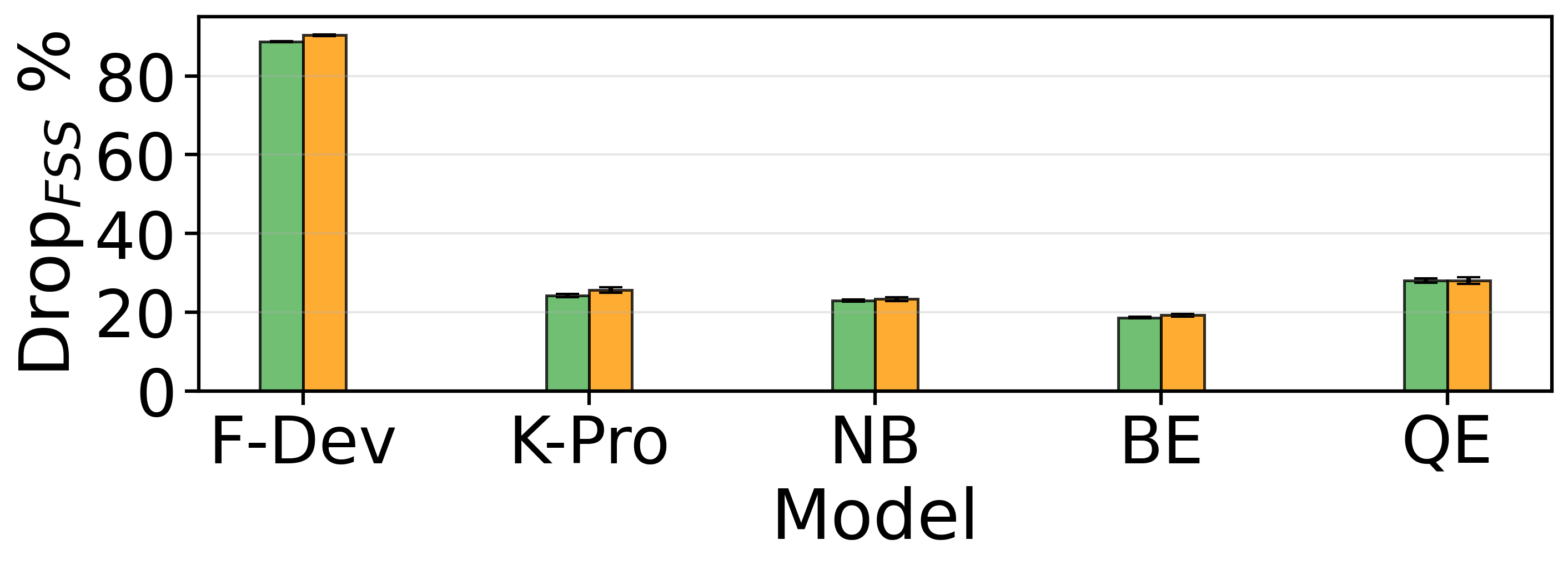}
    \caption{Hair}
  \end{subfigure}

  \vspace{0.10em}

  \begin{subfigure}[t]{0.49\textwidth}
    \centering
    \includegraphics[width=\linewidth]{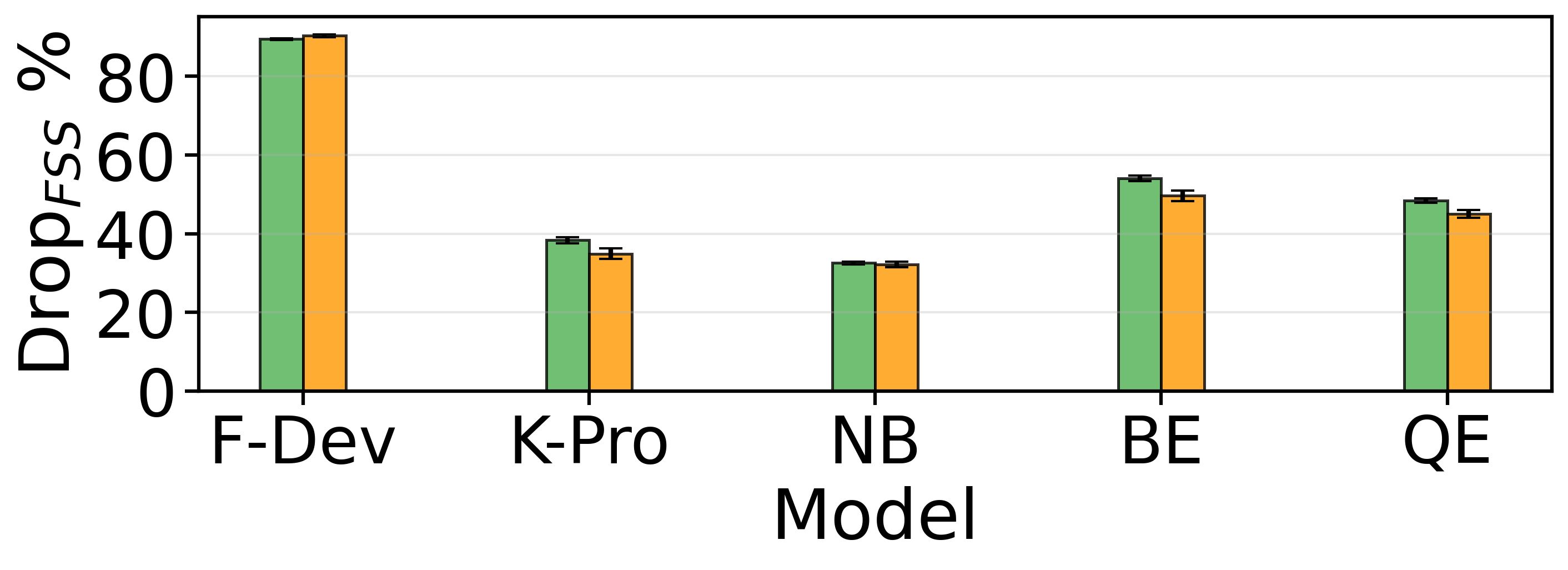}
    \caption{Pose}
  \end{subfigure}\hfill
  \begin{subfigure}[t]{0.49\textwidth}
    \centering
    \includegraphics[width=\linewidth]{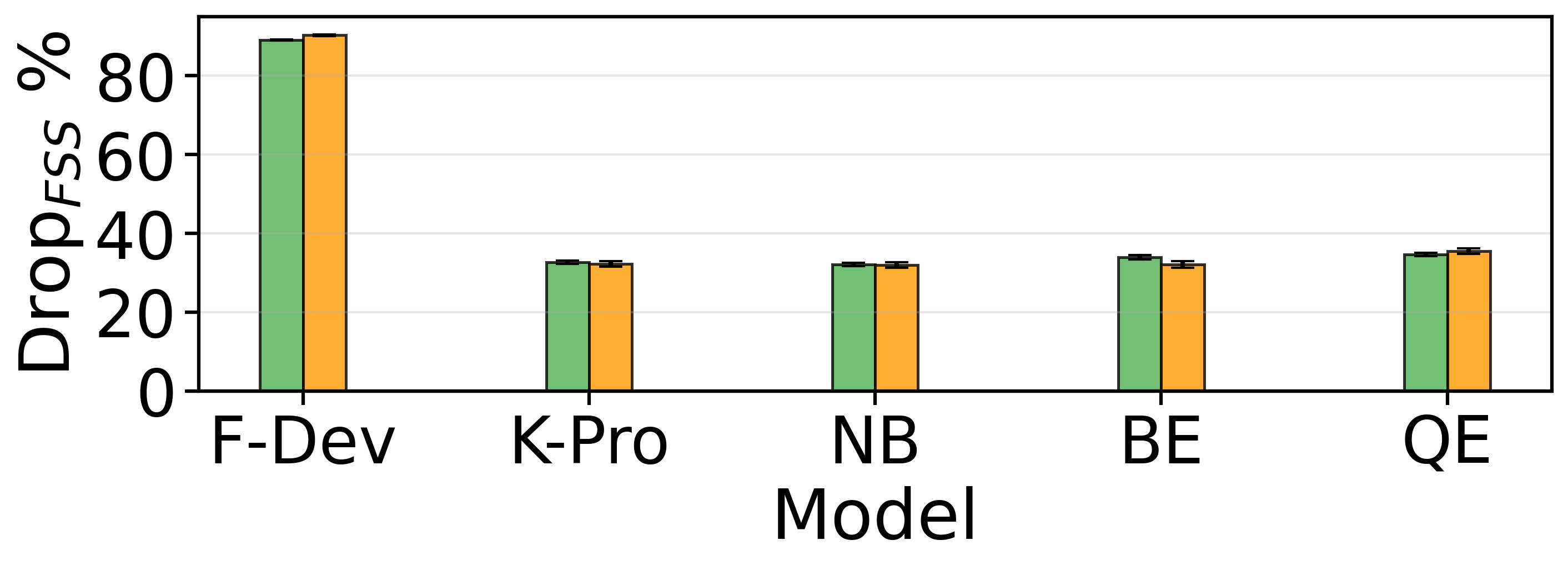}
    \caption{Multi-axis}
  \end{subfigure}

  \vspace{-0.2cm}
  \caption{$\operatorname{Drop}_{\mathrm{FSS}}$ $\%$ (lower is better) across age subgroups in CelebA.}
  \label{fig:face_id_drop_celeba_age}
  \vspace{-0.1cm}
\end{figure*}

\begin{figure*}[h]
  \centering
  \quad
      \legendyoung\ Young\;\;
      \legendold\ Old\;\;
      
  \vspace{0.4em}

  \captionsetup[subfigure]{margin={2.5em,0pt}}

  \begin{subfigure}[t]{0.49\textwidth}
    \centering
    \includegraphics[width=\linewidth]{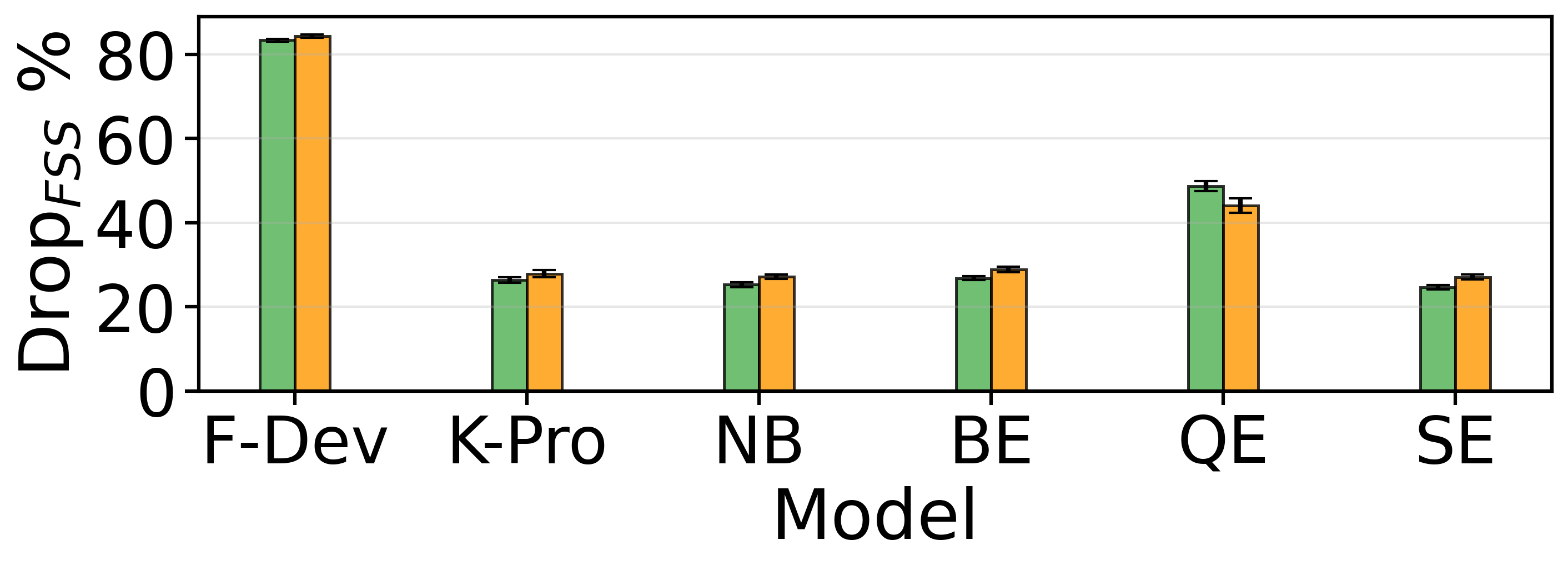}
    \caption{Accessories}
  \end{subfigure}\hfill
  \begin{subfigure}[t]{0.49\textwidth}
    \centering
    \includegraphics[width=\linewidth]{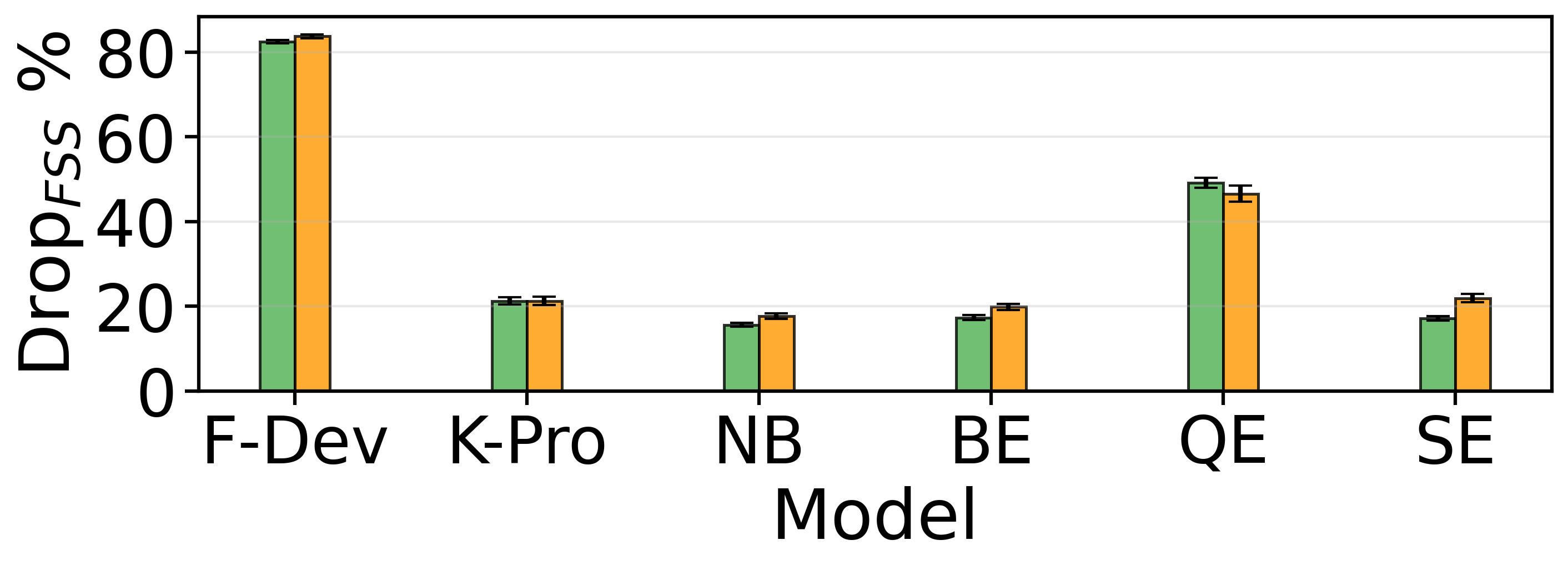}
    \caption{Hair}
  \end{subfigure}

  \vspace{0.10em}

  \begin{subfigure}[t]{0.49\textwidth}
    \centering
    \includegraphics[width=\linewidth]{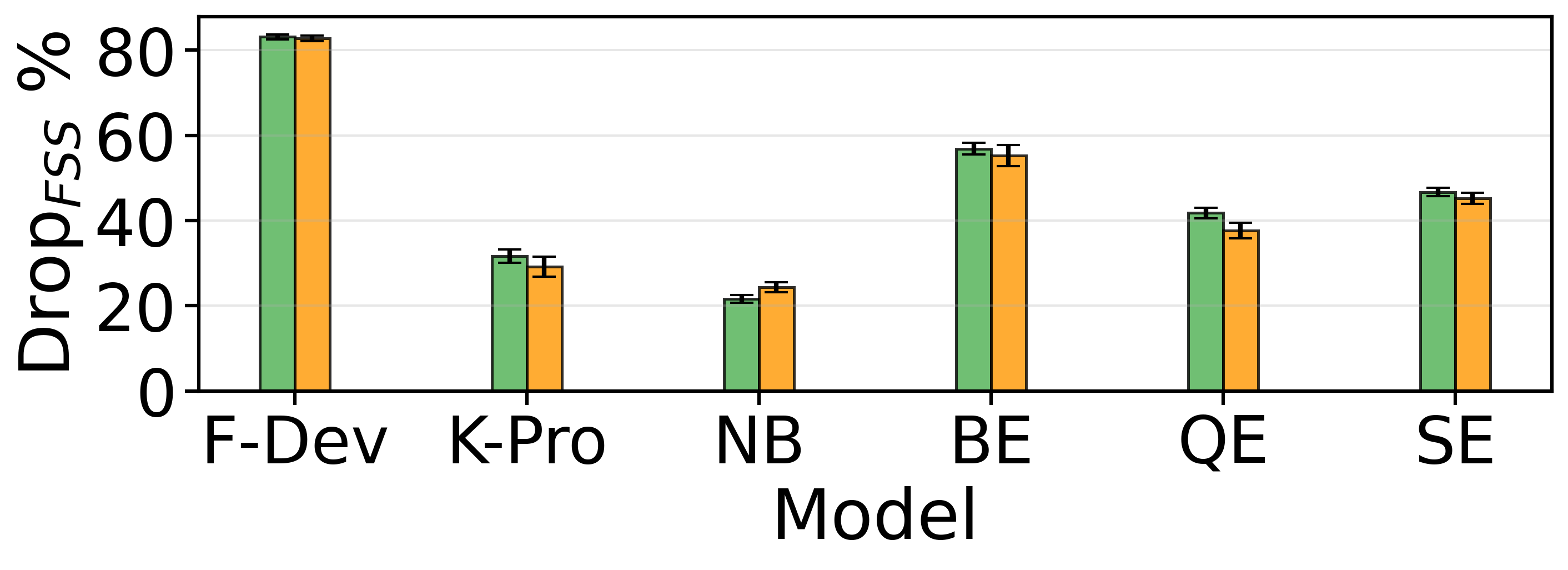}
    \caption{Pose}
  \end{subfigure}\hfill
  \begin{subfigure}[t]{0.49\textwidth}
    \centering
    \includegraphics[width=\linewidth]{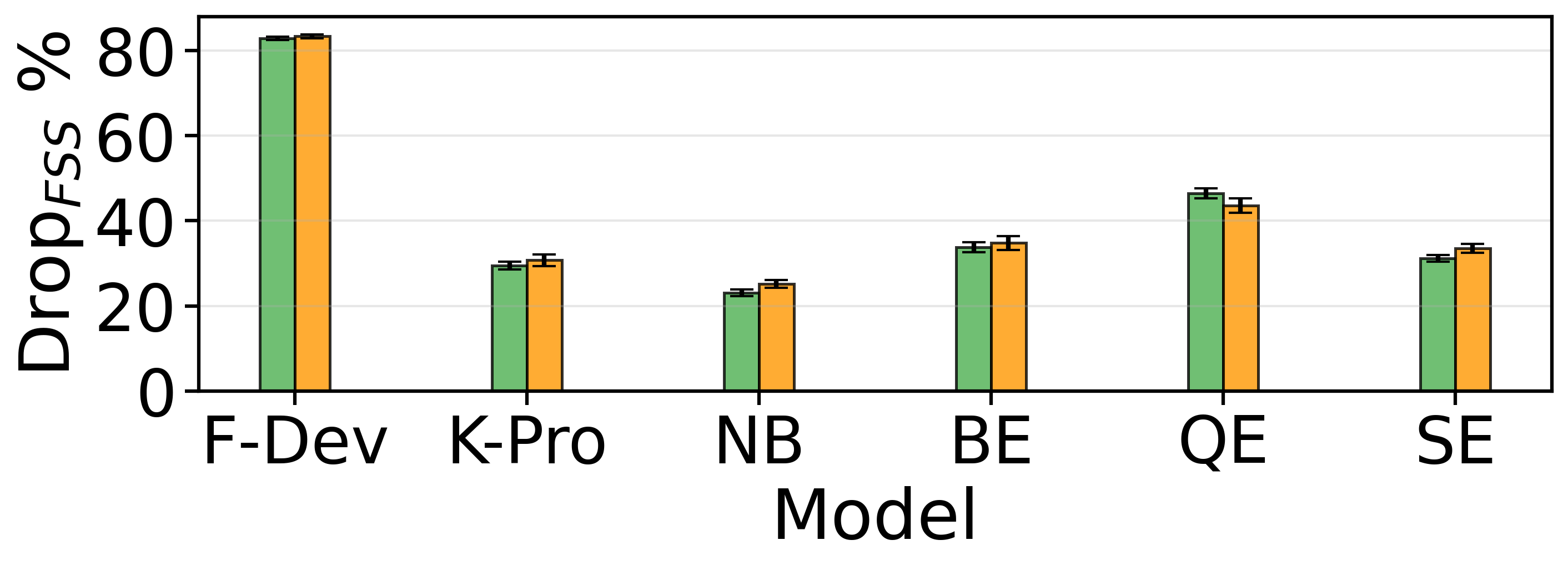}
    \caption{Multi-axis}
  \end{subfigure}

  \vspace{-0.2cm}
  \caption{$\operatorname{Drop}_{\mathrm{FSS}}$ $\%$ (lower is better) across age subgroups in CelebSET.}
  \label{fig:face_id_drop_celebset_age}
  \vspace{-0.1cm}
\end{figure*}

\begin{figure*}[h]
  \centering
  \quad \legendmale\ Male\;\; \legendfemale\ Female \par
  \vspace{0.4em}

  \captionsetup[subfigure]{margin={2.5em,0pt}}

  \begin{subfigure}[t]{0.49\textwidth}
    \centering
    \includegraphics[width=\linewidth]{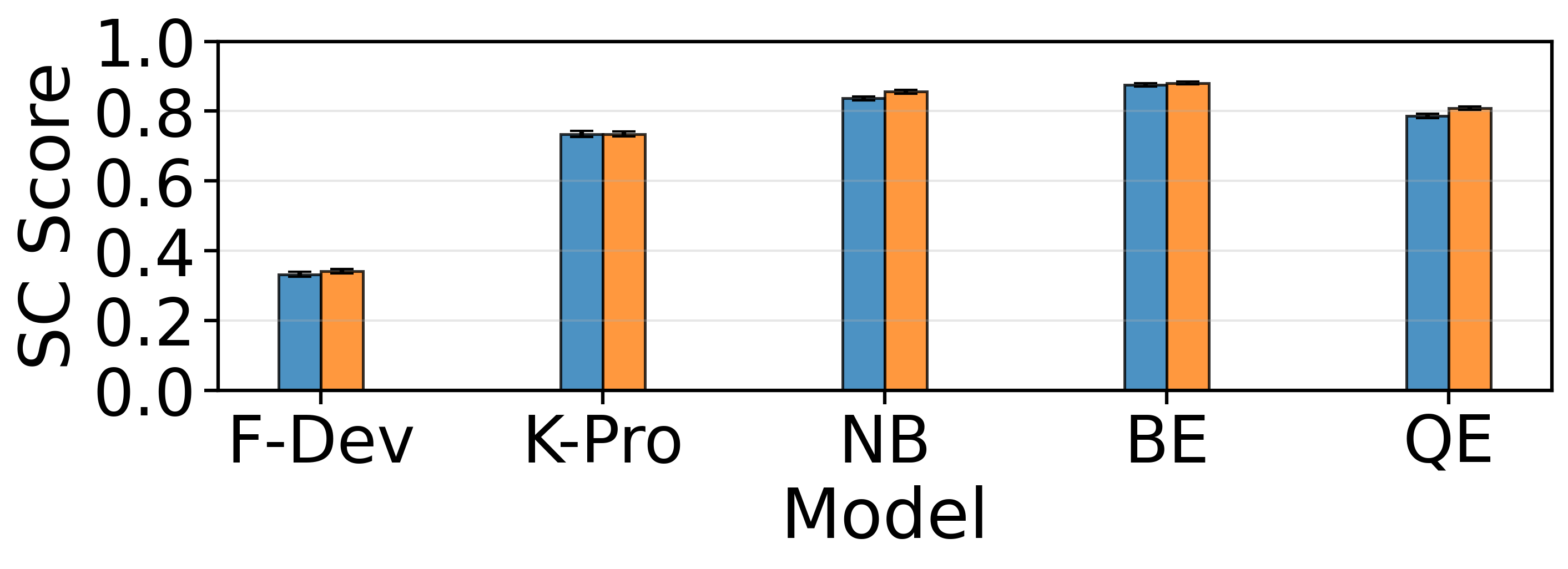}
    \caption{Accessories}
  \end{subfigure}\hfill
  \begin{subfigure}[t]{0.49\textwidth}
    \centering
    \includegraphics[width=\linewidth]{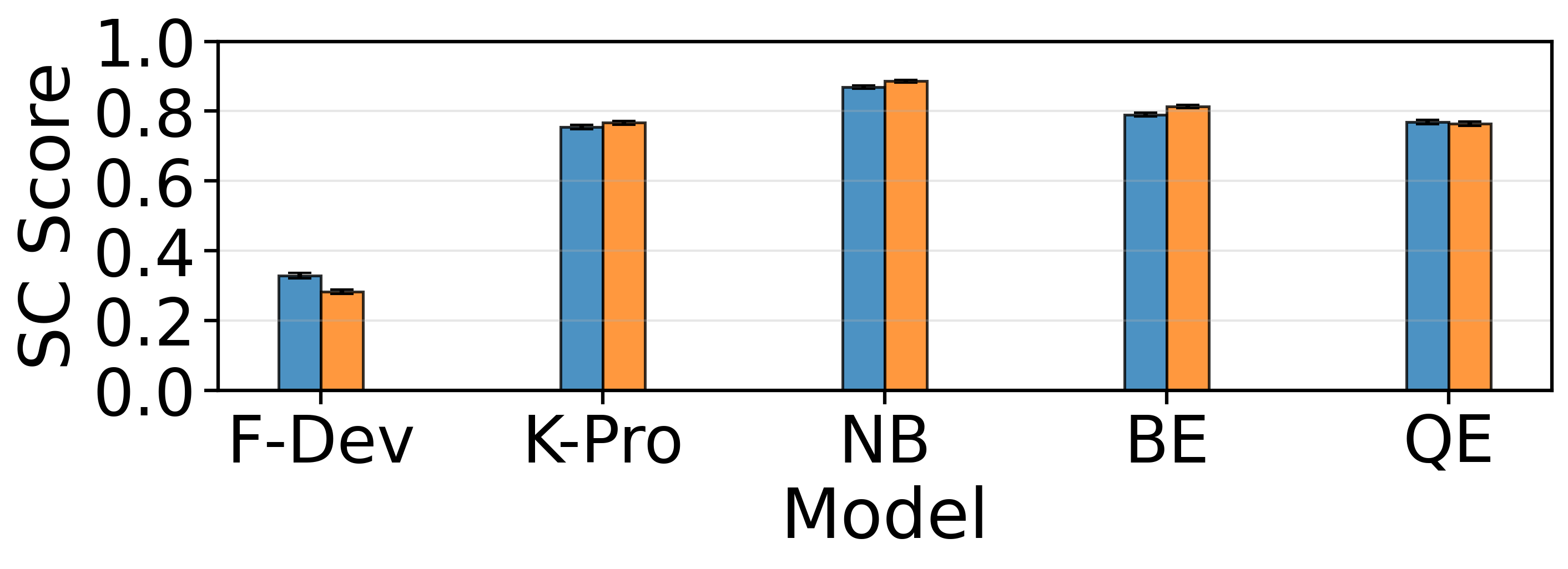}
    \caption{Hair}
  \end{subfigure}

  \vspace{0.10em}

  \begin{subfigure}[t]{0.49\textwidth}
    \centering
    \includegraphics[width=\linewidth]{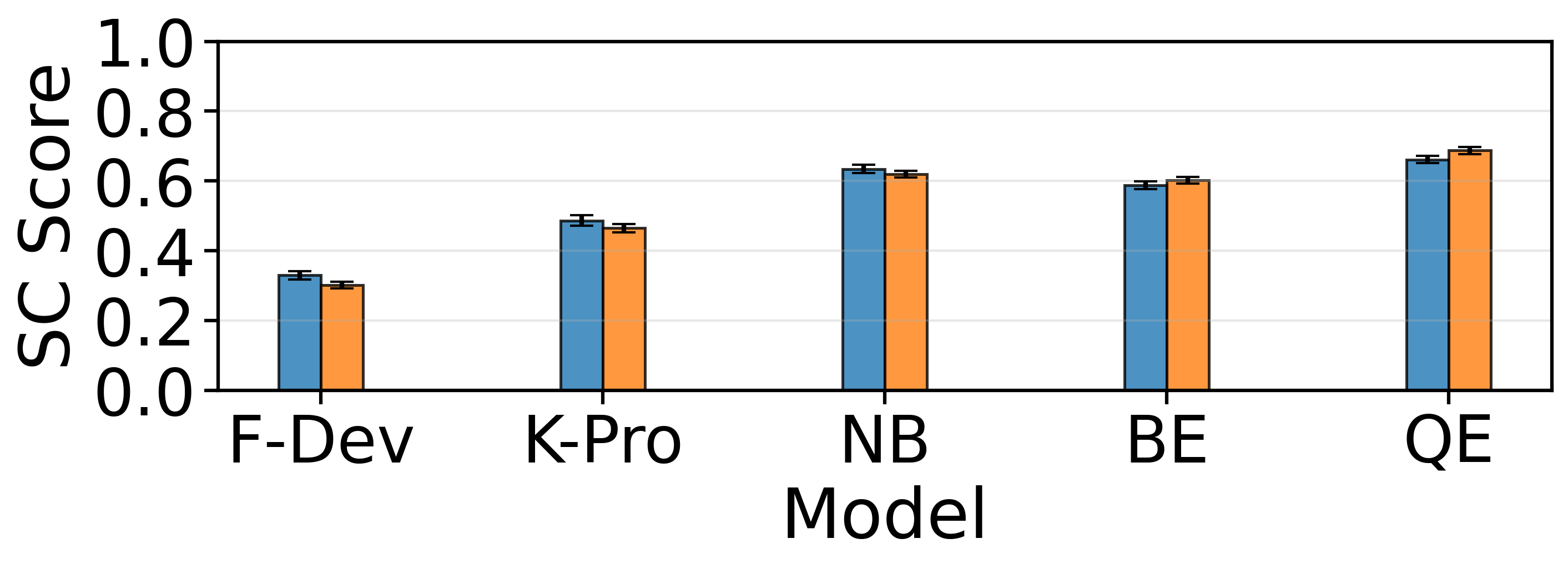}
    \caption{Pose}
  \end{subfigure}\hfill
  \begin{subfigure}[t]{0.49\textwidth}
    \centering
    \includegraphics[width=\linewidth]{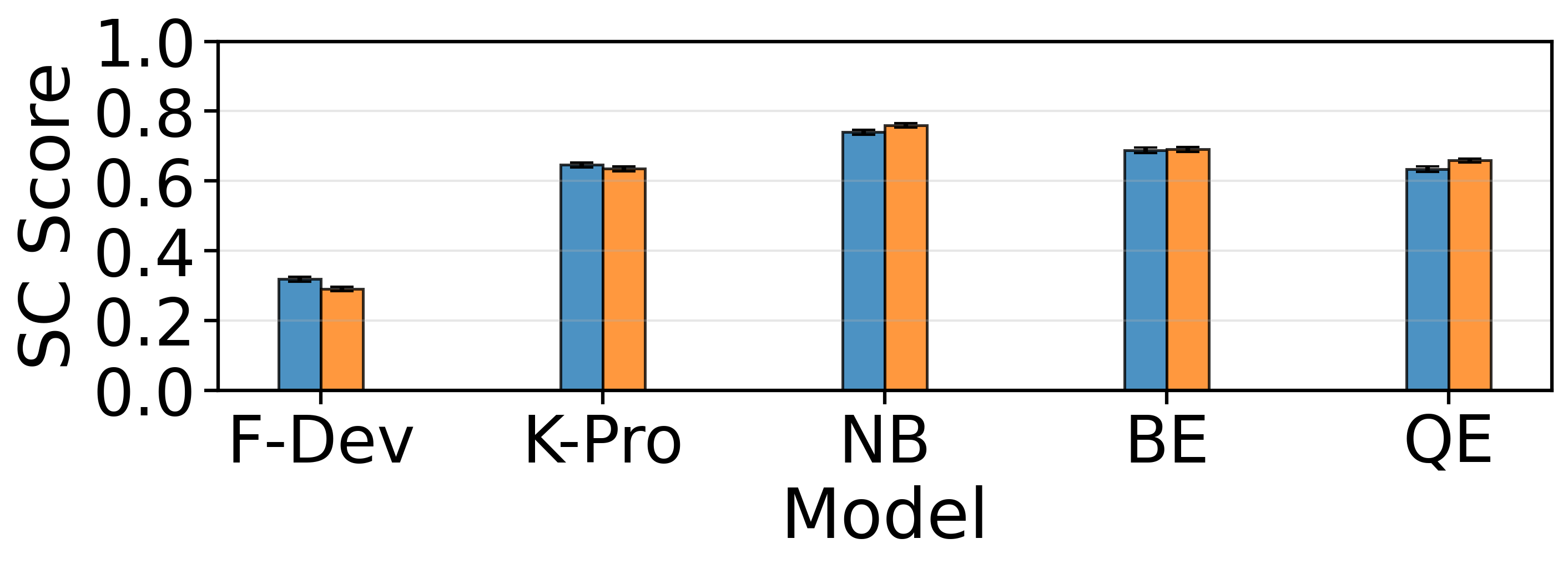}
    \caption{Multi-axis}
  \end{subfigure}

  \vspace{-0.2cm}
  \caption{$SC$ (higher is better) across gender subgroups in CelebA.}
  \label{fig:sem_con_celeba}
  \vspace{-0.1cm}
\end{figure*}

\begin{figure*}[h]
  \centering
  \quad
      \legendml\ Male-Light\;\;
      \legendmd\ Male-Dark\;\;
      \legendfl\ Female-Light\;\;
      \legendfd\ Female-Dark\par
  \vspace{0.4em}

  \captionsetup[subfigure]{margin={2.5em,0pt}}

  \begin{subfigure}[t]{0.49\textwidth}
    \centering
    \includegraphics[width=\linewidth]{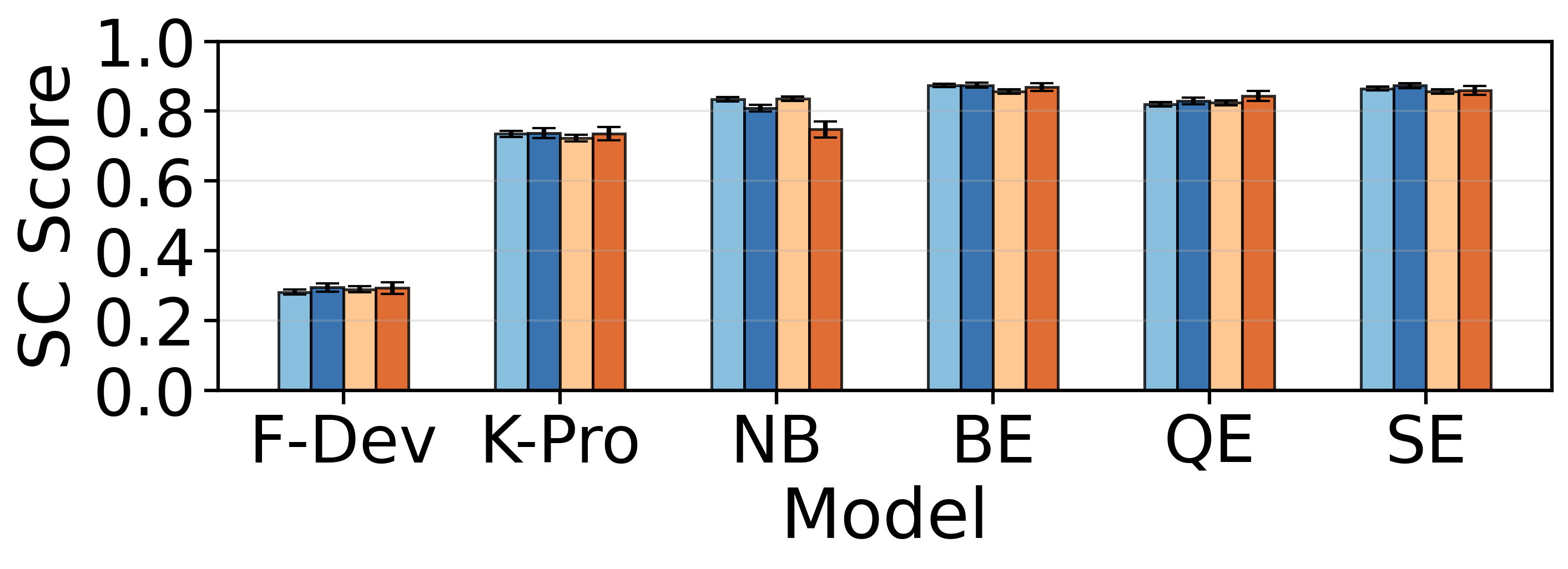}
    \caption{Accessories}
  \end{subfigure}\hfill
  \begin{subfigure}[t]{0.49\textwidth}
    \centering
    \includegraphics[width=\linewidth]{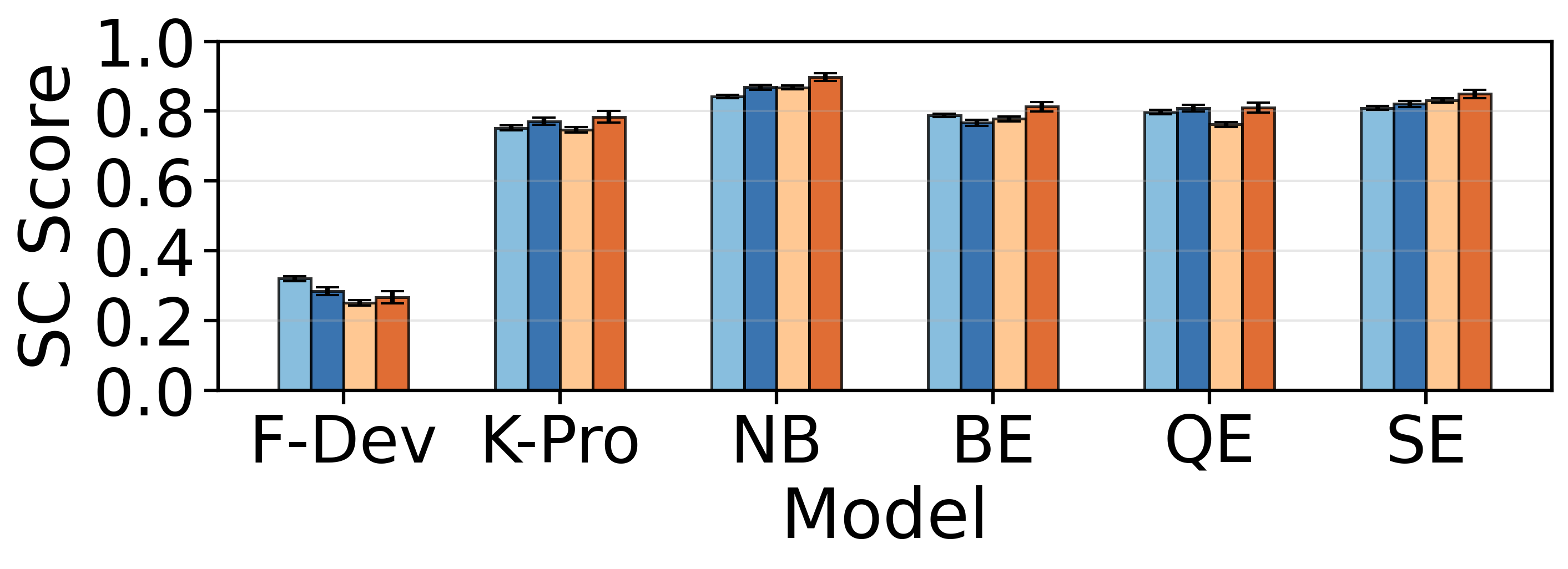}
    \caption{Hair}
  \end{subfigure}

  \vspace{0.10em}

  \begin{subfigure}[t]{0.49\textwidth}
    \centering
    \includegraphics[width=\linewidth]{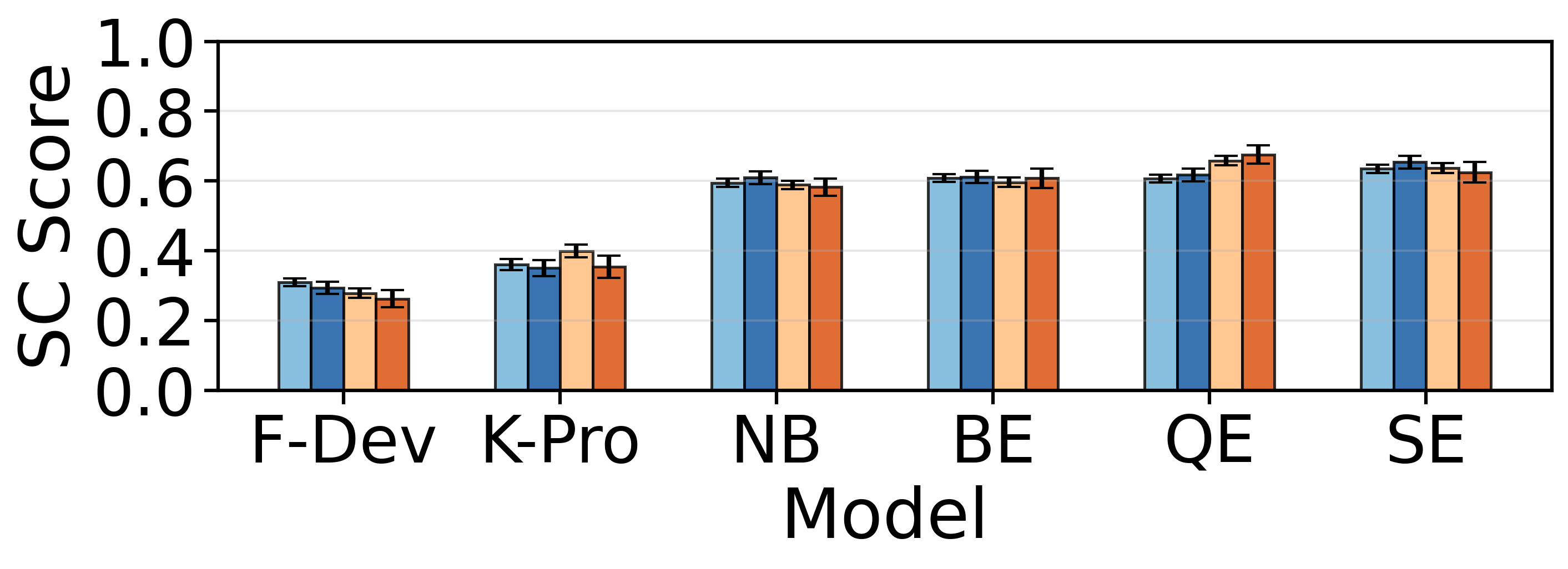}
    \caption{Pose}
  \end{subfigure}\hfill
  \begin{subfigure}[t]{0.49\textwidth}
    \centering
    \includegraphics[width=\linewidth]{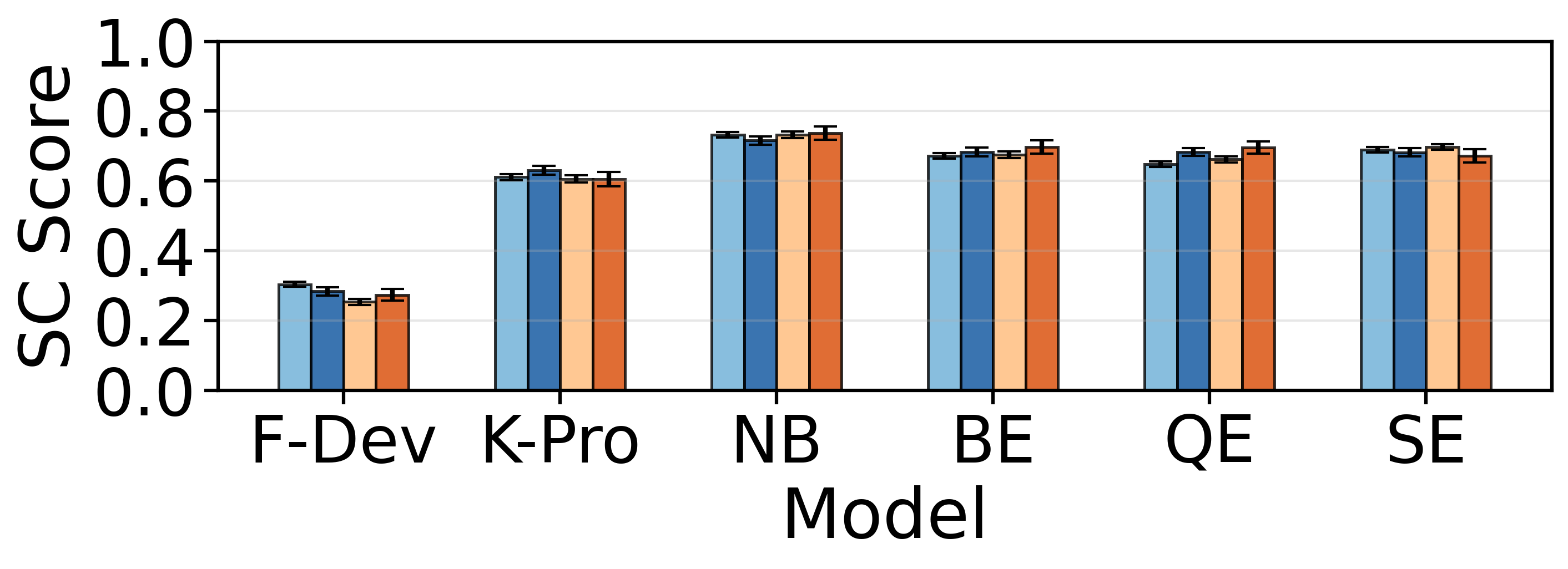}
    \caption{Multi-axis}
  \end{subfigure}

  \vspace{-0.2cm}
  \caption{$SC$ (higher is better) across intersectional demographic subgroups in CelebSET.}
  \label{fig:sem_con_celebset}
  \vspace{-0.1cm}
\end{figure*}

\begin{figure*}[h]
  \centering
  \quad \legendyoung\ Young\;\; \legendold\ Old \par
  \vspace{0.4em}

  \captionsetup[subfigure]{margin={2.5em,0pt}}

  \begin{subfigure}[t]{0.49\textwidth}
    \centering
    \includegraphics[width=\linewidth]{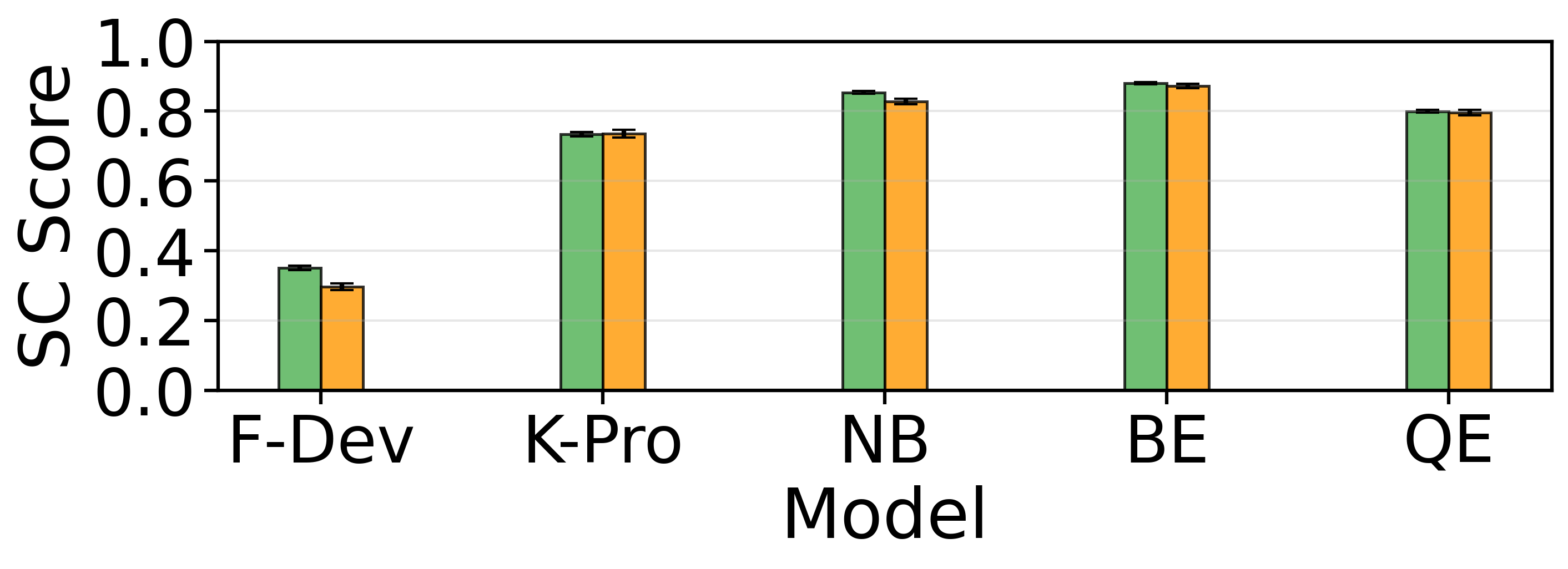}
    \caption{Accessories}
  \end{subfigure}\hfill
  \begin{subfigure}[t]{0.49\textwidth}
    \centering
    \includegraphics[width=\linewidth]{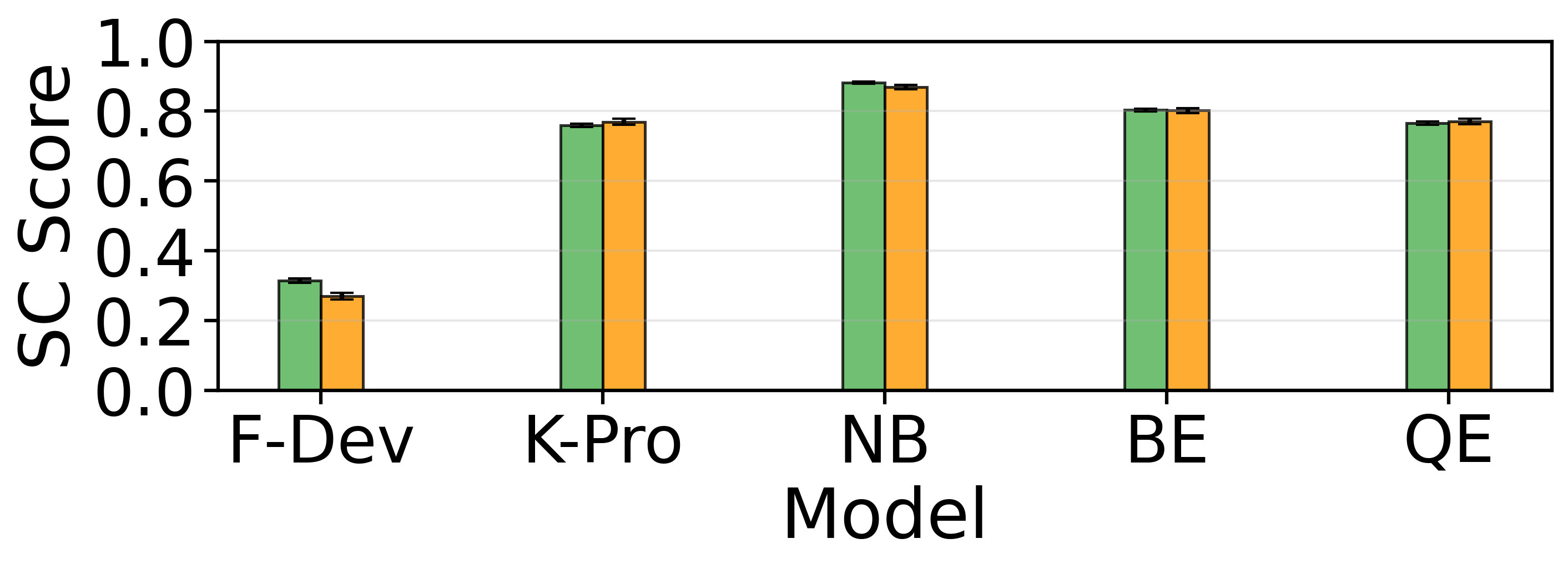}
    \caption{Hair}
  \end{subfigure}

  \vspace{0.10em}

  \begin{subfigure}[t]{0.49\textwidth}
    \centering
    \includegraphics[width=\linewidth]{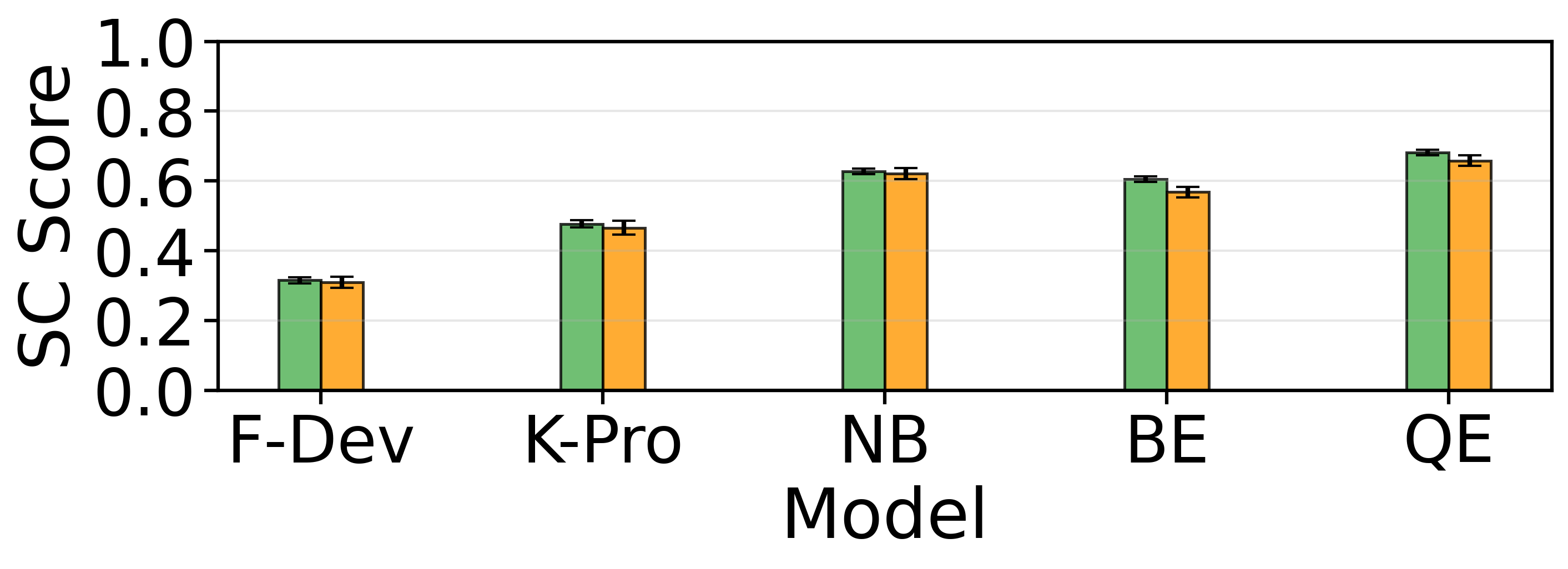}
    \caption{Pose}
  \end{subfigure}\hfill
  \begin{subfigure}[t]{0.49\textwidth}
    \centering
    \includegraphics[width=\linewidth]{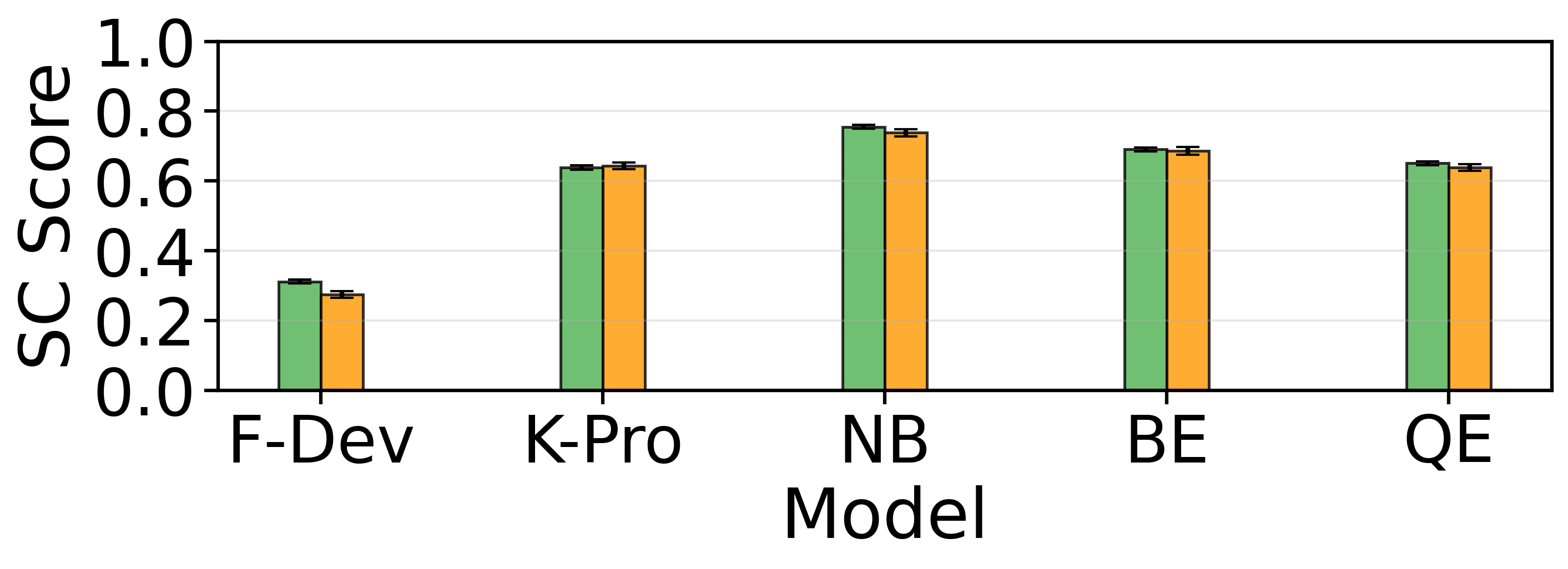}
    \caption{Multi-axis}
  \end{subfigure}

  \vspace{-0.2cm}
  \caption{$SC$ (higher is better) across age subgroups in CelebA.}
  \label{fig:sem_con_celeba_age}
  \vspace{-0.1cm}
\end{figure*}

\begin{figure*}[h]
  \centering
  \quad
      \legendyoung\ Young\;\;
      \legendold\ Old\;\;
      \par
  \vspace{0.4em}

  \captionsetup[subfigure]{margin={2.5em,0pt}}

  \begin{subfigure}[t]{0.49\textwidth}
    \centering
    \includegraphics[width=\linewidth]{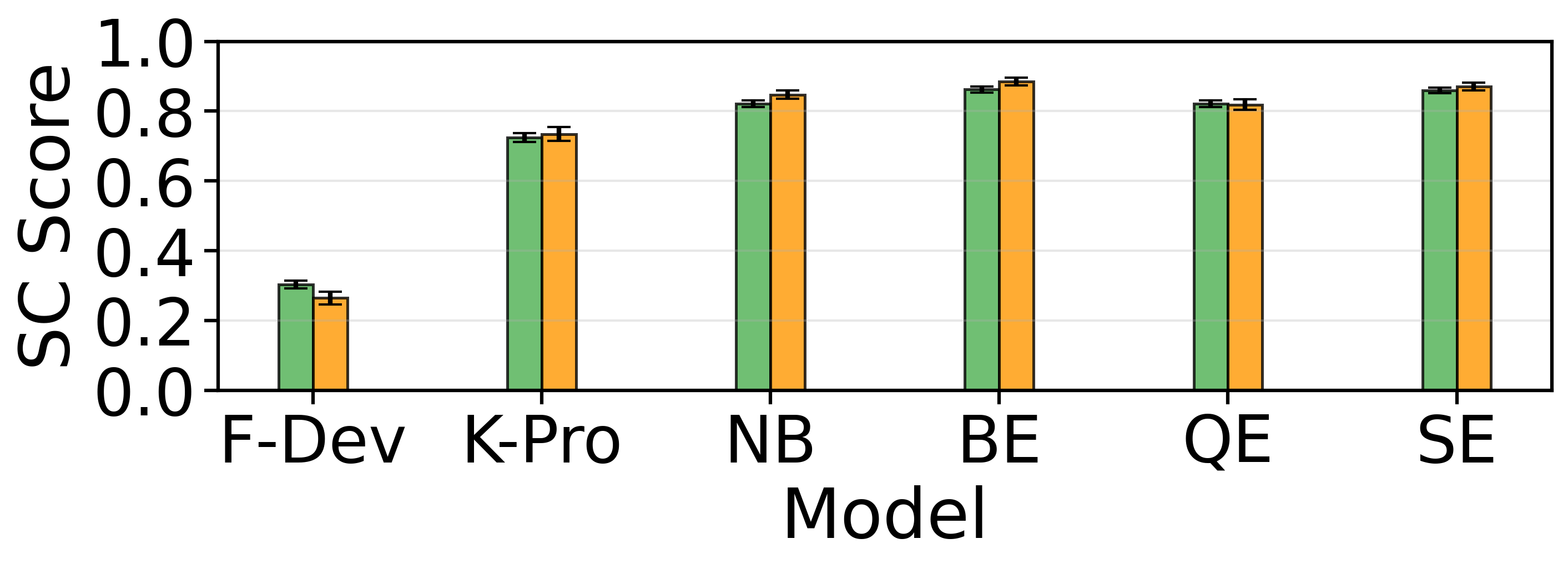}
    \caption{Accessories}
  \end{subfigure}\hfill
  \begin{subfigure}[t]{0.49\textwidth}
    \centering
    \includegraphics[width=\linewidth]{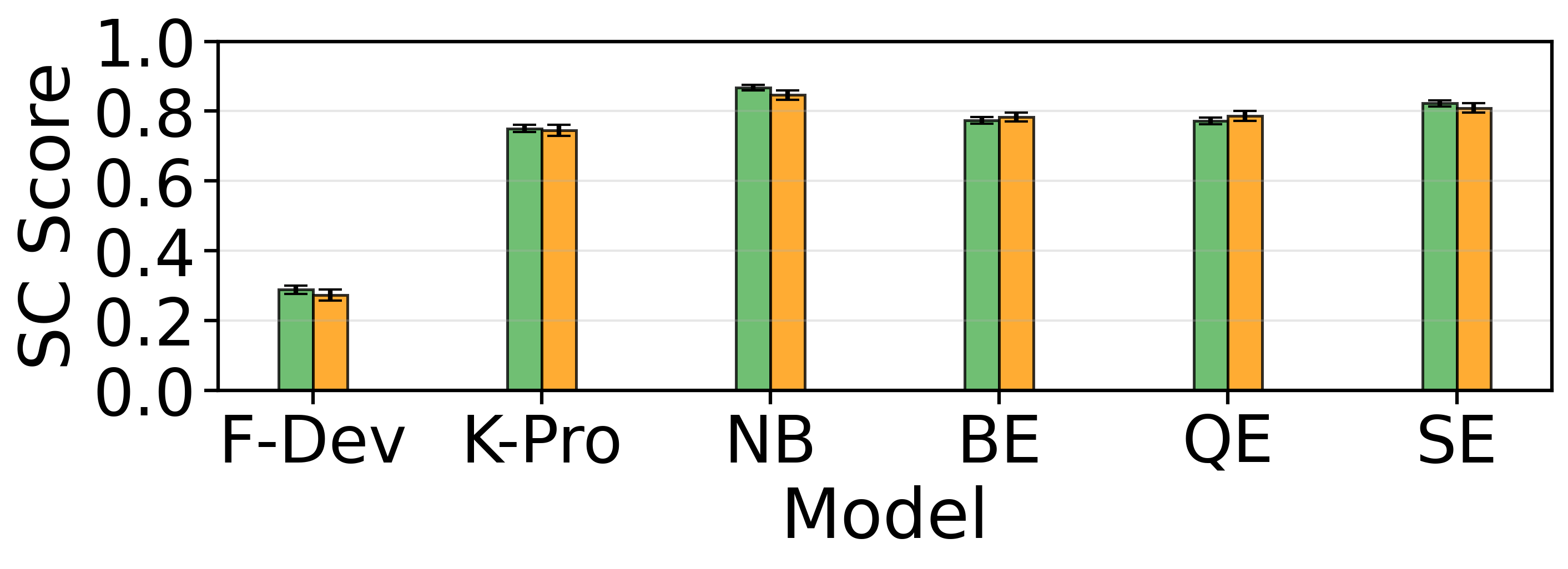}
    \caption{Hair}
  \end{subfigure}

  \vspace{0.10em}

  \begin{subfigure}[t]{0.49\textwidth}
    \centering
    \includegraphics[width=\linewidth]{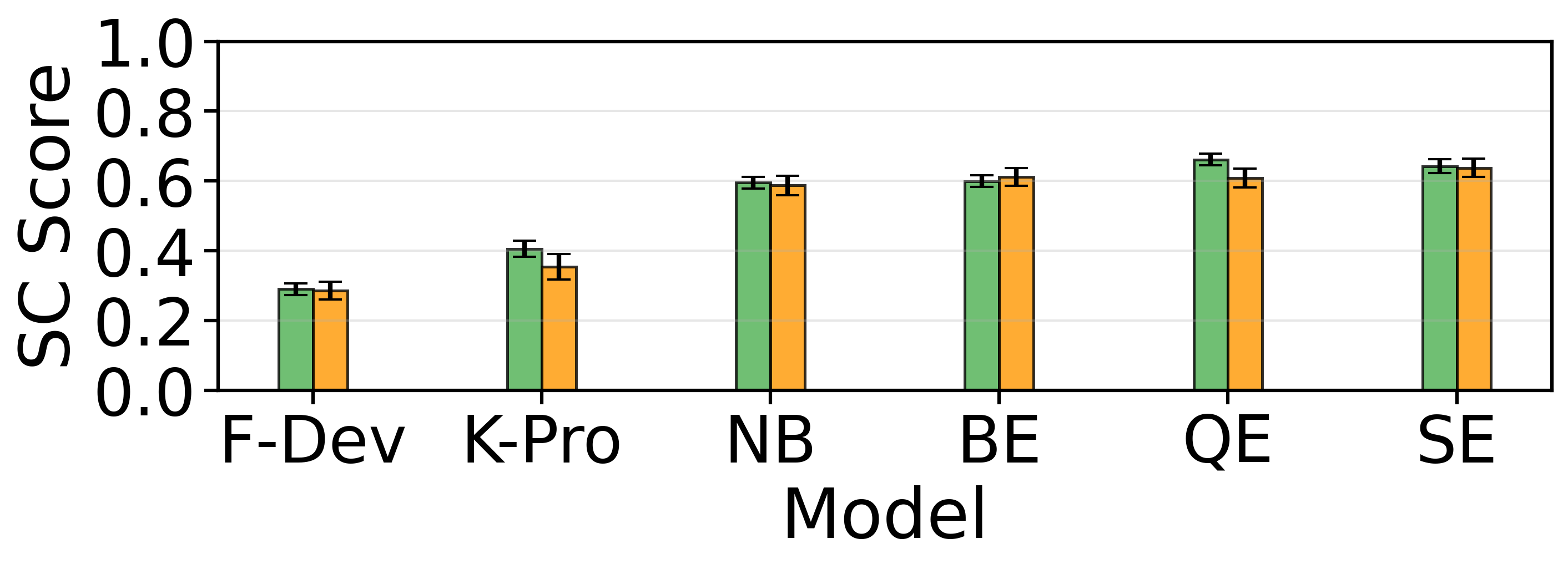}
    \caption{Pose}
  \end{subfigure}\hfill
  \begin{subfigure}[t]{0.49\textwidth}
    \centering
    \includegraphics[width=\linewidth]{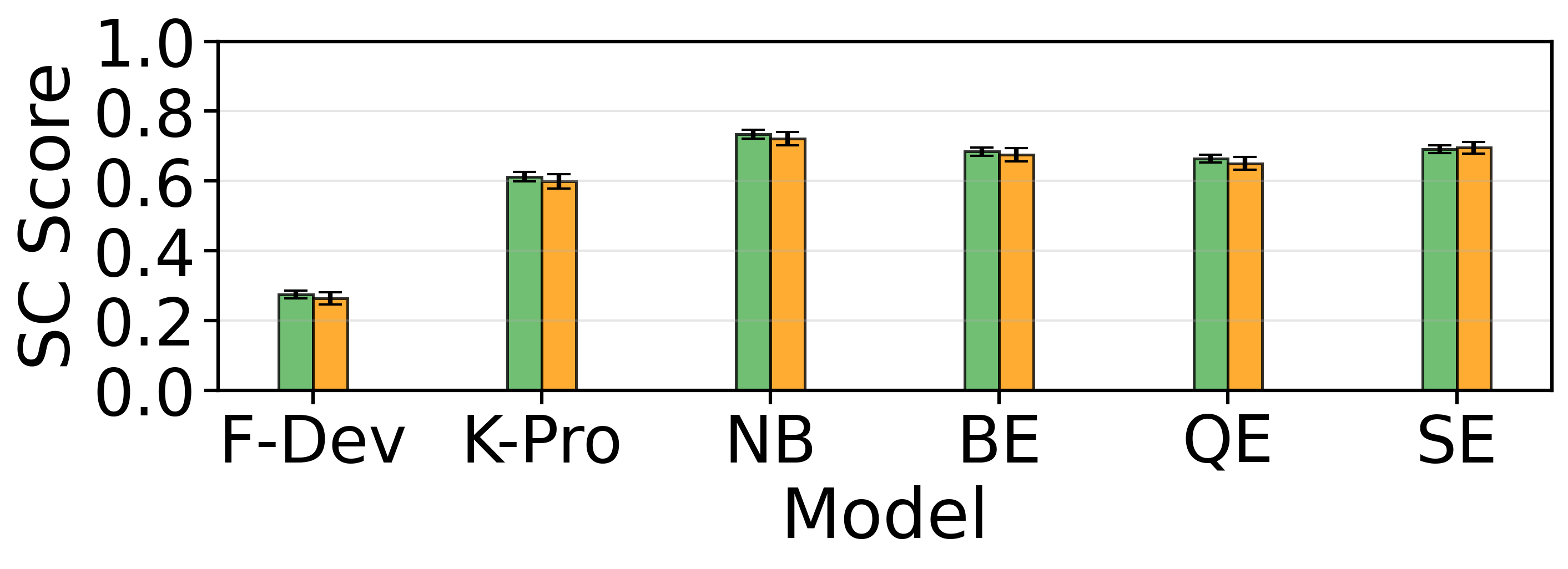}
    \caption{Multi-axis}
  \end{subfigure}

  \vspace{-0.2cm}
  \caption{$SC$ (higher is better) across age subgroups in CelebSET.}
  \label{fig:sem_con_celebset_age}
  \vspace{-0.1cm}
\end{figure*}

\begin{figure*}[h]
  \centering
  \quad \legendmale\ Male\;\; \legendfemale\ Female \par
  \vspace{0.4em}

  \captionsetup[subfigure]{margin={2.5em,0pt}}

  \begin{subfigure}[t]{0.49\textwidth}
    \centering
    \includegraphics[width=\linewidth]{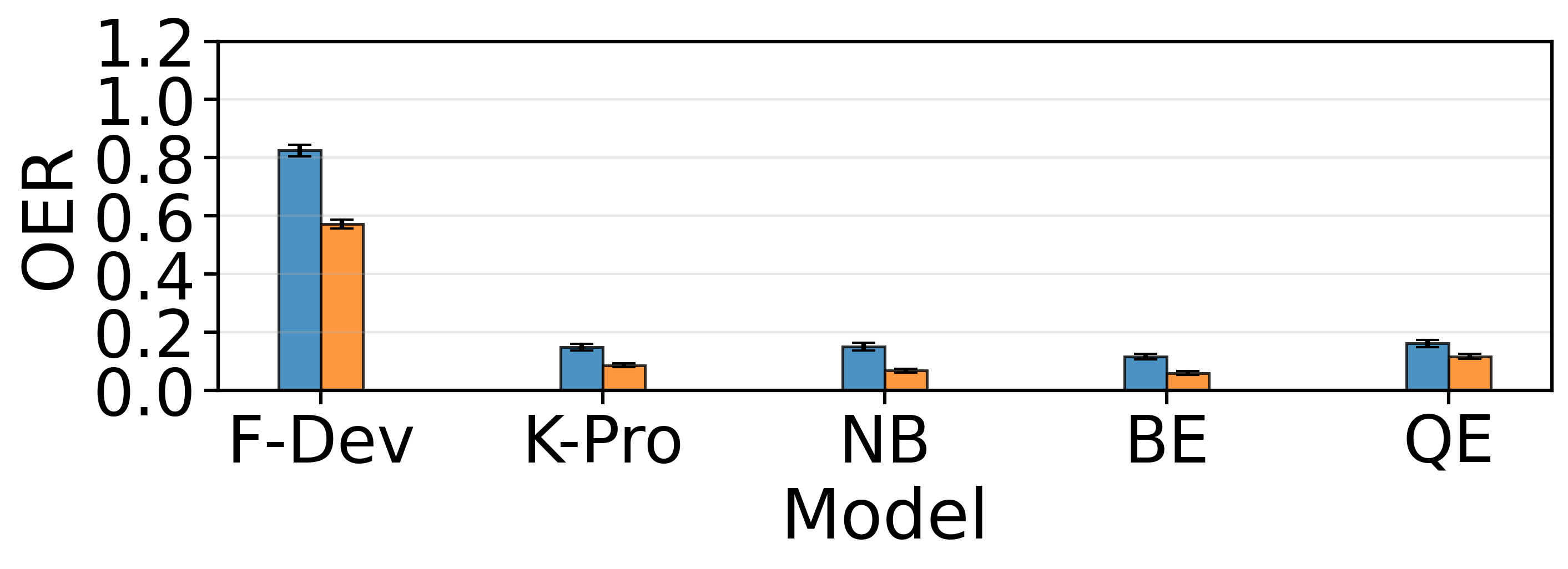}
    \caption{Accessories}
  \end{subfigure}\hfill
  \begin{subfigure}[t]{0.49\textwidth}
    \centering
    \includegraphics[width=\linewidth]{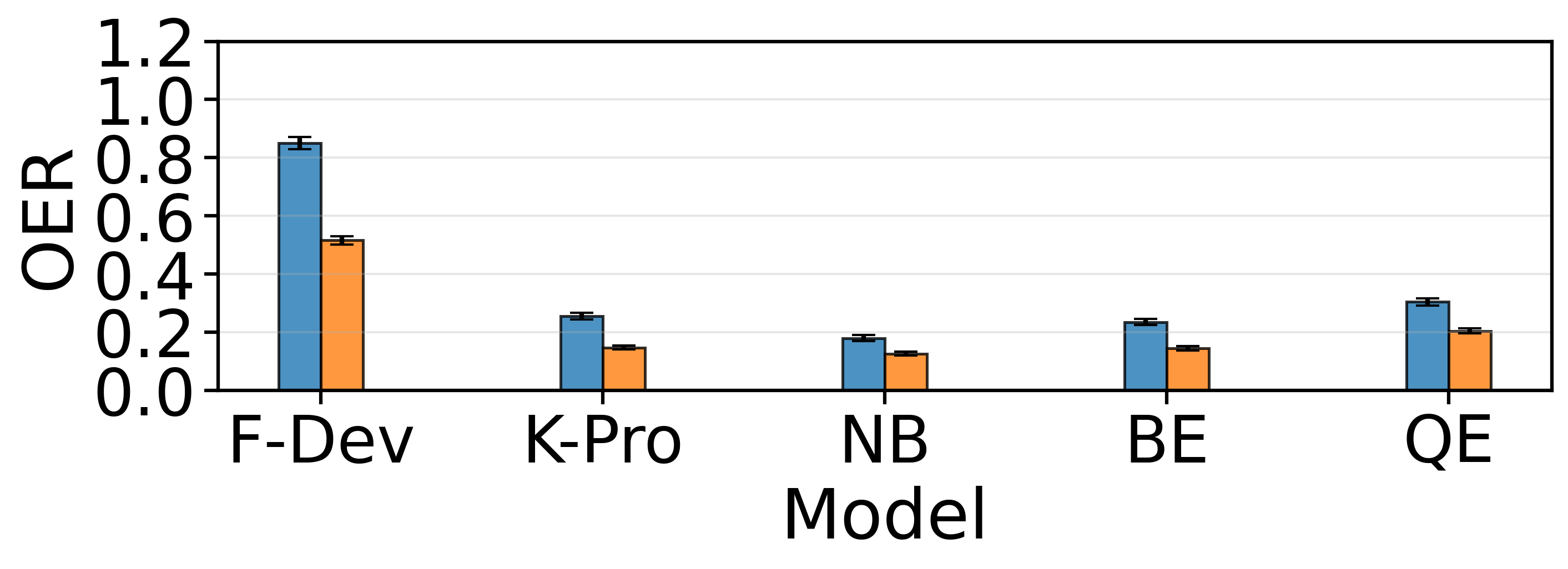}
    \caption{Hair}
  \end{subfigure}

  \vspace{0.10em}

  \begin{subfigure}[t]{0.49\textwidth}
    \centering
    \includegraphics[width=\linewidth]{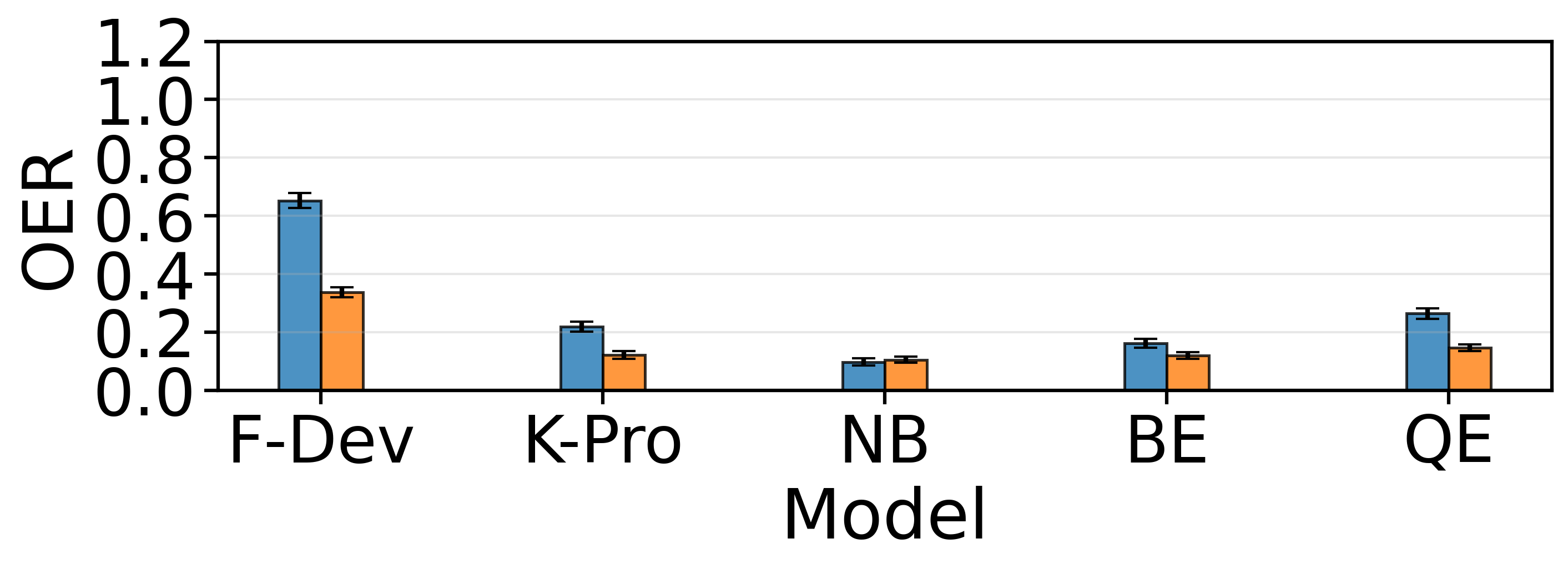}
    \caption{Pose}
  \end{subfigure}\hfill
  \begin{subfigure}[t]{0.49\textwidth}
    \centering
    \includegraphics[width=\linewidth]{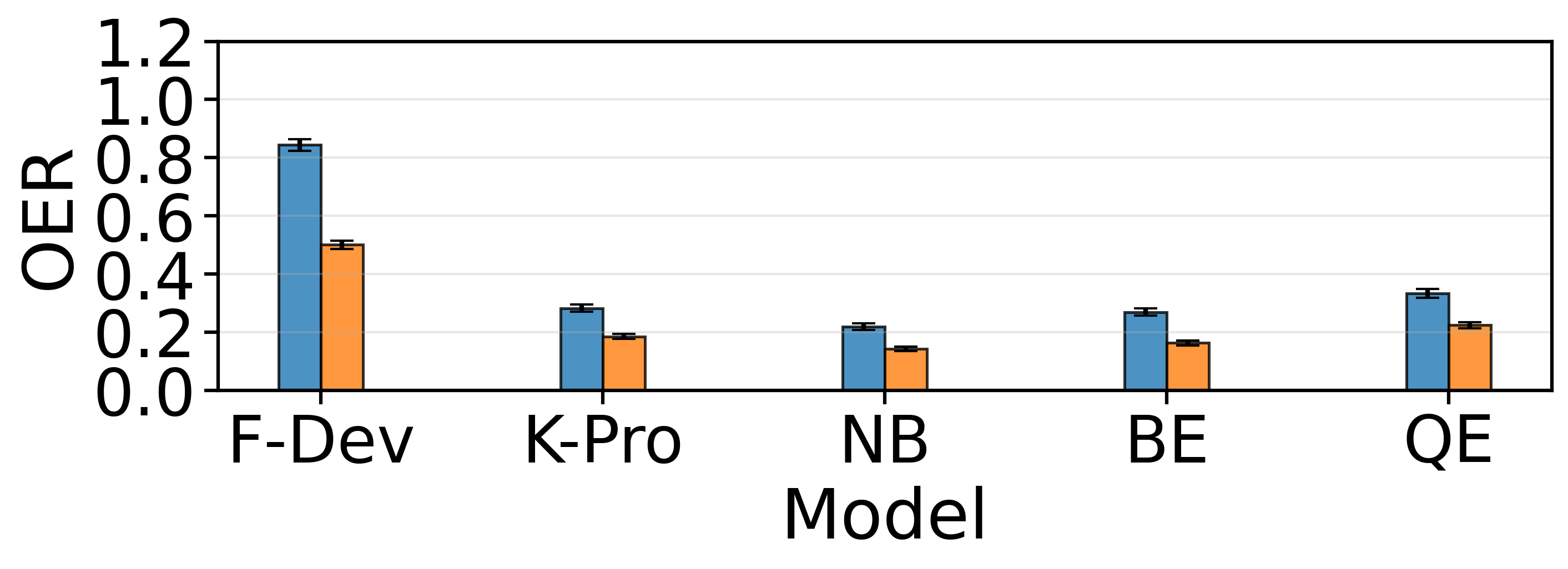}
    \caption{Multi-axis}
  \end{subfigure}

  \vspace{-0.2cm}
  \caption{OER (lower is better) across gender subgroups in CelebA.}
  \label{fig:oer_celeba}
  \vspace{-0.1cm}
\end{figure*}

\begin{figure*}[h]
  \centering
  \quad
      \legendml\ Male-Light\;\;
      \legendmd\ Male-Dark\;\;
      \legendfl\ Female-Light\;\;
      \legendfd\ Female-Dark\par
  \vspace{0.4em}

  \captionsetup[subfigure]{margin={2.5em,0pt}}

  \begin{subfigure}[t]{0.49\textwidth}
    \centering
    \includegraphics[width=\linewidth]{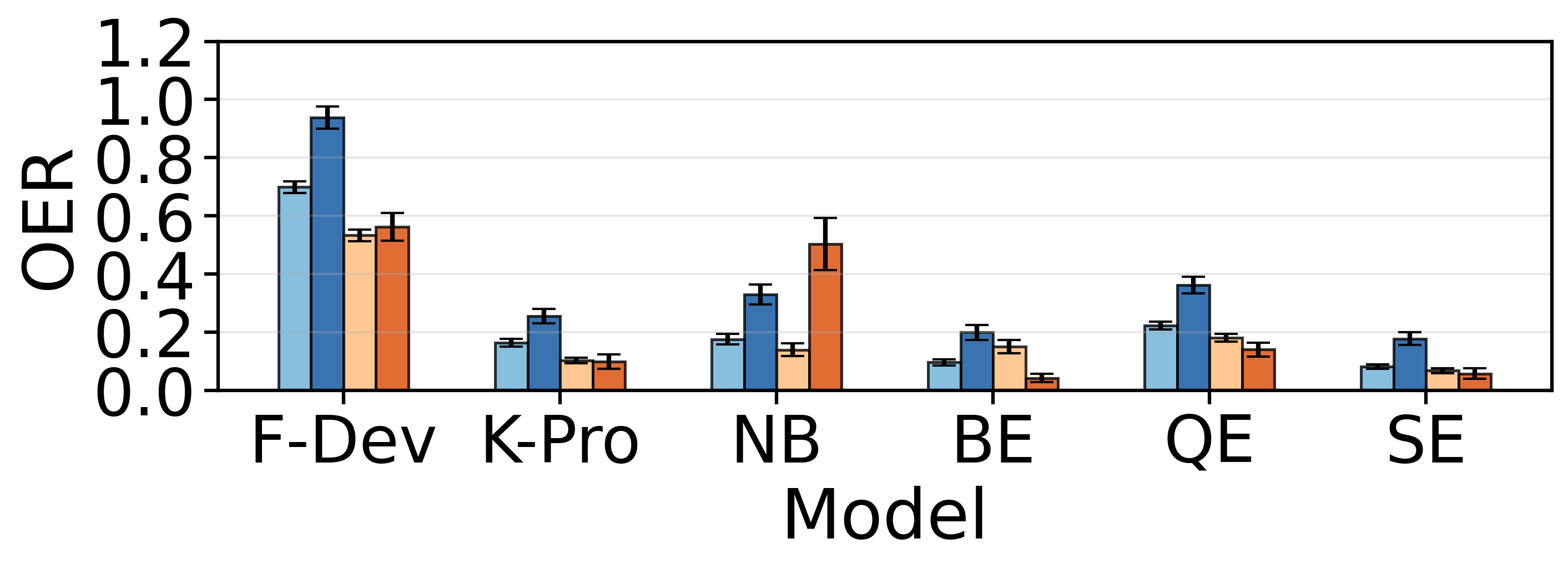}
    \caption{Accessories}
  \end{subfigure}\hfill
  \begin{subfigure}[t]{0.49\textwidth}
    \centering
    \includegraphics[width=\linewidth]{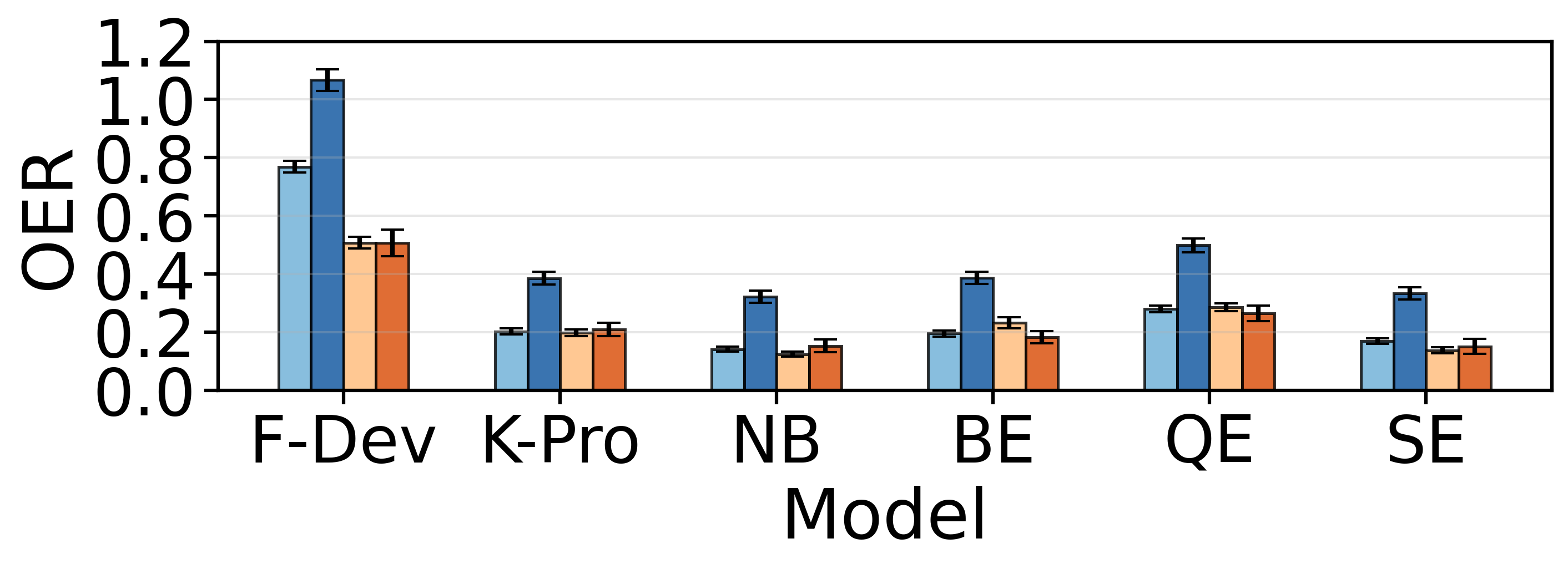}
    \caption{Hair}
  \end{subfigure}

  \vspace{0.10em}

  \begin{subfigure}[t]{0.49\textwidth}
    \centering
    \includegraphics[width=\linewidth]{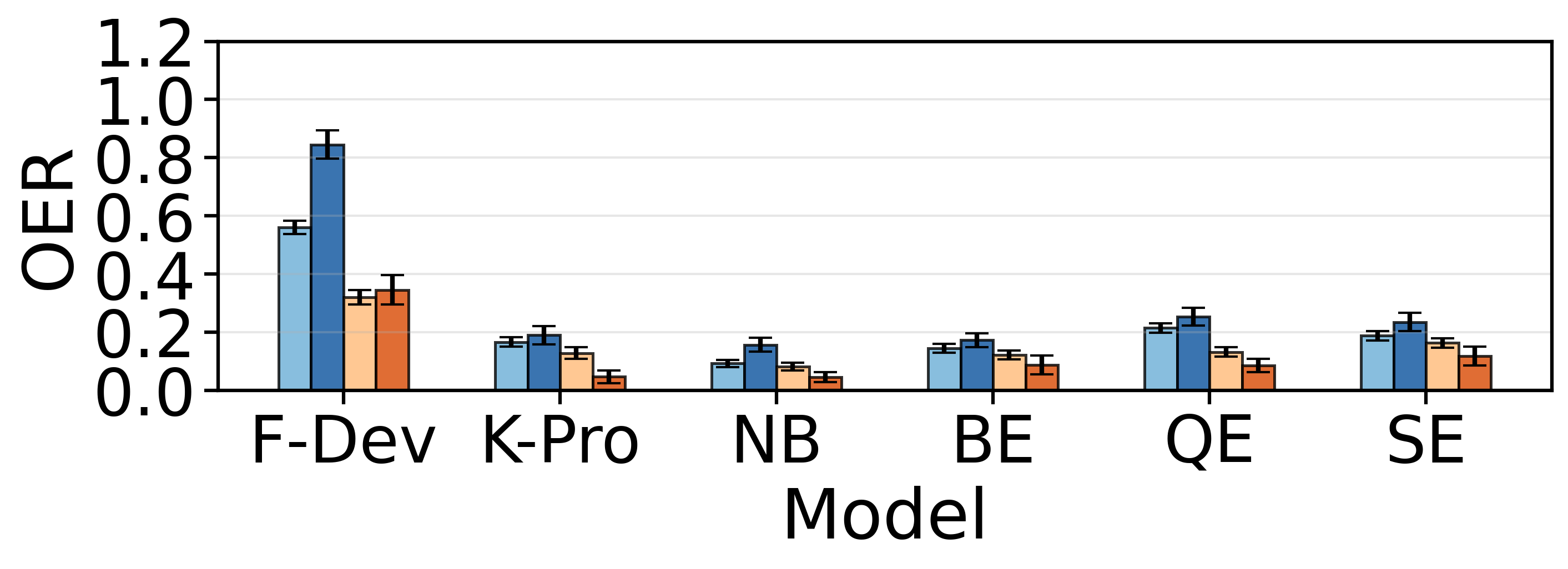}
    \caption{Pose}
  \end{subfigure}\hfill
  \begin{subfigure}[t]{0.49\textwidth}
    \centering
    \includegraphics[width=\linewidth]{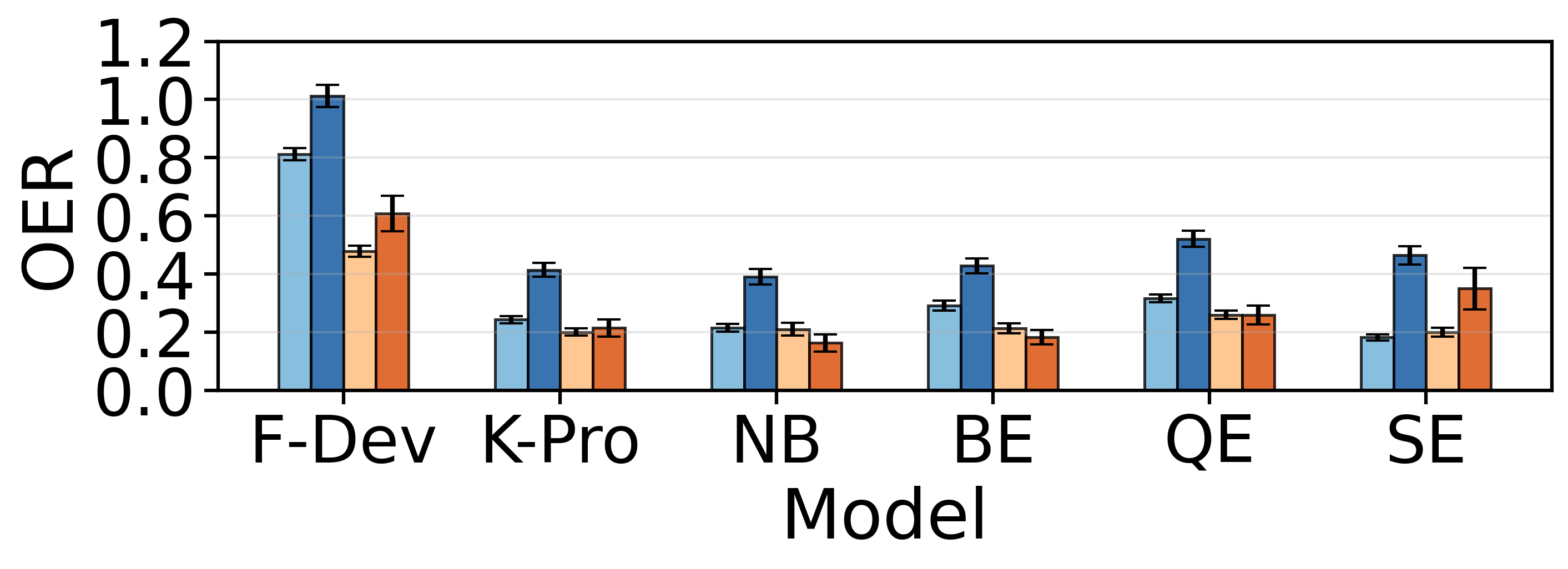}
    \caption{Multi-axis}
  \end{subfigure}

  \vspace{-0.2cm}
  \caption{OER (lower is better) across intersectional demographic subgroups in CelebSET.}
  \label{fig:oer_celebset}
  \vspace{-0.1cm}
\end{figure*}

\begin{figure*}[h]
  \centering
  \quad \legendyoung\ Young\;\; \legendold\ Old \par
  \vspace{0.4em}

  \captionsetup[subfigure]{margin={2.5em,0pt}}

  \begin{subfigure}[t]{0.49\textwidth}
    \centering
    \includegraphics[width=\linewidth]{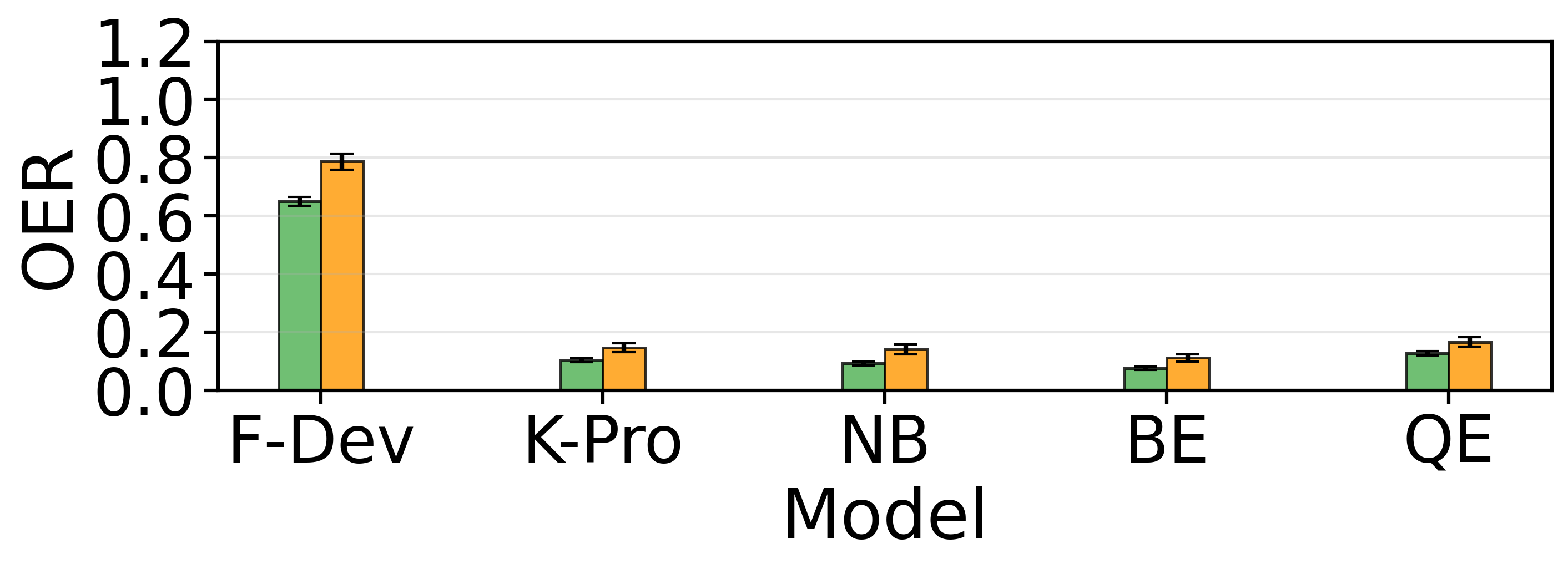}
    \caption{Accessories}
  \end{subfigure}\hfill
  \begin{subfigure}[t]{0.49\textwidth}
    \centering
    \includegraphics[width=\linewidth]{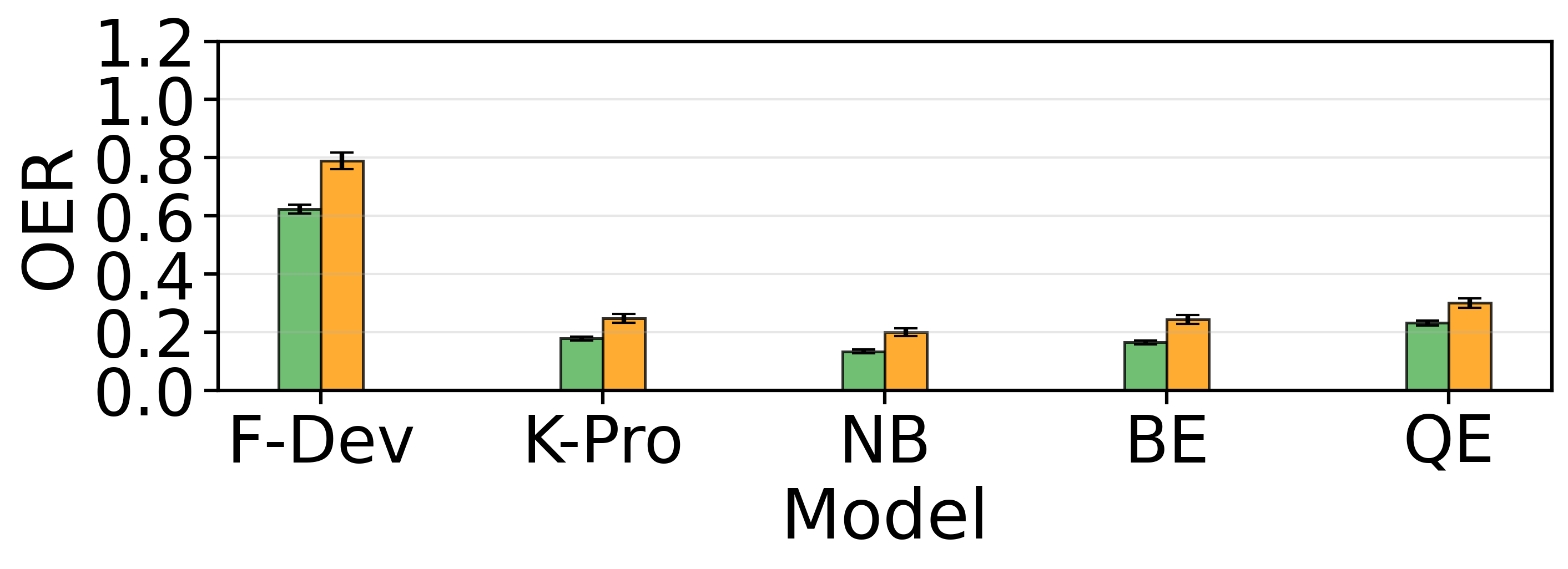}
    \caption{Hair}
  \end{subfigure}

  \vspace{0.10em}

  \begin{subfigure}[t]{0.49\textwidth}
    \centering
    \includegraphics[width=\linewidth]{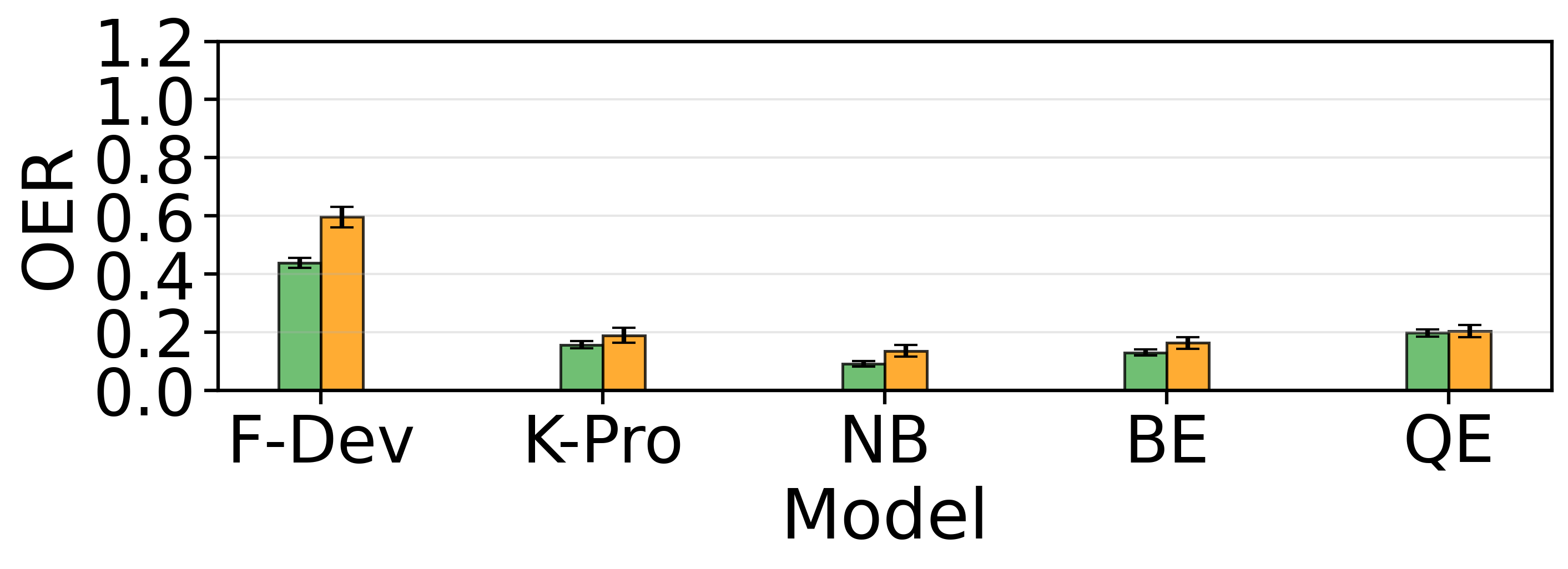}
    \caption{Pose}
  \end{subfigure}\hfill
  \begin{subfigure}[t]{0.49\textwidth}
    \centering
    \includegraphics[width=\linewidth]{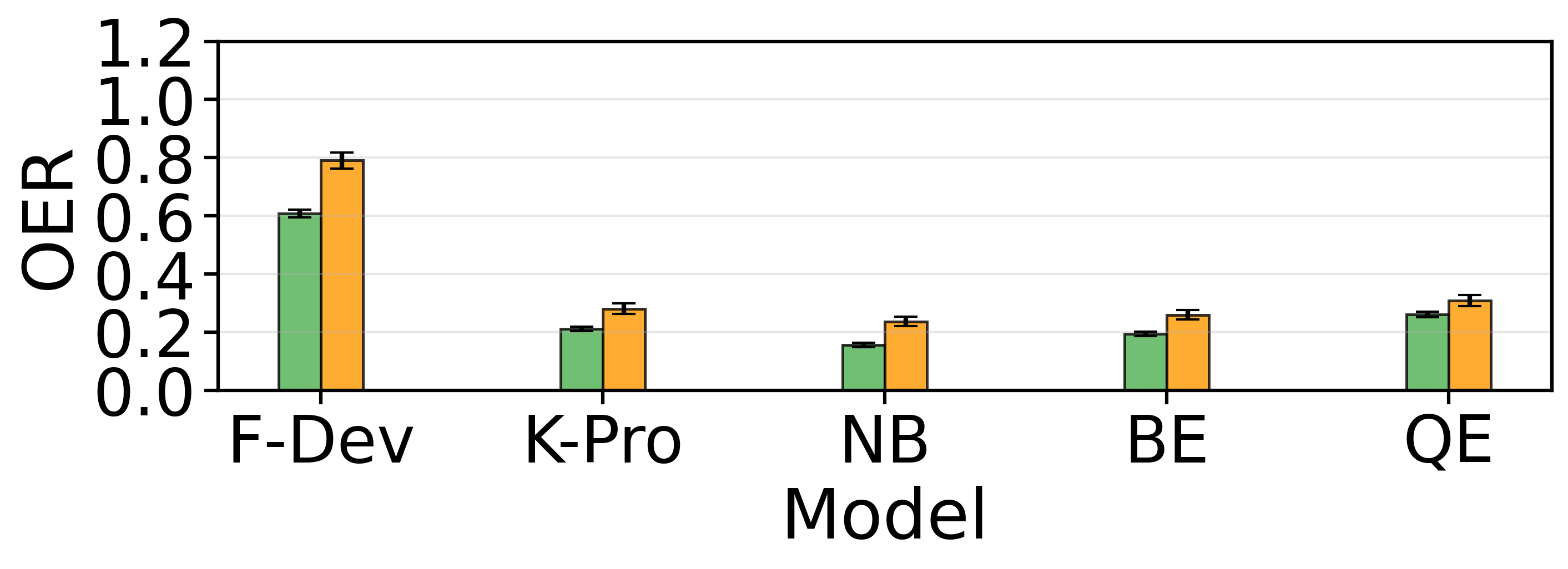}
    \caption{Multi-axis}
  \end{subfigure}

  \vspace{-0.2cm}
  \caption{OER (lower is better) across age subgroups in CelebA.}
  \label{fig:oer_celeba_age}
  \vspace{-0.1cm}
\end{figure*}

\begin{figure*}[h]
  \centering
  \quad
      \legendyoung\ Young\;\;
      \legendold\ Old\;\;
      \par
  \vspace{0.4em}

  \captionsetup[subfigure]{margin={2.5em,0pt}}

  \begin{subfigure}[t]{0.49\textwidth}
    \centering
    \includegraphics[width=\linewidth]{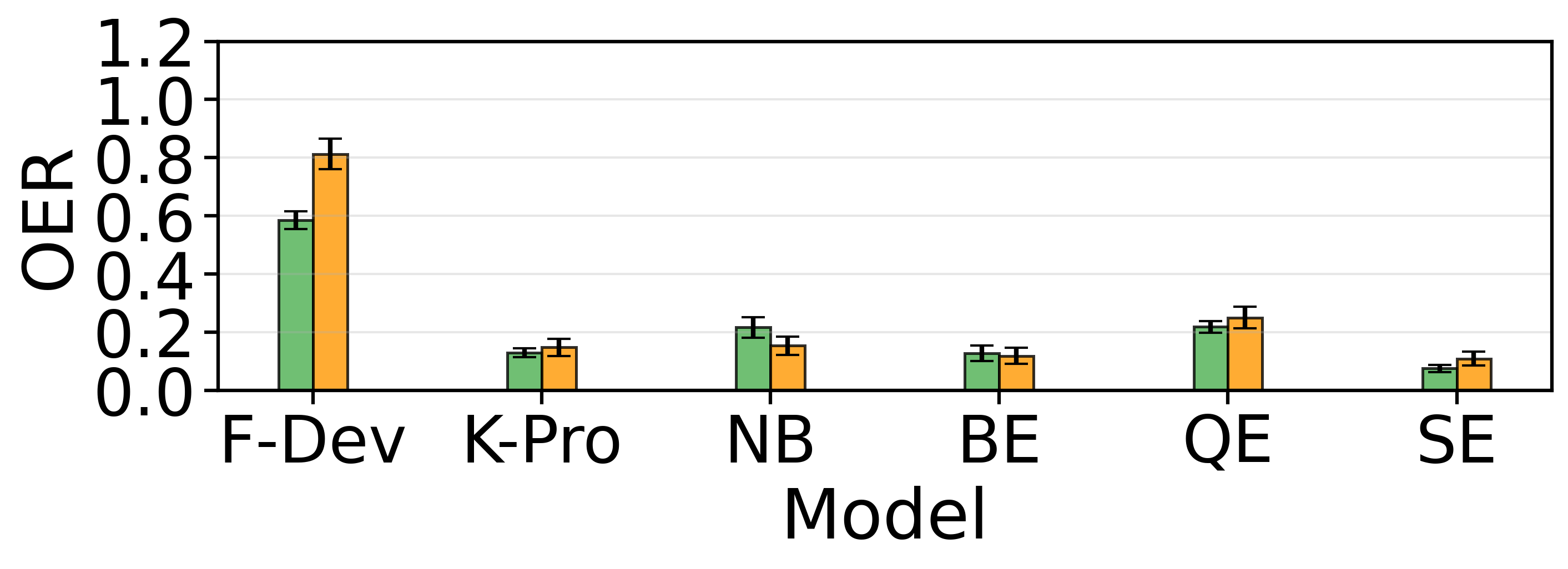}
    \caption{Accessories}
  \end{subfigure}\hfill
  \begin{subfigure}[t]{0.49\textwidth}
    \centering
    \includegraphics[width=\linewidth]{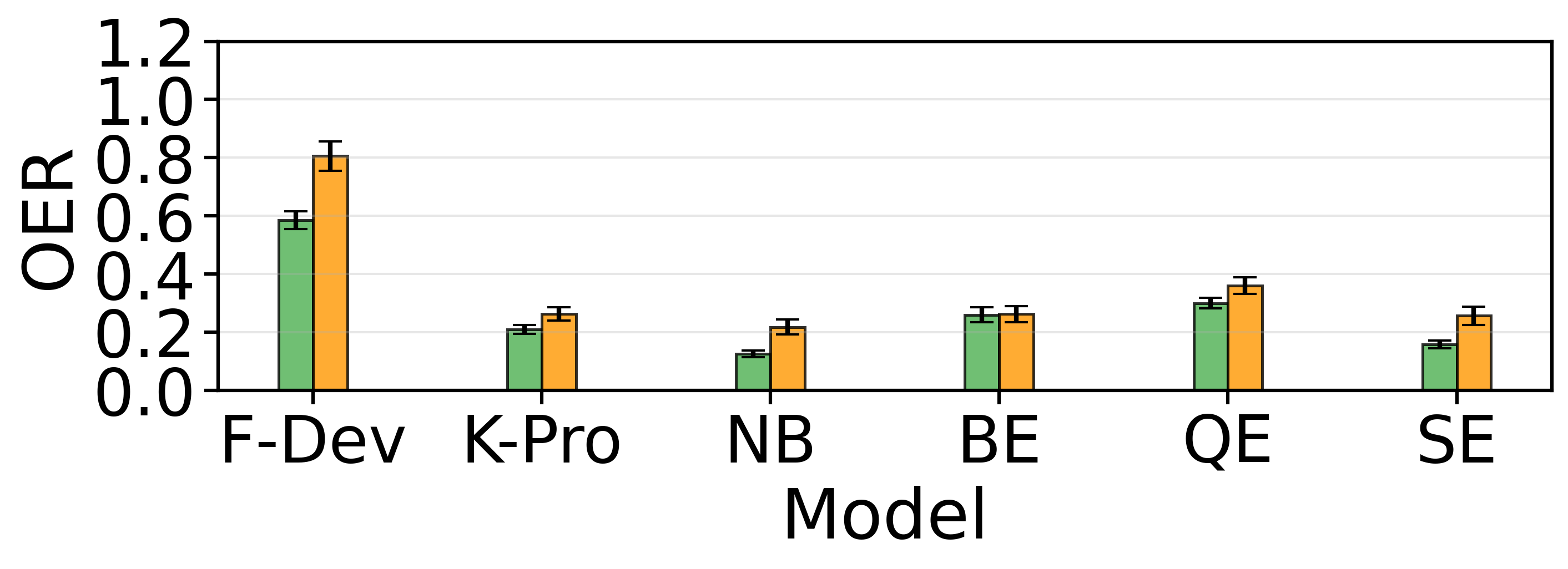}
    \caption{Hair}
  \end{subfigure}

  \vspace{0.10em}

  \begin{subfigure}[t]{0.49\textwidth}
    \centering
    \includegraphics[width=\linewidth]{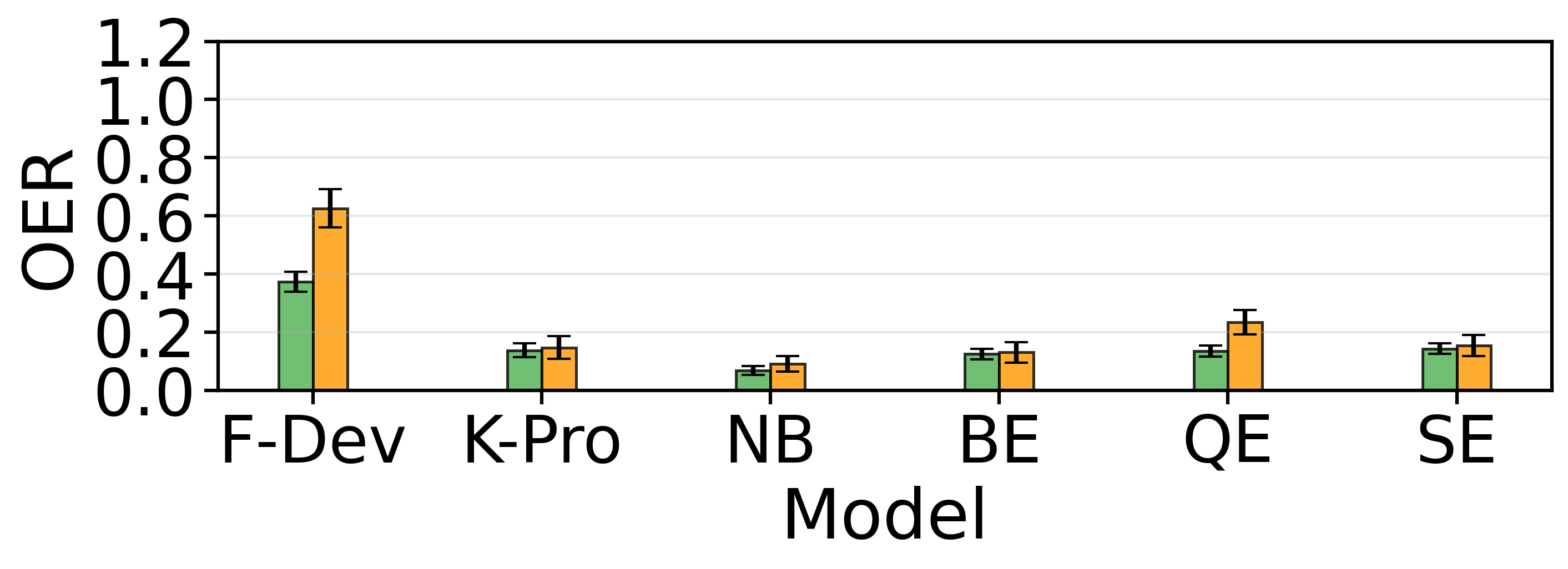}
    \caption{Pose}
  \end{subfigure}\hfill
  \begin{subfigure}[t]{0.49\textwidth}
    \centering
    \includegraphics[width=\linewidth]{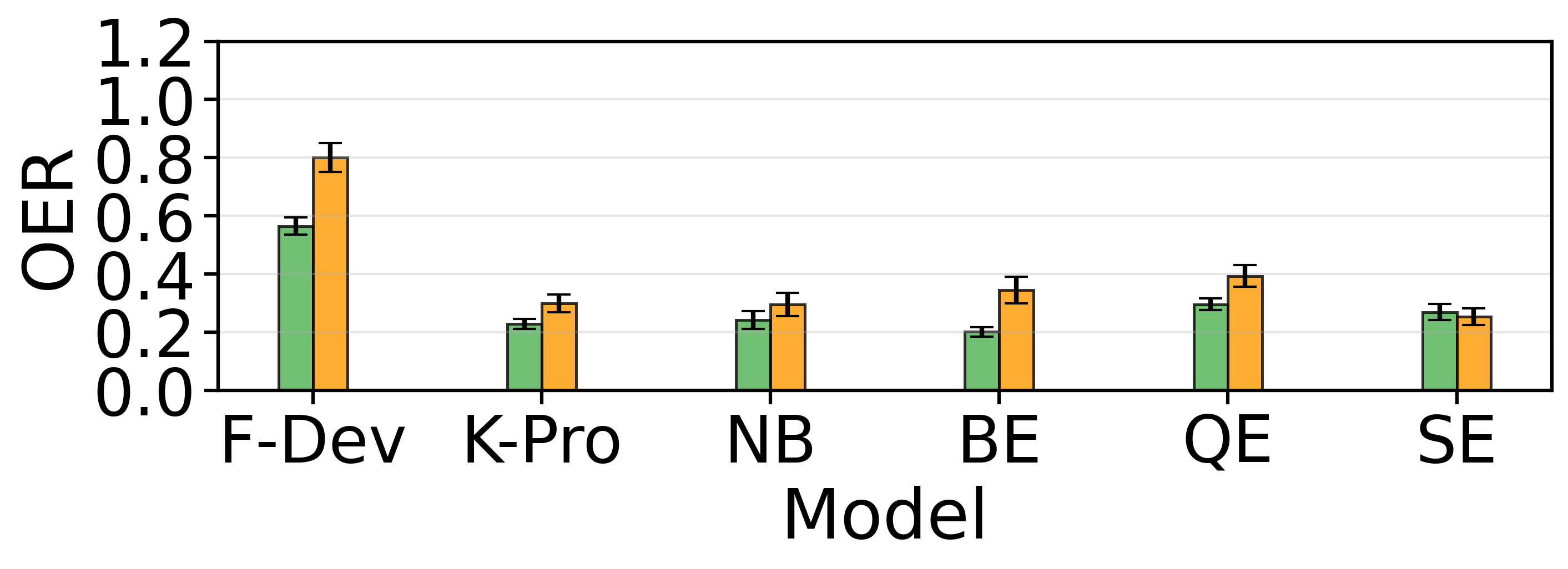}
    \caption{Multi-axis}
  \end{subfigure}

  \vspace{-0.2cm}
  \caption{OER (lower is better) across age subgroups in CelebSET.}
  \label{fig:oer_celebset_age}
  \vspace{-0.1cm}
\end{figure*}

\begin{table*}[t]
\centering
\caption{Attributes that frequently caused demographic biases in overediting. Model abbreviations: Nano-Banana (NB), SeedEdit (SE), Bagel-Edit (BE), Qwen-Edit (QE), Flux-Dev (F-Dev), and Kontext-Pro (K-Pro). Demographic abbreviations: Male (M), Female (F), Old (O), Young (Y), Light skin tone (L), and Dark skin tone (D).}
\label{tab:overedit_attributes_bias}
\vspace{-0.4cm}
\setlength{\tabcolsep}{4pt}
\renewcommand{\arraystretch}{1.05}
\footnotesize

\begin{minipage}[t]{0.48\textwidth}
\centering
\textbf{Gender Bias --- CelebA}\par\vspace{0.3em}
\begin{tabular}{@{} l p{1.3cm} p{0.56\linewidth} @{}}
\toprule
\textbf{Attribute} & \textbf{\# Models} & \textbf{Model Names} \\
\midrule
spiked cut        & 5 (4M, 1F) & F-Dev, K-Pro, NB, BE, QE \\
straight hair     & 5 (5M)     & F-Dev, K-Pro, NB, BE, QE \\
curly hair        & 4 (4M)     & F-Dev, NB, BE, QE \\
fumanchu mustache & 3 (3M)     & F-Dev, K-Pro, NB \\
blue hair         & 3 (3M)     & F-Dev, K-Pro, QE \\
mutton chops      & 3 (1M, 2F) & K-Pro, NB, BE \\
bob cut           & 3 (3M)     & K-Pro, NB, QE \\
green hair        & 3 (3M)     & K-Pro, BE, QE \\
black hair        & 3 (3M)     & K-Pro, NB, BE \\
brown hair        & 3 (3M)     & K-Pro, BE, QE \\
\bottomrule
\end{tabular}
\end{minipage}
\hfill
\begin{minipage}[t]{0.48\textwidth}
\centering
\textbf{Gender Bias --- CelebSET}\par\vspace{0.3em}
\begin{tabular}{@{} l p{1.3cm} p{0.56\linewidth} @{}}
\toprule
\textbf{Attribute} & \textbf{\# Models} & \textbf{Model Names} \\
\midrule
curly hair          & 6 (6M)     & F-Dev, K-Pro, NB, BE, QE, SE \\
fumanchu mustache   & 6 (6M)     & F-Dev, K-Pro, NB, BE, QE, SE \\
brown hair          & 6 (6M)     & F-Dev, K-Pro, NB, BE, QE, SE \\
straight hair       & 6 (6M)     & F-Dev, K-Pro, NB, BE, QE, SE \\
beanie hat          & 5 (5M)     & K-Pro, NB, BE, QE, SE \\
cowboy hat          & 5 (5M)     & K-Pro, NB, BE, QE, SE \\
toothbrush mustache & 4 (4M)     & F-Dev, K-Pro, BE, SE \\
spiked cut          & 4 (4M)     & K-Pro, BE, QE, SE \\
bob cut             & 3 (3M)     & F-Dev, K-Pro, QE \\
mutton chops        & 3 (1M, 2F) & K-Pro, NB, SE \\
\bottomrule
\end{tabular}
\end{minipage}

\vspace{0.7em}

\begin{minipage}[t]{0.48\textwidth}
\centering
\textbf{Age Bias --- CelebA}\par\vspace{0.3em}
\begin{tabular}{@{} l p{1.3cm} p{0.56\linewidth} @{}}
\toprule
\textbf{Attribute} & \textbf{\# Models} & \textbf{Model Names} \\
\midrule
curly hair    & 5 (5O) & F-Dev, K-Pro, NB, BE, QE \\
bob cut       & 5 (5O) & F-Dev, K-Pro, NB, BE, QE \\
spiked cut    & 4 (4O) & F-Dev, K-Pro, BE, QE \\
green hair    & 4 (4O) & F-Dev, K-Pro, BE, QE \\
straight hair & 4 (4O) & K-Pro, NB, BE, QE \\
brown hair    & 4 (4O) & K-Pro, NB, BE, QE \\
blue hair     & 3 (3O) & F-Dev, K-Pro, BE \\
full beard    & 3 (3O) & K-Pro, NB, QE \\
cowboy hat    & 3 (3O) & K-Pro, NB, BE \\
black hair    & 3 (3O) & K-Pro, NB, BE \\
\bottomrule
\end{tabular}
\end{minipage}
\hfill
\begin{minipage}[t]{0.48\textwidth}
\centering
\textbf{Age Bias --- CelebSET}\par\vspace{0.3em}
\begin{tabular}{@{} l p{1.3cm} p{0.56\linewidth} @{}}
\toprule
\textbf{Attribute} & \textbf{\# Models} & \textbf{Model Names} \\
\midrule
curly hair        & 6 (6O)     & F-Dev, K-Pro, NB, BE, QE, SE \\
spiked cut        & 6 (6O)     & F-Dev, K-Pro, NB, BE, QE, SE \\
bob cut           & 5 (5O)     & F-Dev, K-Pro, BE, QE, SE \\
straight hair     & 5 (5O)     & K-Pro, NB, BE, QE, SE \\
brown hair        & 4 (4O)     & F-Dev, K-Pro, QE, SE \\
beret hat         & 3 (1O, 2Y) & F-Dev, NB, QE \\
beanie hat        & 3 (3O)     & F-Dev, K-Pro, NB \\
blue hair         & 3 (3O)     & K-Pro, NB, SE \\
pencil mustache   & 3 (2O, 1Y) & NB, BE, SE \\
fumanchu mustache & 2 (2O)     & F-Dev, QE \\
\bottomrule
\end{tabular}
\end{minipage}

\vspace{0.7em}

\makebox[\textwidth][c]{%
\begin{minipage}{0.48\textwidth}
\centering
\textbf{Skin Tone Bias --- CelebSET}\par\vspace{0.3em}
\begin{tabular}{@{} l p{1.3cm} p{0.56\linewidth} @{}}
\toprule
\textbf{Attribute} & \textbf{\# Models} & \textbf{Model Names} \\
\midrule
fumanchu mustache & 6 (6D) & F-Dev, K-Pro, NB, BE, QE, SE \\
curly hair        & 6 (6D) & F-Dev, K-Pro, NB, BE, QE, SE \\
brown hair        & 6 (6D) & F-Dev, K-Pro, NB, BE, QE, SE \\
straight hair     & 5 (5D) & K-Pro, NB, BE, QE, SE \\
beanie hat        & 5 (5D) & K-Pro, NB, BE, QE, SE \\
spiked cut        & 4 (4D) & F-Dev, K-Pro, NB, SE \\
green hair        & 4 (4D) & F-Dev, K-Pro, BE, QE \\
beret hat         & 3 (3D) & F-Dev, NB, BE \\
cowboy hat        & 3 (3D) & K-Pro, NB, BE \\
black hair        & 3 (3D) & BE, QE, SE \\
\bottomrule
\end{tabular}
\end{minipage}%
}

\end{table*}

\FloatBarrier
\onecolumn
\twocolumn

\section{Which editing attributes commonly led to demographic biases in overediting?}
\label{app:per_operation_bias}
Across our four evaluation sets (hair, accessories, pose, and multi-axis), we used $40$ unique editing attributes in this study (e.g., curly hair, handlebar mustache). We identified which of these $40$ attributes commonly contributed to demographic bias in overediting. For each \{dataset, bias axis\} combination (e.g., CelebA, gender), we computed WD using the OER metric ($WD_{g}$ for this example) for all $40$ attributes, across all editing models. 

For each \{dataset, bias axis\} combination, we identified attributes that frequently occurred in the top-$10$ WD scores (highest overediting bias) across all models. We reported our results in Table \ref{tab:overedit_attributes_bias}. In each table, we showed: ($1$) editing attribute (e.g., curly hair), ($2$) the number of models in which that attribute came in the top-$10$ WD scores. We also showed the bias direction in parentheses. For instance, ($4$M, $1$F) means that in four models, males showed more overedits and in one model, females showed more overedits, $(3)$ the names of all those models. We reported $2$ tables for CelebA (gender, age), and $3$ tables for CelebSET (gender, skin-tone, and age). 

Across all five tables, we see a common trend: in almost all models, the highest demographic biases in overediting are caused by hair-based editing operations. This includes almost all types of hair edit operations like hair style (e.g., spiked cut), facial hair (e.g., fumanchu mustache), and hair color (e.g., green hair). Across the five tables, results are consistent with Section \ref{subsec: demographic_results}: models showed more overedits on men, dark-skinned faces, and old faces.

\section{FSS: Demographic Bias Analysis on Base Images}
\label{app:demographic_bias_fss_base_images}

To check if FSS showed demographic bias in the base images, we computed the average FSS using the base images for each demographic group. To measure disparity (or bias) between groups, we used the WD metric. For each group, we also reported mean values (with $95\%$ confidence intervals). For CelebA, we showed gender comparisons in Table \ref{tab:baseline_fss_celeba_gender} and age comparisons in Table \ref{tab:baseline_fss_celeba_age}. For CelebSET, we showed gender-skin-tone comparisons in Table \ref{tab:baseline_fss_celebset_gender_skin} and age comparisons in Table \ref{tab:baseline_fss_celebset_age}. Across all tables, we observed no significant demographic bias (WD always $<3$). 

\section{FSS: Occlusion Sensitivity Analysis on Base Images}
\label{app:occlusion_sensitivity}

We used CelebA base images because only CelebA has annotations for facial accessories such as glasses and hats. We created two groups: ``with accessory'' (faces with hats, glasses, or both: $34$ images), and ``without accessory'' ($515$ images). Mean FSS for “with accessory”: $96.77 \pm 0.37$, “without accessory”: $96.82 \pm 0.10$. WD between these groups: $0.14$ (WD much less than $3$), with a p-val $= 0.9$ (WD is not statistically significant). FSS is not sensitive to occlusions: scores are similar even when base images include accessories such as glasses or hats.

\begin{table*}[!tb]
\centering
\caption{Baseline FSS (\%) across demographic groups on CelebA and CelebSET. Subscripts denote $95\%$ CI half-widths. $^{*}$ superscript indicates a non-significant p-value ($p > 0.05$). ML: Male-Light, FL: Female-Light, MD: Male-Dark, FD: Female-Dark.}
\vspace{-0.2cm}
\label{tab:baseline_fss}

\setlength{\tabcolsep}{5pt}
\renewcommand{\arraystretch}{1.0}
\scriptsize

\begin{subtable}[t]{0.23\textwidth}
\centering
\caption{CelebA - Gender}
\label{tab:baseline_fss_celeba_gender}
\vspace{0.1cm}
\begin{tabular}{lcc}
\toprule
Group & Mean$_{\text{CI}}$ & Count \\
\midrule
Male   & 96.79$_{0.15}$ & 244 \\
Female & 96.85$_{0.14}$ & 305 \\
\midrule
WD$_{g}$ $\downarrow$ & \multicolumn{2}{c}{0.13$^{*}$} \\
\bottomrule
\end{tabular}
\end{subtable}
\hfill
\begin{subtable}[t]{0.24\textwidth}
\centering
\caption{CelebA - Age}
\label{tab:baseline_fss_celeba_age}
\vspace{0.1cm}
\begin{tabular}{lcc}
\toprule
Group & Mean$_{\text{CI}}$ & Count \\
\midrule
Young & 96.79$_{0.11}$ & 413 \\
Old   & 96.91$_{0.18}$ & 136 \\
\midrule
WD$_{age}$ $\downarrow$ & \multicolumn{2}{c}{0.28} \\
\bottomrule
\end{tabular}
\end{subtable}
\hfill
\begin{subtable}[t]{0.24\textwidth}
\centering
\caption{CelebSet - Gender \& Skin}
\label{tab:baseline_fss_celebset_gender_skin}
\vspace{0.1cm}
\begin{tabular}{lcc}
\toprule
Group & Mean$_{\text{CI}}$ & Count \\
\midrule
ML & 89.26$_{0.51}$ & 222 \\
MD & 89.59$_{0.85}$ & 95 \\
FL & 90.06$_{0.69}$ & 161 \\
FD & 89.67$_{1.22}$ & 39 \\
\midrule
WD$_g$ $\downarrow$ & \multicolumn{2}{c}{0.63$^{*}$} \\
WD$_s$ $\downarrow$ & \multicolumn{2}{c}{0.30$^{*}$} \\
\bottomrule
\end{tabular}
\end{subtable}
\hfill
\begin{subtable}[t]{0.24\textwidth}
\centering
\caption{CelebSet - Age}
\label{tab:baseline_fss_celebset_age}
\vspace{0.1cm}
\begin{tabular}{lcc}
\toprule
Group & Mean$_{\text{CI}}$ & Count \\
\midrule
Young & 90.14$_{0.91}$ & 98 \\
Old   & 88.47$_{0.89}$ & 41 \\
\midrule
WD$_{age}$ $\downarrow$ & \multicolumn{2}{c}{1.74} \\
\bottomrule
\end{tabular}
\end{subtable}
\end{table*}

\begin{table*}[htbp]
\centering
\caption{Stereotype analysis using $SC$. For the male-centric instruction subset, we reported the mean $SC$ for demographic groups. In the Mean column, for CelebA: left is the worst group; right value is the best group. For CelebSET: left: worst group; right: second-worst group. $*$ marks non-significant values ($p > 0.05$). WD $> 3$ is highlighted. Values scaled by $100$.}
\vspace{-0.35cm}
\label{tab:sc_comparison_gender_rep}

\setlength{\tabcolsep}{4.2pt}
\scriptsize

\begin{subtable}[t]{0.48\textwidth}
\centering
\caption{CelebA}
\label{tab:sc_celeba_stereo}
\vspace{0.1cm}

\begin{tabular}{lcc}
\toprule
&
\multicolumn{2}{c}{Male-Centric Subset$\uparrow$} \\
\cmidrule(lr){2-3}
Model &
Mean & $WD_{g}$ \\
\midrule

Flux-Dev &
F:25.4 \quad M:31.2 & \textbf{5.8} \\

Kontext-Pro &
F:54.1 \quad F:55.9 & 1.8 \\

Nano Banana &
M:70.7 \quad F:74.2 & \textbf{4.4} \\

Bagel-Edit &
F:58.2 \quad M:61.4 & \textbf{3.2} \\

Qwen-Edit &
F:39.3 \quad M:39.8 & 1.1$^{*}$ \\

\bottomrule
\end{tabular}
\end{subtable}
\hfill
\begin{subtable}[t]{0.48\textwidth}
\centering
\caption{CelebSET}
\label{tab:sc_celebset_stereo}
\vspace{0.1cm}

\begin{tabular}{lccc}
\toprule
&
\multicolumn{3}{c}{Male-Centric Subset$\uparrow$} \\
\cmidrule(lr){2-4}
Model &
Mean & $WD_g$ & $WD_s$  \\
\midrule

Flux-Dev &
FL:24.8 \quad FD:25.5 & \textbf{9.4} & 0.9$^{*}$ \\

Kontext-Pro &
FL:54.0 \quad ML:55.9 & 2.1 & 2.1 \\

Nano Banana &
MD:66.1 \quad ML:72.0 & \textbf{5.3} & \textbf{4.8} \\

Bagel-Edit &
FL:56.8 \quad MD:61.0 & \textbf{3.4} & 1.8$^{*}$ \\

Qwen-Edit &
FL:43.1 \quad ML:44.5 & \textbf{3.6} & \textbf{5.8} \\

SeedEdit 3.0 &
ML:43.4 \quad FL:44.5 & 1.9 & 2.3  \\

\bottomrule
\end{tabular}
\end{subtable}
\end{table*}

\begin{figure*}[hbp]
  \centering
  \begin{subfigure}[t]{0.48\textwidth}
    \centering
    \includegraphics[width=\linewidth]{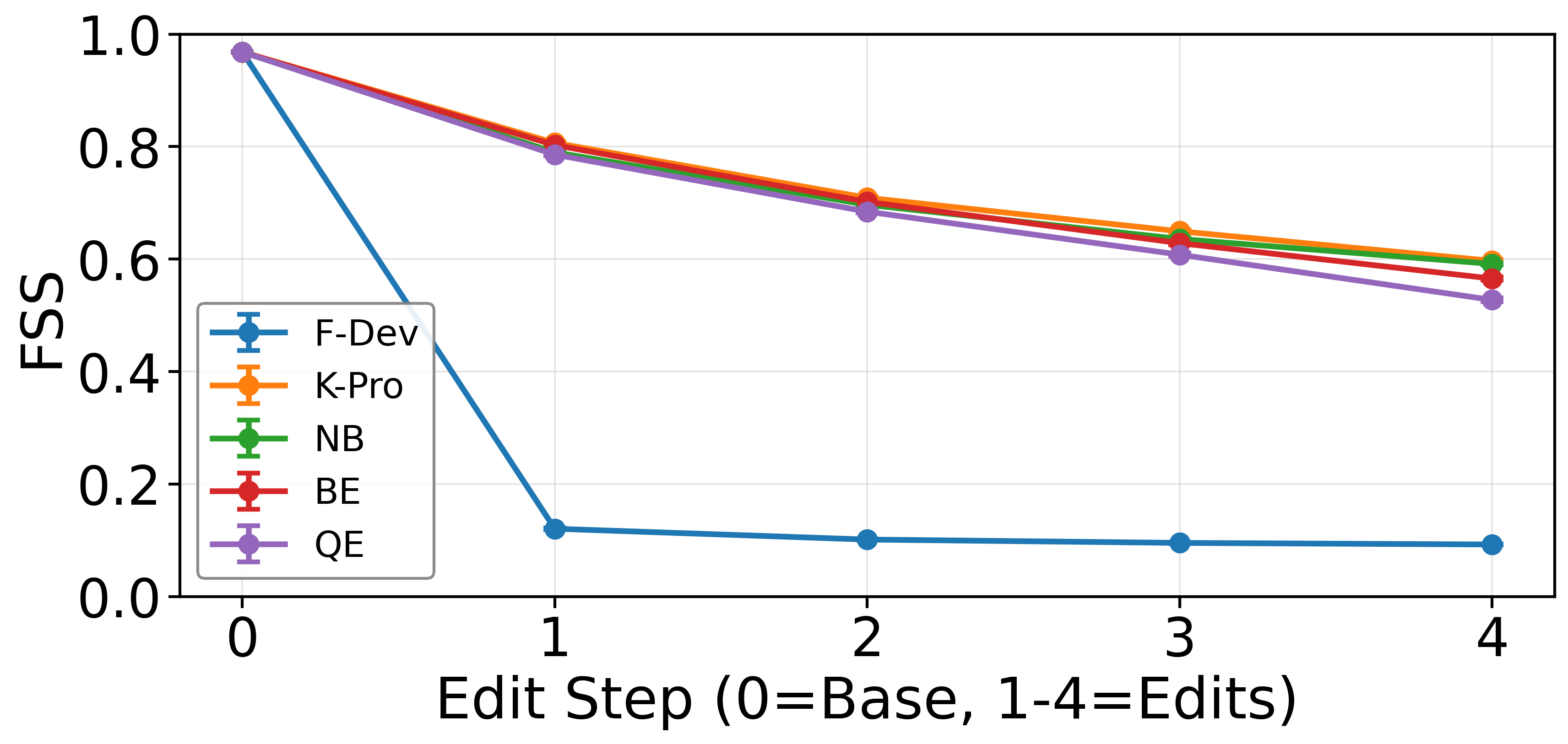}
    \vspace{-0.4cm}
    \caption{CelebA}
    \label{fig:first_image}
  \end{subfigure}\hfill
  \begin{subfigure}[t]{0.48\textwidth}
    \centering
    \includegraphics[width=\linewidth]{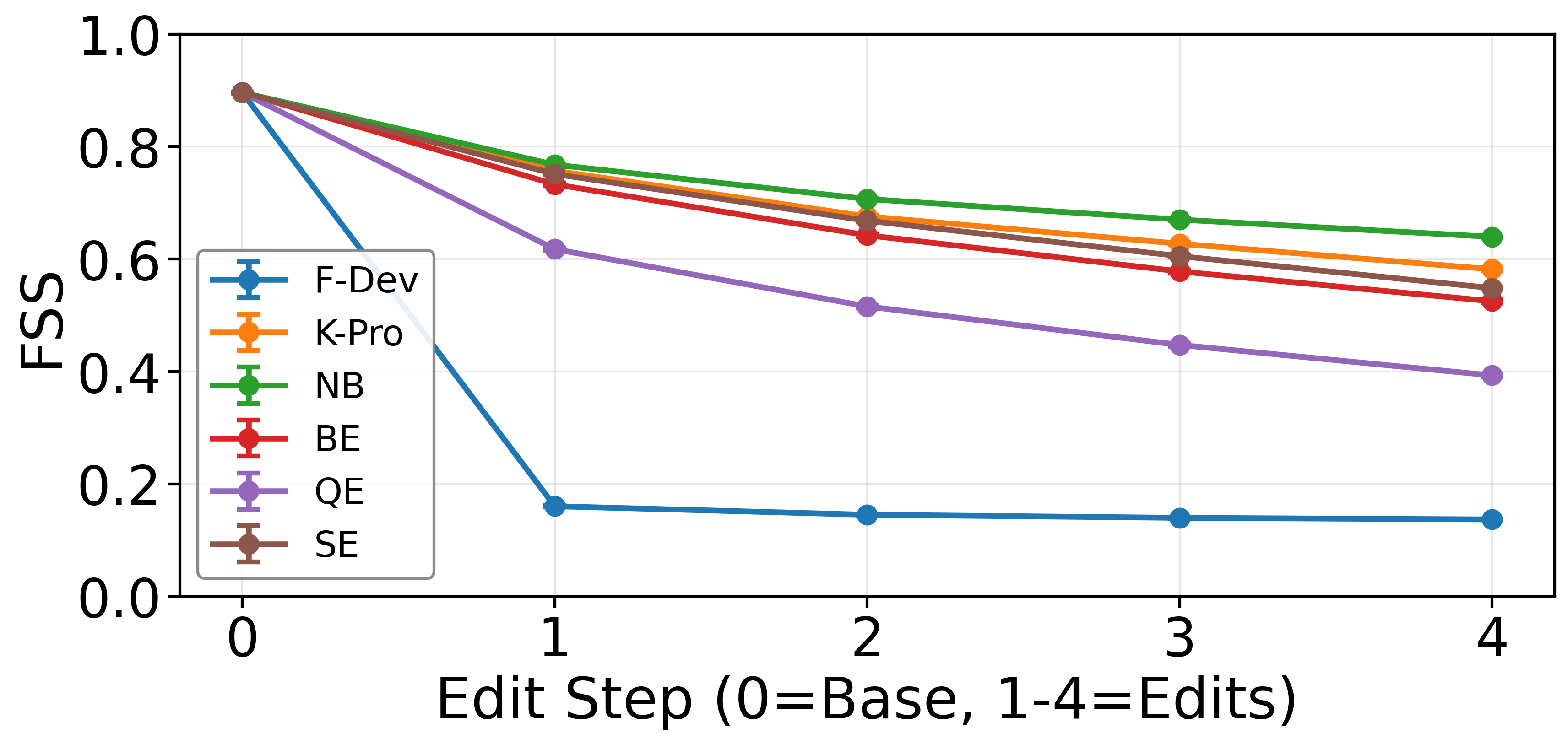}
    \vspace{-0.4cm}
    \caption{CelebSET}
    \label{fig:second_image}
  \end{subfigure}
  \vspace{-0.5cm}
  \caption{Facial similarity scores after each edit. The x-axis indicates the editing step, from $0$ (base image) to $4$ (final edited image). At each edit step, the Y-axis is the average facial similarity of a model for all images in that edit step. Note that $95\%$ confidence intervals are added, but are too narrow to be visible.}
  \label{fig:fss_progression}
  \vspace{-0.2cm}
\end{figure*}

\section{Stereotype Analysis}
\label{app:gendered_representation_analysis}

We conducted a controlled experiment to check if models performed edits better on male faces, if editing instructions are male-centric (e.g., add a handlebar mustache). From the $140$ instructions that we used for evaluation, we selected $17$ instructions that generally apply to males (e.g., add a pencil mustache, add a bushy beard). All of these instructions focused on editing facial hair. 

For this subset of instructions (called the male-centric instruction subset), we computed the average $SC$ across all models, for male and female groups in CelebA, and for four intersectional subgroups (male-light, female-light, male-dark, female-dark) in CelebSET. CelebA results are shown in Table \ref{tab:sc_celeba_stereo}, and CelebSET results are shown in Table \ref{tab:sc_celebset_stereo}. In the ``Mean'' column for CelebA (Table \ref{tab:sc_celeba_stereo}), the left value is the worst group, and the right value is the best group. For CelebSET (Table \ref{tab:sc_celebset_stereo}), we reported two worst groups (out of four intersectional groups): the left value is the worst, and the right is the second-worst performing group. For CelebA, we reported $WD_{g}$ and for CelebSET, we reported both $WD_{g}$ and $WD_{s}$. 

In CelebA (Table \ref{tab:sc_celeba_stereo}), we saw gender bias in three models. Out of these, only two models showed worse $SC$ on females (Kontext-Pro and Bagel-Edit). In CelebSET (Table \ref{tab:sc_celebset_stereo}), only for Flux-Dev, the worst and the second-worst groups were both females (FL and FD). For all other models, at least one male intersectional group was present in the two worst groups. 

For male-centric editing instructions, we did not find clear trends indicating that models perform edits better for males.

\section{Face ID Progression Analysis}
\label{app:face_id_progression}

We created line plots (in Figure \ref{fig:fss_progression}) to assess how facial similarity scores changed in models after each edit. In the x-axis, we marked the editing steps, starting from the base image (edit $0$) to the final edited image (edit $4$). At each edit step, the y-axis is the average facial similarity score (FSS) of a model (with $95\%$ confidence intervals) for all images in that edit step. The left plot in Figure \ref{fig:fss_progression} shows FSS when base images are from CelebA, while the right plot shows results on CelebSET. In both these plots, we showed the aggregated FSS across all four test sets (hair, accessories, pose, and multi-axis).

\textbf{For all models, the facial similarity score (FSS) consistently dropped with every edit.} In Figure \ref{fig:fss_progression}, for both CelebA and CelebSET, we see the biggest drop in FSS from edit $0$ to $1$. This is expected, as in edit $1$, we transition from a real image to a model-edited image. In subsequent edits, we see smaller (but consistent) drops.

\end{document}